\documentclass[10pt,journal,compsoc]{IEEEtran}
\ifCLASSOPTIONcompsoc
  \usepackage[nocompress]{cite}
\else
  \usepackage{cite}
\fi
\ifCLASSINFOpdf
\else
\fi
\usepackage{color}
\usepackage{tikz}
\usepackage{comment}
\usepackage{amsmath,amssymb,amssymb}
\usepackage{booktabs} 
\usepackage{mathrsfs}
\usepackage{multirow}
\usepackage{comment}
\usepackage{changepage}
\usepackage{wrapfig}
\usepackage{colortbl} 
\usepackage{arydshln} 
\usepackage{algorithm}  
\usepackage{algorithmic}  
\usepackage[colorlinks,linkcolor=red]{hyperref}
\usepackage{cleveref}

\begin{document}
\title{A Comprehensive Survey of Continual Learning: Theory, Method and Application}
%
% \title{A Comprehensive Review of Continual Learning: Theory, Method and Application}

\author{Liyuan Wang, Xingxing Zhang, Hang Su, Jun Zhu, \textit{Fellow}, \textit{IEEE} 
\thanks{
Liyuan Wang, Xingxing Zhang, Hang Su, and Jun Zhu are with Dept. of Comp. Sci. \& Tech., Institute for AI, BNRist Center, THBI Lab, Tsinghua-Bosch Joint Center for ML, Tsinghua University, Beijing, China (email: wly19@tsinghua.org.cn; xxzhang1993@gmail.com; \{suhangss, dcszj\}@tsinghua.edu.cn). Corresponding author: Jun Zhu.
%Liyuan Wang, Hang Su, and Jun Zhu are with Dept. of Comp. Sci. \& Tech., Institute for AI, BNRist Center, THBI Lab, Tsinghua-Bosch Joint Center for ML, Tsinghua University, Beijing, China; Liyuan Wang is also with School of Life Sciences, IDG/McGovern Institute for Brain Research, Tsinghua University, Beijing, China, and Tsinghua-Peking Center for Life Sciences, Beijing, China (email: wly19@mails.tsinghua.edu.cn; \{suhangss, dcszj\}@tsinghua.edu.cn). Xingxing Zhang is with Qiyuan Lab, Beijing, China (email: xxzhang1993@gmail.com). Corresponding author: Jun Zhu.
%Liyuan Wang, Hang Su and Jun Zhu are with Dept. of Comp. Sci. \& Tech., Institute for AI, BNRist Center, THBI Lab, Tsinghua-Bosch Joint Center for ML, Tsinghua University, Beijing, China (email: wly19@mails.tsinghua.edu.cn; \{suhangss, dcszj\}@tsinghua.edu.cn). Liyuan Wang is also with School of Life Sciences, IDG/McGovern Institute for Brain Research, Tsinghua University, Beijing, China, and Tsinghua-Peking Center for Life Sciences, Beijing, China. Xingxing Zhang is with Qiyuan Lab, Beijing, China (email: xxzhang1993@gmail.com). Corresponding author: Jun Zhu.
}
 
}
%ly\_wang94@126.com


\markboth{Journal of \LaTeX\ Class Files,~Vol.~14, No.~8, August~2015}%
{Wang \MakeLowercase{\textit{et al.}}}
\IEEEtitleabstractindextext{%
\begin{abstract}
To cope with real-world dynamics, an intelligent system needs to incrementally acquire, update, accumulate, and exploit knowledge throughout its lifetime. This ability, known as continual learning, provides a foundation for AI systems to develop themselves adaptively. In a general sense, continual learning is explicitly limited by catastrophic forgetting, where learning a new task usually results in a dramatic performance degradation of the old tasks. Beyond this, increasingly numerous advances have emerged in recent years that largely extend the understanding and application of continual learning. The growing and widespread interest in this direction demonstrates its realistic significance as well as complexity. In this work, we present a comprehensive survey of continual learning, seeking to bridge the basic settings, theoretical foundations, representative methods, and practical applications. Based on existing theoretical and empirical results, we summarize the general objectives of continual learning as ensuring a proper stability-plasticity trade-off and an adequate intra/inter-task generalizability in the context of resource efficiency. Then we provide a state-of-the-art and elaborated taxonomy, extensively analyzing how representative methods address continual learning, and how they are adapted to particular challenges in realistic applications. 
Through an in-depth discussion of promising directions, we believe that such a holistic perspective can greatly facilitate subsequent exploration in this field and beyond.
\end{abstract}

}

\maketitle

\IEEEdisplaynontitleabstractindextext

% \fi
\IEEEpeerreviewmaketitle

\IEEEraisesectionheading{\section{Introduction}\label{sec:introduction}}


%\IEEEPARstart{T}{his} demo file is intended to serve as a ``starter file'' for IEEE Computer Society journal papers produced under \LaTeX\ using IEEEtran.cls version 1.8b and latter.

\begin{figure}[t]
    \vspace{-.1cm}
	\centering
	\includegraphics[width=0.95\columnwidth]{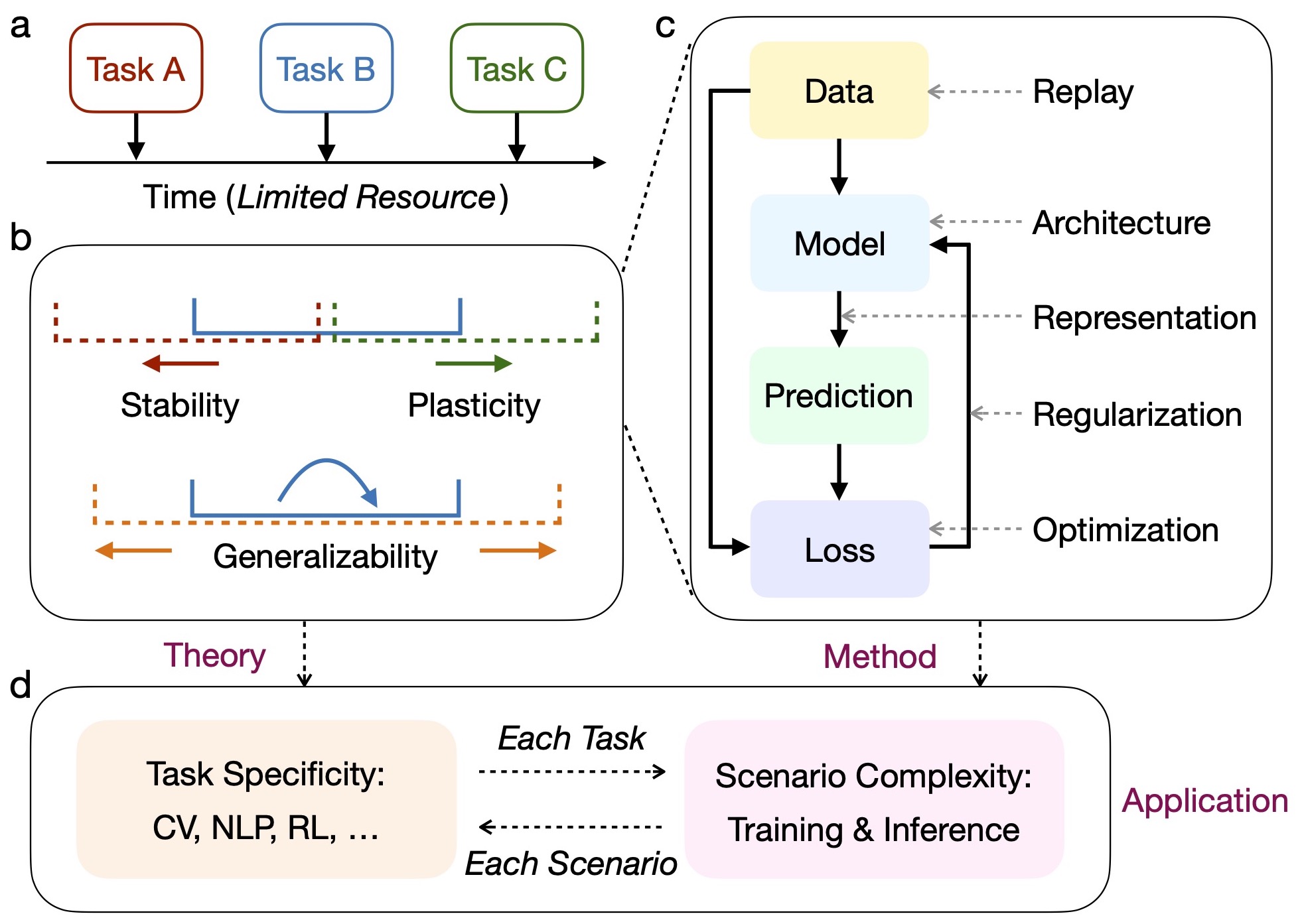} 
    \vspace{-.1cm}
 	\caption{A conceptual framework of continual learning. \textbf{a}, Continual learning requires adapting to incremental tasks with dynamic data distributions (Sec.~\ref{Sec.2_Setup}). \textbf{b}, A desirable solution should ensure an appropriate trade-off between stability (red arrow) and plasticity (green arrow), as well as an adequate generalizability to intra-task (blue arrow) and inter-task (orange arrow) distribution differences (Sec.~\ref{Sec.3_Theory}). \textbf{c}, To achieve the objective of continual learning, representative methods have targeted various aspects of machine learning (Sec.~\ref{Sec.4_Method}). \textbf{d}, Continual learning is adapted to practical applications to address particular challenges such as scenario complexity and task specificity (Sec.~\ref{Sec.5_Scenario}).} 
	\label{fig:CL_Framework}
    \vspace{-.4cm}
\end{figure}

Learning is the basis for intelligent systems to accommodate dynamic environments.
In response to external changes, evolution has empowered human and other organisms with strong adaptability to continually acquire, update, accumulate and exploit knowledge~\cite{kudithipudi2022biological,parisi2019continual,hadsell2020embracing}. 
Naturally, we expect artificial intelligence (AI) systems to adapt in a similar way.
This motivates the study of \textbf{continual learning}, where a typical setting is to learn a sequence of contents one by one and behave as if they were observed simultaneously (see Fig.~\ref{fig:CL_Framework}, a).
Such contents could be new skills, new examples of old skills, different environments, different contexts, etc., with particular realistic challenges incorporated~\cite{parisi2019continual,van2019three}.
As the contents are provided incrementally over a lifetime, continual learning is also referred to as \textbf{incremental learning} or \textbf{lifelong learning} in much of the literature, without a strict distinction~\cite{chen2018lifelong,kudithipudi2022biological}.

Unlike conventional machine learning models built on the premise of capturing a static data distribution, continual learning is characterized by learning from dynamic data distributions. A major challenge is known as \textbf{catastrophic forgetting}~\cite{mccloskey1989catastrophic,mcclelland1995there}, where adaptation to a new distribution generally results in a largely reduced ability to capture the old ones. 
This dilemma is a facet of the trade-off between \textbf{learning plasticity} and \textbf{memory stability}: an excess of the former interferes with the latter, and vice versa.
Beyond simply balancing the ``proportions'' of these two aspects, a desirable solution for continual learning should obtain strong \textbf{generalizability} to accommodate distribution differences within and between tasks (see Fig.~\ref{fig:CL_Framework}, b).
As a naive baseline, reusing all old training samples (if allowed) makes it easy to address the above challenges, but creates huge computational and storage overheads, as well as potential privacy issues. In fact, continual learning is primarily intended to ensure the \textbf{resource efficiency} of model updates, preferably close to learning only new training samples.

A number of continual learning methods have been proposed in recent years for various aspects of machine learning, which can be conceptually separated into five groups (see Fig.~\ref{fig:CL_Framework}, c): adding regularization terms with reference to the old model (regularization-based approach); approximating and recovering the old data distributions (replay-based approach); explicitly manipulating the optimization programs (optimization-based approach); learning robust and well-distributed representations (representation-based approach); and constructing task-adaptive parameters with a properly-designed architecture (architecture-based approach).
This taxonomy extends the commonly-used ones with current advances, and provides refined sub-directions for each category.
We summarize how these methods achieve the objective of continual learning, with an extensive analysis of their theoretical foundations and specific implementations.
In particular, these methods are \emph{closely connected}, e.g., regularization and replay ultimately act to rectify the gradient directions in optimization, and \emph{highly synergistic}, e.g., the efficacy of replay can be facilitated by distilling knowledge from the old model.

Realistic applications present particular challenges for continual learning, categorized into \emph{scenario complexity} and \emph{task specificity} (see Fig.~\ref{fig:CL_Framework}, d).
As for the former, for example, the task identity is probably missing in training and testing, and the training samples might be introduced in tiny batches or even one pass. Due to the expense and scarcity of data labeling, continual learning needs to be effective for few-shot, semi-supervised and even unsupervised scenarios.
As for the latter, although current advances mainly focus on visual classification, other vision domains such as object detection and semantic segmentation, as well as other related fields, such as conditional generation, reinforcement learning (RL), natural language processing (NLP) and ethic considerations, are receiving increasing attention with their own characteristics.
We summarize their particular challenges and analyze how continual learning methods are adapted to them.
%For example, the oracle of which task to perform is probably missing in training and testing, and training samples might be introduced in tiny batches or even one pass. Due to the expense and scarcity of data labeling, continual learning needs to be effective for few-shot, semi-supervised and even unsupervised training samples. Most of the current work in continual learning is targeted at visual classification tasks. Other typical vision scenarios such as conditional generation, semantic segmentation and object detection, as well as other directions such as reinforcement learning (RL), natural language processing (NLP) and scientific domains are also receiving increasing attention.

Considering the significant growth of interest in continual learning, we believe that such an \emph{up-to-date} and \emph{comprehensive} survey can provide a holistic perspective for subsequent work. 
Although there are some early surveys of continual learning with relatively broad coverage~\cite{parisi2019continual,chen2018lifelong}, important advances in recent years have not been incorporated. 
In contrast, the latest surveys typically capture only a partial aspect of continual learning, with respect to its biological underpinnings~\cite{hadsell2020embracing,hayes2021replay,kudithipudi2022biological,jedlicka2022contributions}, specialized settings for visual classification~\cite{masana2020class,de2021continual_review,qu2021recent,mai2022online,kilickaya2023towards}, as well as specific extensions for NLP~\cite{biesialska2020continual_review,ke2022continual_review} or RL~\cite{khetarpal2022towards}.
To the best of our knowledge, this is the first survey that systematically summarizes the latest advances in continual learning. %this is the only systematic overview of recent advances in continual learning. 
Building on such strengths, we provide an in-depth discussion of continual learning, in terms of current trends, cross-directional prospects and interdisciplinary connections with neuroscience. % such as diffusion models, large-scale pre-training, vision transformer, embodied AI, neural compression, etc.,

%Our main contributions include: (1) We present an up-to-date and comprehensive survey of continual learning, in order to bridge the advances in theory, method, and application; (2) We summarize the general objectives of continual learning based on existing theoretical and empirical results, with a state-of-the-art and elaborated taxonomy of representative methods; (3) We categorize the particular challenges of realistic applications into scenario complexity and task specificity, analyzing extensively how continual learning methods are adapted to them; and (4) We provide an in-depth discussion of current trends and promising directions to facilitate subsequent work in related fields.

The paper is organized as follows: In Sec.~\ref{Sec.2_Setup}, we introduce the setups of continual learning, including its basic formulation, typical scenarios and evaluation metrics. 
In Sec.~\ref{Sec.3_Theory}, we summarize the theoretical efforts on continual learning in response to its general objectives, which motivate the development of various continual learning methods.
In Sec.~\ref{Sec.4_Method}, we present a state-of-the-art and elaborated taxonomy of representative methods, analyzing their motivations and typical implementations.
%In Sec. 4, we present a state-of-the-art and elaborated taxonomy of representative strategies and analyze how they address continual learning.
In Sec.~\ref{Sec.5_Scenario} and~\ref{Sec.6_Task}, we describe how these methods are adapted to realistic applications in terms of scenario complexity and task specificity.
In Sec.~\ref{Sec.7_Discussion}, we further discuss current trends, cross-directional prospects and interdisciplinary connections with neuroscience.

\newcommand{\tabincell}[2]{\begin{tabular}{@{}#1@{}}#2\end{tabular}}
\begin{table*}[th]
    \renewcommand\arraystretch{2.10}
	\centering
   % \vspace{-0.4cm}
	\caption{A formal comparison of typical continual learning scenarios. $\mathcal{D}_{t,b}$: the training samples of task $t$ and batch $b$. %an incoming batch of 
	$| b |$: the size of batch $b$.
	$\mathcal{B}_t$: the space of incremental batches belonging to task $t$. $\mathcal{D}_{t}$: the training set of task $t$ (further specified as $\mathcal{D}_{t}^{pt}$ for pre-training). $\mathcal{T}$: the space of all incremental tasks (further specified as $\mathcal{T}^{pt}$ for pre-training). $\mathcal{X}_{t}$: the input data in $\mathcal{D}_{t}$. $p(\mathcal{X}_{t})$: the distribution of $\mathcal{X}_{t}$. $\mathcal{Y}_{t}$: the data label of $\mathcal{X}_{t}$. } 
     \vspace{-0.2cm}
	\smallskip
        \renewcommand\arraystretch{1.8}
	\resizebox{17 cm}{!}{ 
	\begin{tabular}{l|c|c}
	 \hline
	\,\,\,\,\,\,\,  Scenario & Training & Testing \\
    \hline
     IIL~\cite{lomonaco2017core50}&  $\{\{\mathcal{D}_{t,b}, t\}_{b \in \mathcal{B}_t}\}_{t = j}$ & $\{p(\mathcal{X}_{t})\}_{t = j}$; $t$ is not required  \\
          \hline
     DIL~\cite{hsu2018re,van2019three}& $\{\mathcal{D}_{t}, t \}_{t \in \mathcal{T}}$; $p(\mathcal{X}_{i}) \neq p(\mathcal{X}_{j})$ and $\mathcal{Y}_{i} = \mathcal{Y}_{j}$ for $i \neq j$ & $\{p(\mathcal{X}_{t})\}_{t \in \mathcal{T}}$, $t$ is not required  \\
          \hline
     TIL~\cite{hsu2018re,van2019three}& $\{\mathcal{D}_{t}, t \}_{t \in \mathcal{T}}$; $p(\mathcal{X}_{i}) \neq p(\mathcal{X}_{j})$ and $\mathcal{Y}_{i} \cap \mathcal{Y}_{j} = \emptyset $  for $i \neq j$ & $\{p(\mathcal{X}_{t})\}_{t \in \mathcal{T}}$; $t$ is available  \\
      \hline
     CIL~\cite{hsu2018re,van2019three}& $\{\mathcal{D}_{t}, t \}_{t \in \mathcal{T}}$; $p(\mathcal{X}_{i}) \neq p(\mathcal{X}_{j})$ and $\mathcal{Y}_{i} \cap \mathcal{Y}_{j} = \emptyset $  for $i \neq j$ & $\{p(\mathcal{X}_{t})\}_{t \in \mathcal{T}}$; $t$ is unavailable  \\
      \hline
     TFCL~\cite{aljundi2019task}& $\{\{\mathcal{D}_{t,b} \}_{b \in \mathcal{B}_t}\}_{t \in \mathcal{T}}$; $p(\mathcal{X}_{i}) \neq p(\mathcal{X}_{j})$ and $\mathcal{Y}_{i} \cap \mathcal{Y}_{j} = \emptyset $  for $i \neq j$ & $\{p(\mathcal{X}_{t})\}_{t \in \mathcal{T}}$; $t$ is optionally available
     % is available or unavailable  
     \\
      \hline
     OCL~\cite{aljundi2019gradient}& $\{\{\mathcal{D}_{t,b} \}_{b \in \mathcal{B}_t}\}_{t \in \mathcal{T}}$, $| b |=1$; $p(\mathcal{X}_{i}) \neq p(\mathcal{X}_{j})$ and $\mathcal{Y}_{i} \cap \mathcal{Y}_{j} = \emptyset $  for $i \neq j$ & $\{p(\mathcal{X}_{t})\}_{t \in \mathcal{T}}$; $t$ is optionally available  \\
     \hline
     BBCL~\cite{buzzega2020dark,bang2021rainbow}& $\{\mathcal{D}_{t}, t\}_{t \in \mathcal{T}}$; $p(\mathcal{X}_{i}) \neq p(\mathcal{X}_{j})$, $\mathcal{Y}_{i} \neq \mathcal{Y}_{j}$ and $\mathcal{Y}_{i} \cap \mathcal{Y}_{j} \neq \emptyset $  for $i \neq j$ & $\{p(\mathcal{X}_{t})\}_{t \in \mathcal{T}}$; $t$ is unavailable  \\
      \hline
     CPT~\cite{sun2020ernie}& $\{\mathcal{D}_{t}^{pt}, t \}_{t \in \mathcal{T}^{pt}}$, followed by a downstream task $j$  & $\{p(\mathcal{X}_{t})\}_{t=j}$; $t$ is not required\\
     %(x_{T}^{ds}, y_{T}^{ds}) \sim p(\mathcal{X}_{T}^{ds}, \mathcal{Y}_{T}^{ds})
       \hline
	\end{tabular}
	}
	\label{table:scenario}
	\vspace{-.2cm}
\end{table*}

\section{Setup}\label{Sec.2_Setup}
In this section, we first present a basic formulation of continual learning. Then we introduce typical scenarios and evaluation metrics.

\subsection{Basic Formulation}\label{Sec_2_1_Formulation}
Continual learning is characterized as learning from dynamic data distributions. In practice, training samples of different distributions arrive in sequence. A continual learning model parameterized by $\theta$ needs to learn corresponding task(s) with no or limited access to old training samples and perform well on their test sets. 
Formally, an incoming batch of training samples belonging to a task $t$ can be represented as $\mathcal{D}_{t,b} = \{ \mathcal{X}_{t,b}, \mathcal{Y}_{t,b}\}$, where $\mathcal{X}_{t,b}$ is the input data, $\mathcal{Y}_{t,b}$ is the data label, $t \in \mathcal{T}=\{1,\cdots,k\}$ is the task identity and $b \in \mathcal{B}_t$ is the batch index ($\mathcal{T}$ and $\mathcal{B}_t$ denote their space, respectively). Here we define a ``task'' by its training samples $\mathcal{D}_{t}$ following the distribution
$\mathbb{D}_{t} := p(\mathcal{X}_{t}, \mathcal{Y}_{t}) $ ($\mathcal{D}_{t}$
% Here we define a task by its training samples following the same data distribution $\mathcal{D}_{t} \sim p(\mathcal{X}_{t}, \mathcal{Y}_{t}) :=\mathbb{D}_{t} $ ($\mathcal{D}_{t}$
denotes the entire training set by omitting the batch index, likewise for $\mathcal{X}_{t}$ and $\mathcal{Y}_{t}$), and assume that there is no difference in distribution between training and testing. Under realistic constraints, the data label $\mathcal{Y}_{t}$ and the task identity $t$ might not be always available. In continual learning, the training samples of each task can arrive incrementally in batches (i.e., $\{\{\mathcal{D}_{t,b} \}_{b \in \mathcal{B}_t}\}_{t \in \mathcal{T}}$) or simultaneously (i.e., $\{\mathcal{D}_{t}\}_{t \in \mathcal{T}}$).

\subsection{Typical Scenario}\label{Sec.2.2_Scenario}
According to the division of incremental batches and the availability of task identities, we describe typical continual learning scenarios as follows (please refer to Table~\ref{table:scenario} for a formal comparison):
\begin{enumerate}[\leftmargin=-2em]
\item[$\bullet$] \emph{Instance-Incremental Learning} (IIL): 
All training samples belong to the same task and arrive in batches.

\item[$\bullet$] \emph{Domain-Incremental Learning} (DIL): Tasks have the same data label space but different input distributions. Task identities are not required.

\item[$\bullet$] \emph{Task-Incremental Learning} (TIL): Tasks have disjoint data label spaces. Task identities are provided in both training and testing.

\item[$\bullet$] \emph{Class-Incremental Learning} (CIL): Tasks have disjoint data label spaces. Task identities are only provided in training. % (also called \emph{task-free evaluation}).

\item[$\bullet$] \emph{Task-Free Continual Learning} (TFCL): Tasks have disjoint data label spaces. Task identities are not provided in either training or testing.

\item[$\bullet$] \emph{Online Continual Learning} (OCL): Tasks have disjoint data label spaces. Training samples for each task arrive as a one-pass data stream. 

\item[$\bullet$] \emph{Blurred Boundary Continual Learning} (BBCL): Task boundaries are blurred, characterized by distinct but overlapping data label spaces.

\item[$\bullet$] \emph{Continual Pre-training} (CPT): Pre-training data arrives in sequence. The goal is to improve knowledge transfer to downstream tasks.

%\item[$\bullet$] \emph{Hybrid}: A combination of the above scenarios.
%Combinatorial challenges of the above scenarios.

\end{enumerate}

If not specified, each task is often assumed to have a sufficient number of labeled training samples, i.e., \emph{Supervised Continual Learning}.
According to the provided $\mathcal{X}_{t}$ and $\mathcal{Y}_{t}$ in each $\mathcal{D}_{t}$, continual learning is further extended to zero-shot~\cite{ghosh2021dynamic,singh2021rectification}, few-shot~\cite{tao2020few}, semi-supervised~\cite{wang2021ordisco}, open-world (i.e., to identify unknown classes and then incorporate their labels)~\cite{joseph2021towards,wang2022learngene} and un-/self-supervised~\cite{rao2019continual,hu2021well} scenarios.
Besides, other practical challenges have been considered and incorporated, such as multiple labels~\cite{kim2020imbalanced}, noisy labels~\cite{kim2021continual,bang2022online}, hierarchical granularity~\cite{abdelsalam2021iirc} and sub-populations~\cite{liang2022balancing}, mixture of task similarity~\cite{ke2020continual}, long-tailed distribution~\cite{liu2022long}, domain alignment~\cite{lin2022prototype}, domain shifting~\cite{simon2022generalizing}, anytime inference~\cite{koh2021online}, novel class discovery~\cite{roy2022class,joseph2022novel}, multi-modality~\cite{srinivasan2022climb}, etc. 
Some recent work has focused on various combinations of these scenarios~\cite{lomonaco2017core50,mi2020generalized,xie2022general,bang2022online,koh2021online,caccia2020osaka}, as a way to better simulate the complexity of the real world.

\subsection{Evaluation Metric}\label{Sec.2.3_Metric}

In general, the performance of continual learning can be evaluated from three aspects: overall performance of the tasks learned so far, memory stability of old tasks, and learning plasticity of new tasks. Here we summarize the common evaluation metrics, using classification as an example.
%\hangx{we may add some comments about 1) why and how we define these metric 2) for which scenario 3) from what aspects, and move the details to appendix. }

\textbf{Overall performance} is typically evaluated by \emph{average accuracy} (AA)~\cite{chaudhry2018riemannian_rwalk,lopez2017gradient_gem} and \emph{average incremental accuracy} (AIA)~\cite{hou2019learning_lucir,douillard2020podnet}. Let ${a}_{k,j} \in [0,1]$ denote the classification accuracy evaluated on the test set of the $j$-th task after incremental learning of the $k$-th task ($j \leq k$). The output space to compute ${a}_{k,j}$ consists of the classes in either $\mathcal{Y}_{j}$ or $\cup_{i=1}^{k} \mathcal{Y}_{i}$, corresponding to the use of multi-head evaluation (e.g., TIL) or single-head evaluation (e.g., CIL)~\cite{chaudhry2018riemannian_rwalk}. The two metrics at the $k$-th task are then defined as

\vspace{-0.2cm}
\begin{equation} {{\rm{AA}}_{k}} = \frac{1}{k} \sum_{j=1}^{k} {a}_{k,j}, 
\end{equation}
\begin{equation} 
{{\rm{AIA}}_{k}} = \frac{1}{k} \sum_{i=1}^{k} {\rm{AA}}_{i}, 
\end{equation}
where AA represents the overall performance at the current moment and AIA further reflects the historical performance.

\textbf{Memory stability} can be evaluated by \emph{forgetting measure} (FM)~\cite{chaudhry2018riemannian_rwalk} and \emph{backward transfer} (BWT)~\cite{lopez2017gradient_gem}. As for the former, the forgetting of a task is calculated by the difference between its maximum performance obtained in the past and its current performance:
\begin{equation}
f_{j,k} = \max_{i \in \{1,...,k-1\}} ({a}_{i,j} - {a}_{k,j}), \forall j<k.
\end{equation}
FM at the $k$-th task is the average forgetting of all old tasks:
\begin{equation}
    {{\rm{FM}}_{k}} = \frac{1}{k-1} \sum_{j=1}^{k-1} f_{j,k}.
\end{equation}
As for the latter, BWT evaluates the average influence of learning the $k$-th task on all old tasks: 
%\begin{equation}
%    {\rm{BWT}}_k = \frac{1}{k-1} \sum_{j=1}^{k-1} {a}_{k,j} - {a}_{j,j},
%\end{equation}
\begin{equation}
    {\rm{BWT}}_k = \frac{1}{k-1} \sum_{j=1}^{k-1} ({a}_{k,j} - {a}_{j,j}),
\end{equation}
where the forgetting is usually reflected as a negative BWT.

\textbf{Learning plasticity} can be evaluated by \emph{intransience measure} (IM)~\cite{chaudhry2018riemannian_rwalk} and \emph{forward transfer} (FWT)~\cite{lopez2017gradient_gem}. IM is defined as the inability of a model to learn new tasks, calculated by the difference of a task between its joint training performance and continual learning performance:
\begin{equation}
    {\rm{IM}}_k = a_{k}^{*} - a_{k,k},
\end{equation}
where $a_{k}^{*}$ is the classification accuracy of a randomly-initialized reference model jointly trained with $\cup_{j=1}^{k} \mathcal{D}_{j}$ for the $k$-th task. In comparison, FWT evaluates the average influence of all old tasks on the current $k$-th task:
\begin{equation}
    {\rm{FWT}}_k = \frac{1}{k-1} \sum_{j=2}^{k} (a_{j,j} - \tilde{a}_{j}),
\end{equation}
where $\tilde{a}_{j}$ is the classification accuracy of a randomly-initialized reference model trained with $\mathcal{D}_{j}$ for the $j$-th task.
Note that, $a_{k,j}$ can be adapted to other forms depending on the task type, such as average precision (AP) for object detection~\cite{shmelkov2017incremental_ilod}, Intersection-over-Union (IoU) for semantic segmentation~\cite{cermelli2020modeling}, Fréchet Inception Distance (FID) for image generation~\cite{wu2018memory_mrgan}, normalized reward for reinforcement learning~\cite{ahn2019uncertainty}, etc, and the above evaluation metrics should be adjusted accordingly.

Besides, there are many other useful metrics, such as linear probes for representation forgetting~\cite{davari2022probing}, maximum eigenvalue of the Hessian matrix for flatness of loss landscape~\cite{mirzadeh2020understanding}, area under the curve of accuracy for anytime inference~\cite{koh2021online}, as well as the overheads of storage and computation for resource efficiency~\cite{chaudhry2018efficient_agem}, etc. We refer readers to their original papers.

\section{Theoretical Foundation}\label{Sec.3_Theory} % and Analysis

In this section, we summarize the theoretical efforts on continual learning with respect to both stability-plasticity trade-off and generalizability analysis, and relate them to the motivations of various continual learning methods.
%In this section, we summarize some theoretical efforts on continual learning, covering the general objectives of stability-plasticity trade-off and solution generalizability, and relate them to representative approaches.

\subsection{Stability-Plasticity Trade-off}
%\hangx{Important section. We may include more comments and remove the technical details, which will make our point more prominent}
%\hangx{\textbf{In general, we can include the conclusive equation in a survey but remove the derivation }}

\begin{figure}[t]
	\centering
	\includegraphics[width=0.99\columnwidth]{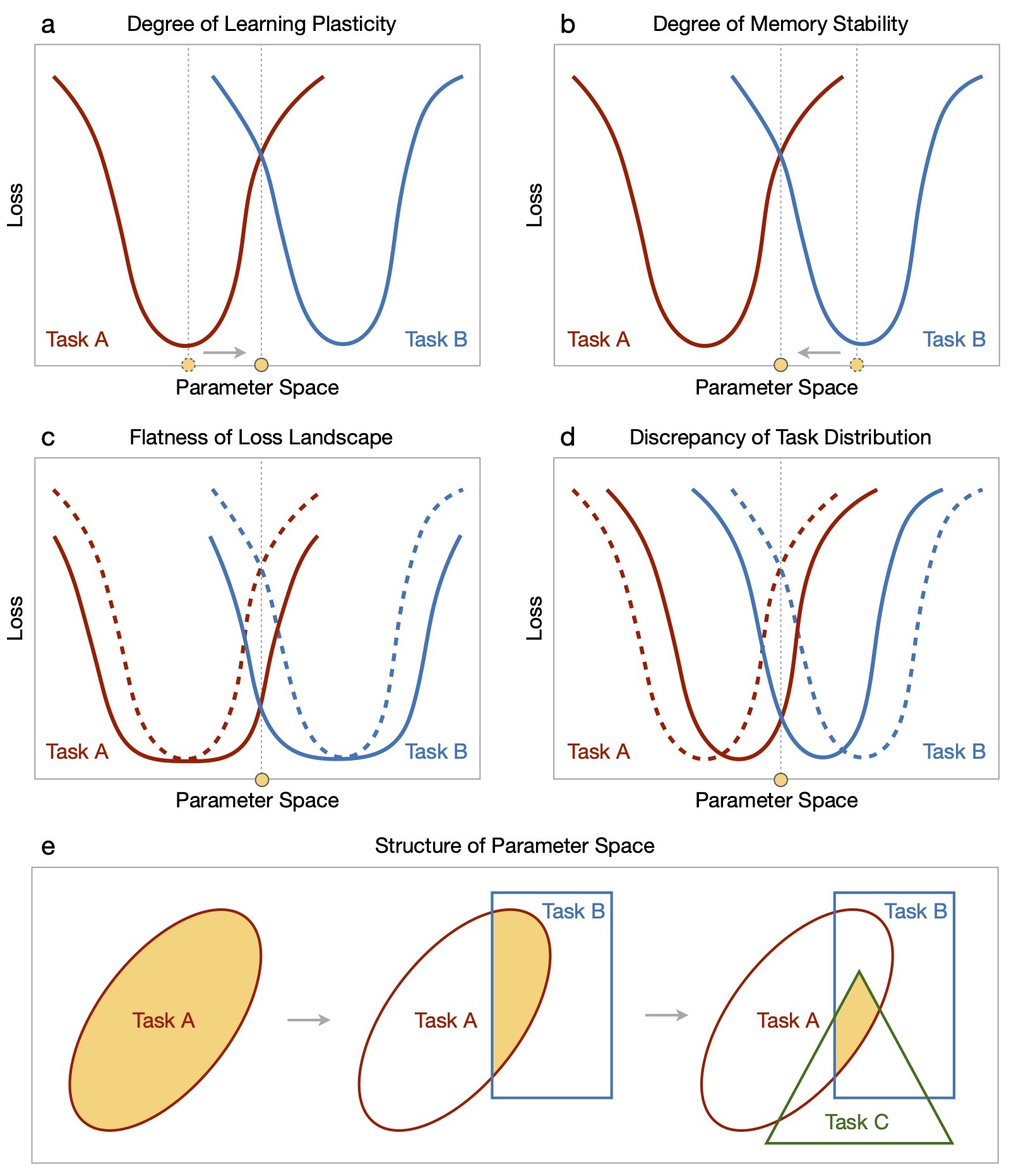} 
    %\vspace{-.1cm}
	\caption{Analysis of critical factors for continual learning. \textbf{a}, \textbf{b}, Continual learning requires a proper balance between learning plasticity and memory stability, where excess of either can affect the overall performance (adapted from~\cite{wang2022coscl}).
    \textbf{c}, \textbf{d}, When the converged loss landscape is flatter and the observed data distributions are more similar, a properly balanced solution can better generalize to the task sequence (adapted from~\cite{wang2022coscl}).
    \textbf{e}, The structure of parameter space determines the complexity and possibility of finding a desirable solution (adapted from~\cite{knoblauch2020optimal}). The yellow area indicates the feasible parameter space shared by individual tasks, which tends to be narrow and irregular as more incremental tasks are introduced. }
	\label{fig:Theory}
    \vspace{-.3cm}
\end{figure}

With the basic formulation in Sec.~\ref{Sec_2_1_Formulation}, let's consider a general setup for continual learning, where a neural network with parameters $\theta \in \mathbb{R}^{|\theta|}$
%\junz{use $\mathbb{R}$ to denote real space} 
needs to learn $k$ incremental tasks.
The training set and test set of each task are assumed to follow the same distribution $\mathbb{D}_{t}$, $t=1,...,k$, where the training set $\mathcal{D}_{t}=\{\mathcal{X}_{t}, \mathcal{Y}_{t}\} = \{(x_{t,n}, y_{t,n})\}_{n=1}^{N_t}$ includes $N_t$ data-label pairs. %$\mathcal{D}_{t}=\{(x_{t,n}, y_{t,n})\}_{n=1}^{N_t}$
The objective is to learn a probabilistic model $p(\mathcal{D}_{1:k}|\theta) = \prod_{t=1}^{k} p(\mathcal{D}_{t} | \theta)$  (with assumption of conditional independence)
%($\mathcal{D}_{1:k} := \{\mathcal{D}_{1},...,\mathcal{D}_{k}\}$)
that can perform well on all tasks denoted as $\mathcal{D}_{1:k} := \{\mathcal{D}_{1},...,\mathcal{D}_{k}\}$.
The task-related performance for discriminative models can be expressed as $\log p(\mathcal{D}_{t}|\theta) = \sum_{n=1}^{N_t} \log p_{\theta}(y_{t,n}|x_{t,n})$. The central challenge of continual learning generally arises from the sequential nature of learning: when learning the $k$-th task from $\mathcal{D}_{k}$, the old training sets $\{\mathcal{D}_{1},...,\mathcal{D}_{k-1}\}$ are inaccessible. Therefore, it is critical but difficult to capture the distributions of both old and new tasks in a balanced manner, i.e., ensuring an appropriate \textbf{stability-plasticity trade-off}, where excessive learning plasticity or memory stability can largely compromise each other (see Fig.~\ref{fig:Theory}, a, b).

A straightforward idea is to approximate and recover the old data distributions by storing a few old training samples or training a generative model, known as the \emph{replay-based approach} in Sec.~\ref{Sec.4.2_Replay}.
According to the learning theory for supervised learning~\cite{hastie2009elements}, the performance of an old task is improved with replaying more old training samples that approximate its distribution, but resulting in potential privacy issues and a linear increase in \textbf{resource overhead}. The use of generative models is also limited by a huge additional resource overhead, as well as their own catastrophic forgetting and expressiveness. %, which are further detailed in Sec.~\ref{Sec.4.2_Replay}.
% Based on the supervised learning theory, using more training samples of the same distribution can improve the performance, but requiring more resource and privacy issue. On the other hand, training a generative model is limited by catastrophic forgetting and expresiveness of itself, as well as resource comsumption.

% To avoid saving old training samples and the overhead of training a generative model,
An alternative choice is to propagate the old data distributions in updating parameters through formulating continual learning in a Bayesian framework. Based on a prior $p(\theta)$ of the network parameters, the posterior after observing the $k$-th task can be computed with Bayes' rule:
\begin{equation}
    p(\theta | \mathcal{D}_{1:k}) \propto p(\theta) \prod_{t=1}^{k} p(\mathcal{D}_{t} | \theta) 
    \propto p(\theta | \mathcal{D}_{1:k-1}) p(\mathcal{D}_{k} | \theta),
\label{eqn_cl_posterior}
\end{equation}
where the posterior $p(\theta | \mathcal{D}_{1:k-1})$ of the ($k$-$1$)-th task becomes the prior of the $k$-th task, and thus enables the new posterior $p(\theta | \mathcal{D}_{1:k})$ to be computed with only the current training set $\mathcal{D}_{k}$. However, as the posterior is generally intractable (except very special cases), a common option is to approximate it with $q_{k-1}(\theta) \approx p(\theta | \mathcal{D}_{1:k-1})$, likewise for $q_{k}(\theta) \approx p(\theta | \mathcal{D}_{1:k})$. In the following, we will introduce two widely-used approximation strategies:
%the posterior $p(\theta | \mathcal{D}_{1:k-1})$ by a parametric distribution $q_{k-1}(\theta):=q(\theta ; \phi_{k-1})$ with parameters $\phi_{k-1}$ for learning the $k$-th task~\cite{kao2021natural}. Below, we will introduce two representative strategies:

The first is online \emph{Laplace approximation}, which approximates $p(\theta | \mathcal{D}_{1:k-1})$ as a multivariate Gaussian with local gradient information~\cite{kirkpatrick2017overcoming,huszar2017quadratic,ritter2018online,kao2021natural,wang2021afec}. 
Specifically, we can parameterize $q_{k-1}(\theta)$ with $\phi_{k-1}$ and construct an approximate Gaussian posterior $q_{k-1}(\theta) := q(\theta ; \phi_{k-1}) = \mathcal{N}(\theta; \mu_{k-1}, \Lambda_{k-1}^{-1})$ through performing a second-order Taylor expansion around the mode $\mu_{k-1} \in \mathbb{R}^{|\theta|}$ of $p(\theta | \mathcal{D}_{1:k-1})$, where $\Lambda_{k-1}$ denotes the precision matrix and $\phi_{k-1} = \{ \mu_{k-1}, \Lambda_{k-1}\}$, likewise for $q(\theta ; \phi_{k})$, $\mu_{k}$ and $\Lambda_{k}$.
According to Eq.~\eqref{eqn_cl_posterior}, the posterior mode for learning the current $k$-th task can be computed as
%Specifically, learning the current $k$-th task while remembering the previous 1:($k$-1)-th tasks is equivalent to finding a mode $\mu_k$ of the posterior $p(\theta | \mathcal{D}_{1:k}) \propto p(\mathcal{D}_{k} | \theta)p(\theta | \mathcal{D}_{1:k-1})$. Then we can parameterize $q_{k}(\theta)$ with $\phi_{k}$ and construct an approximate Gaussian posterior $q_{k}(\theta) := q(\theta ; \phi_{k}) = \mathcal{N}(\theta; \mu_k, \Lambda_k^{-1})$ through performing a second-order Taylor expansion around $\mu_k \in \mathcal{R}^{|\theta|}$, where $\phi_{k} = \{ \mu_k, \Lambda_k \}$ and $\Lambda_k$ denotes the precision matrix, likewise for $q(\theta ; \phi_{k-1})$, $\mu_{k-1}$ and $\Lambda_{k-1}$. According to Eq.~\eqref{eqn_cl_posterior}, the posterior mode on the $k$-th task can be computed as
\begin{equation}
\begin{split}
    \mu_k &= {\arg\max}_{\theta} \log p(\theta | \mathcal{D}_{1:k}) \\
    & \approx {\arg\max}_{\theta} \log p(\mathcal{D}_{k} | \theta) + \log q(\theta ; \phi_{k-1}) \\
    & = {\arg\max}_{\theta} \log p(\mathcal{D}_{k} | \theta) - \frac{1}{2} (\theta - \mu_{k-1})^\top \Lambda_{k-1} (\theta - \mu_{k-1}),
\end{split}
\label{laplace_mode}
\end{equation} %\ell_{k}(\theta)
which is updated recursively from $\mu_{k-1}$ and $\Lambda_{k-1}$.
Meanwhile, $\Lambda_k$ is updated recursively from $\Lambda_{k-1}$:
\begin{equation}
\begin{split}
    \Lambda_k &= - \nabla_{\theta}^{2} \log p(\theta | \mathcal{D}_{1:k})\big|_{\theta = \mu_k} \\
    &\approx - \nabla_{\theta}^{2} \log p(\mathcal{D}_{k}|\theta)\big|_{\theta = \mu_k} + \Lambda_{k-1},
\label{laplace_precision}
\end{split}
\end{equation}
where the first term on the right side is the Hessian of the negative log likelihood of $\mathcal{D}_{k}$ at $\mu_k$, denoted as $H(\mathcal{D}_{k}, \mu_k)$. In practice, $H(\mathcal{D}_{k}, \mu_k)$ is often computationally inefficient due to the great dimensionality of $\mathbb{R}^{|\theta|}$, and there is no guarantee that the approximated $\Lambda_k$ is positive semi-definite for the Gaussian assumption. To overcome these issues, the Hessian is usually approximated by the Fisher information matrix (FIM):
\begin{equation}
F_k = \mathbb{E}[\nabla_{\theta} \log p(\mathcal{D}_{k}|\theta)\nabla_{\theta} \log p(\mathcal{D}_{k}|\theta)^{\top}]\big|_{\theta = \mu_k} \approx H(\mathcal{D}_{k}, \mu_k).
\label{eqn_fim}
\end{equation}
For ease of computation, the FIM can be further simplified with a diagonal approximation~\cite{kirkpatrick2017overcoming,huszar2017quadratic} or a Kronecker-factored approximation~\cite{ritter2018online,martens2015optimizing}. Then, Eq.~\eqref{laplace_mode} is implemented by saving a frozen copy of the old model $\mu_{k-1}$ to regularize parameter changes, known as the \emph{regularization-based approach} in Sec.~\ref{Sec.4.1_Regularization}. Here, we use EWC~\cite{kirkpatrick2017overcoming} as an example and present its loss function:
% \begin{equation} 
% \ell_{{\rm{EWC}}}(\theta) = \ell_k(\theta) + \frac{\lambda}{2} \sum_{i} F_{1:k-1,i} (\theta_i - \mu_{k-1,i})^2, 
% \label{eqn_EWC}
% \end{equation}
\begin{equation} 
\mathcal{L}_{{\rm{EWC}}}(\theta) = \ell_k(\theta) + \frac{\lambda}{2}   (\theta - \mu_{k-1})^\top \hat{F}_{1:k-1}  (\theta - \mu_{k-1}),
% F_{1:k-1} \left \| \theta - \mu_{k-1} \right \|_2^2, 
\label{eqn_EWC}
\end{equation}
where $\ell_k$ denotes the task-specific loss, the FIM $\hat{F}_{1:k-1} = \sum_{t=1}^{k-1} {\rm{diag}}({F}_{t})$ with a diagonal approximation ${\rm{diag}}(\cdot)$ of each ${F}_{t}$, and $\lambda$ is a hyperparameter to control the strength of regularization. 

The second is online \emph{variational inference} (VI)~\cite{nguyen2018variational,swaroop2019improving,adel2019continual,kurle2019continual,li2020variational,loo2020generalized,kapoor2021variational,rudner2022continual,derakhshani2021kernel}. There are many different ways to do this, and a representative one is to minimize the following KL-divergence over a family $\mathcal{Q}$ that satisfies $p(\theta|\mathcal{D}_{1:k}) \in \mathcal{Q}$ at 
the current $k$-th task: %every step $k$:
\begin{equation}
q_k(\theta) = {\arg \min}_{q \in \mathcal{Q}} {\rm{KL}}(q(\theta) \parallel \frac{1}{Z_{k}} q_{k-1}(\theta)p(\mathcal{D}_{k}|\theta)),
\end{equation}
where $Z_{k}$ is the normalizing constant of $q_{k-1}(\theta)p(\mathcal{D}_{k}|\theta)$. In practice, the above minimization can be achieved by employing an additional Monte Carlo approximation, with specifying $q_k(\theta) := q(\theta ; \phi_{k}) = \mathcal{N}(\theta; \mu_k, \Lambda_k^{-1})$ as a multivariate Gaussian. Here we use VCL~\cite{nguyen2018variational} as an example, which minimizes the following objective (i.e., maximize its negative):
%and present the minimization objective as follows:
\begin{equation}
\mathcal{L}_{{\rm{VCL}}}(q_k(\theta)) = \mathbb{E}_{q_k(\theta)}(\ell_k(\theta)) + {\rm{KL}}(q_k(\theta) \parallel q_{k-1}(\theta)),
\label{eqn_VCL}
\end{equation}
where the KL-divergence can be computed in a closed-form and serves as an implicit regularization term.
In particular, although the loss functions of Eq.~\eqref{eqn_EWC} and Eq.~\eqref{eqn_VCL} take similar forms, the former is a local approximation at a set of deterministic parameters $\theta$, while the latter is computed by sampling from the variational distribution $q_{k}(\theta)$. 
This is attributed to the fundamental difference between the two approximation strategies~\cite{nguyen2018variational,tserannatural}, with slightly different performance in particular settings. 

In addition to the parameter space, the idea of sequential Bayesian inference is also applicable to the function space~\cite{titsias2019functional,pan2020continual,rudner2022continual}, which tends to enable more flexibility in updating parameters. Also, there are many other extensions of VI, such as improving posterior updates with variational auto-regressive Gaussian processes (VAR-GPs)~\cite{kapoor2021variational}, constructing task-specific parameters~\cite{adel2019continual,loo2020generalized,kumar2021bayesian,lee2019neural_cndpm}, and adapting to a non-stationary data stream~\cite{kurle2019continual}.

%task-specific / adaptive weights in a data-driven fashion~\cite{adel2019continual}, online non-stationary datastream~\cite{kurle2019continual}, improve VCL with task-specific FiLM layers~\cite{loo2020generalized}, sequential function-space variational inference~\cite{rudner2022continual}, improved posterior updating strategy with variational auto-regressive Gaussian processes (VAR-GPs)~\cite{kapoor2021variational}, construct task-specific weights~\cite{kumar2021bayesian}
% , or optimizing each new task in the orthogonal or null space of the old tasks~\cite{zeng2019continual,wang2021training,farajtabar2020orthogonal,saha2020gradient,kao2021natural,kong2022balancing,liu2021continual,lin2022towards,peng2022continual,chaudhry2020continual,guo2022adaptive}.

In essence, the constraint on continual learning for either replay or regularization is ultimately reflected in gradient directions. As a result, some recent work directly manipulates the gradient-based optimization process, categorized as the \emph{optimization-based approach} in Sec.~\ref{Sec.4.3_Optimization}. Specifically, when a few old training samples $\mathcal{M}_t$ for task $t$ are maintained in a memory buffer, gradient directions of the new training samples are encouraged to stay close to that of the $\mathcal{M}_t$~\cite{lopez2017gradient_gem,chaudhry2018efficient_agem,tang2021layerwise}.
This is formulated as $\left \langle \nabla_{\theta} \mathcal{L}_k(\theta; \mathcal{D}_k), \nabla_{\theta} \mathcal{L}_k(\theta; \mathcal{M}_t) \right \rangle   \geq 0$ for $t \in \{1,...,k-1\}$,
%$t<k$
% Eq.~\eqref{eqn_replay_grad}, 
so as to essentially enforce non-increase in the loss of old tasks, i.e., 
$\mathcal{L}_k(\theta; \mathcal{M}_t) \leq \mathcal{L}_k(\theta_{k-1}; \mathcal{M}_t) $, where $\theta_{k-1}$ is the network parameters at the end of learning the ($k$-1)-th task. %\theta_{1:k-1}

Alternatively, gradient projection can also be performed without storing old training samples~\cite{zeng2019continual,wang2021training,farajtabar2020orthogonal,saha2020gradient,kao2021natural,kong2022balancing,liu2021continual,lin2022towards,peng2022continual,chaudhry2020continual,guo2022adaptive}.
Here we take NCL~\cite{kao2021natural} as an example, which manipulates gradient directions with $\mu_{k-1}$ and $\Lambda_{k-1}$ in online Laplace approximation.
As shown in Eq.~\eqref{eqn_reg_grad}, NCL performs continual learning by minimizing the task-specific loss $\ell_k(\theta)$ within a region of radius $r$ centered around $\theta$ with a distance metric $d(\theta, \theta+\delta)=\sqrt{\delta^\top\Lambda_{k-1}\delta/2}$ that
takes into account the curvature of the prior via its precision matrix $\Lambda_{k-1}$:
\begin{equation}
\begin{split}
% &d(\theta, \theta+\delta)=\sqrt{\delta^\top\Lambda_{k-1}\delta/2}\\
\delta^* &= \arg \min_{\delta}\ell_k(\theta+\delta) \\
& \approx \arg \min_{\delta} \ell_k(\theta)+\nabla_{\theta} \ell_k(\theta)^\top \delta, \\
% & \propto \Lambda_{k-1}^{-1}\nabla_{\theta} \ell_k(\theta)-(\theta-\mu_{k-1})\\
& \mathrm{s.t.}, d(\theta, \theta+\delta)=\sqrt{\delta^\top\Lambda_{k-1}\delta/2} \leq r.
\label{eqn_reg_grad}
\end{split}
\end{equation}
The solution to such an optimization problem in Eq.~\eqref{eqn_reg_grad} is given by $\delta^* \propto \Lambda_{k-1}^{-1}\nabla_{\theta} \ell_k(\theta)-(\theta-\mu_{k-1})$, which gives rise to the following update rule for a learning rate $\lambda$:
\begin{equation}
\begin{split}
\theta \leftarrow \theta+\lambda [ \Lambda_{k-1}^{-1}\nabla_{\theta} \ell_k(\theta)-(\theta-\mu_{k-1})],
\label{eqn_update_rule}
\end{split}
\end{equation}
in which the first term encourages parameter changes predominantly in directions that do not interfere with the old tasks via a preconditioner $\Lambda_{k-1}^{-1}$, while the second term enforces $\theta$ to stay close to the old task solution $\mu_{k-1}$.

%We categorize the above two strategies as the \textbf{optimization-based approach} and provide a detailed discussion in Sec. 4.3.
Of note, the above analyses are mainly based on finding a shared solution for all incremental tasks, which is subject to severe inter-task interference~\cite{wang2021afec,ramesh2021model,wang2022coscl}. 
In contrast, incremental tasks can also be learned in a (partially) separated way, which is the dominant idea of the \emph{architecture-based approach} in Sec.~\ref{Sec.4.5_Architecture}. This can be formulated as constructing a continual learning model with parameters $\theta = \cup_{t=1}^{k} \theta^{(t)}$, where $\theta^{(t)} = \{e^{(t)}, \psi\}$, $e^{(t)}$ is the task-specific/adaptive parameters, and $\psi$ is the task-sharing parameters. 
%Note that $e^{(t)}$ and $\psi$ are usually network parameters or their matrix decomposition. 
%can be induced by different modules of the network (e.g., different layers) or a matrix decomposition of parameters.
%, \xx{serving as a fixed point or sampled from a distribution}.
The task-sharing parameters $\psi$ are omitted in some cases, where the task-specific parameters $e^{(i)}$ and $e^{(j)}$ ($i<j$) may overlap to enable parameter reuse and knowledge transfer. The overlapping part $e^{(i)} \cap e^{(j)}$ is frozen when learning the $j$-th task to avoid catastrophic forgetting~\cite{rusu2016progressive,serra2018overcoming}. Then, each task can be performed as $p(\mathcal{D}_{t}|\theta^{(t)})$ instead of $p(\mathcal{D}_{t}|\theta)$ if given the task identity 
$\mathbb{I}_{\mathcal{D}_{t}}$, in which the conflicts between tasks are explicitly controlled or even avoided if $\psi$ is omitted:
\vspace{-0.1cm}
\begin{equation}
\begin{split}
p(\mathcal{D}_{t} | \theta) &= \sum_{i=1}^k p(\mathcal{D}_{t} |\mathbb{I}_{\mathcal{D}_{t}} =  i, \theta) p(\mathbb{I}_{\mathcal{D}_{t}} =  i |\theta) \\
&= p(\mathcal{D}_{t} |\mathbb{I}_{\mathcal{D}_{t}} =  t, \theta) p(\mathbb{I}_{\mathcal{D}_{t}} =  t |\mathcal{D}_{t}, \theta) \\
 &= p(\mathcal{D}_{t} |\theta^{(t)}) p(\mathbb{I}_{\mathcal{D}_t }=  t |\mathcal{D}_{t}, \theta) \\
 &=p(\mathcal{D}_{t} |e^{(t)}, \psi) p(\mathbb{I}_{\mathcal{D}_t }=  t |\mathcal{D}_{t}, \theta).
\label{eqn_task_inference}
\end{split}
\end{equation}
%\vspace{-0.1cm}
However, there are two major challenges. The first is the scalability of model size due to progressive allocation of $\theta^{(t)}$, which depends on the sparsity of $e^{(t)}$, reusability of $e^{(i)} \cap e^{(j)}$ ($i<j$), and transferability of $\psi$. % (detailed in Sec.~\ref{Sec.4.5_Architecture}). 
The second is the accuracy of task-identity prediction, denoted as $p(\mathbb{I}_{\mathcal{D}_{t}} =  t |\mathcal{D}_{t}, \theta)$. %where $\mathbb{I}_{\mathcal{D}_{t}}$ is the task label of $\mathcal{D}_{t}$. 
Except for the TIL setting that always provides the task identity $\mathbb{I}_{\mathcal{D}_{t}}$~\cite{rusu2016progressive,serra2018overcoming,fernando2017pathnet,ebrahimi2019uncertainty}, other scenarios generally require the model to determine which $\theta^{(t)}$ to use based on the input data, as shown in Eq.~\eqref{eqn_task_inference}. This is closely related to the out-of-distribution (OOD) detection, where the predictive uncertainty should be low for in-distribution data and high for OOD data~\cite{henning2021posterior,kim2022theoretical,d2021uncertainty}. %(i.e., the ones following $p(X_{t})$)
More importantly, since the function of task-identity prediction as Eq.~\eqref{eqn_task_inference_2} (equivalent to classifying tasks) needs to be continually updated, it also suffers from catastrophic forgetting. 
%To address this issue, the memory buffer~\cite{henning2021posterior,kj2020meta,gong2022continual} or generative model~\cite{lee2019neural_cndpm} is usually incorporated to recover the $i$-th task's distribution $p(\mathcal{D}_{t} |  i,\theta)$:
To address this issue, the $i$-th task's distribution $p(\mathcal{D}_{t} |  i,\theta)$ can be recovered by incorporating replay~\cite{henning2021posterior,kj2020meta,gong2022continual,lee2019neural_cndpm}:
\begin{equation} \label{eqn_task_inference_2}
p(\mathbb{I}_{\mathcal{D}_t }=  i |\mathcal{D}_{t},\theta) \propto
p(\mathcal{D}_{t} |  i,\theta)p(i),
\end{equation}
where the marginal task distribution $p(i) \propto N_i$ in general.
\vspace{-.2cm}

\begin{figure*}[ht]
	\centering	\includegraphics[width=1.95\columnwidth]{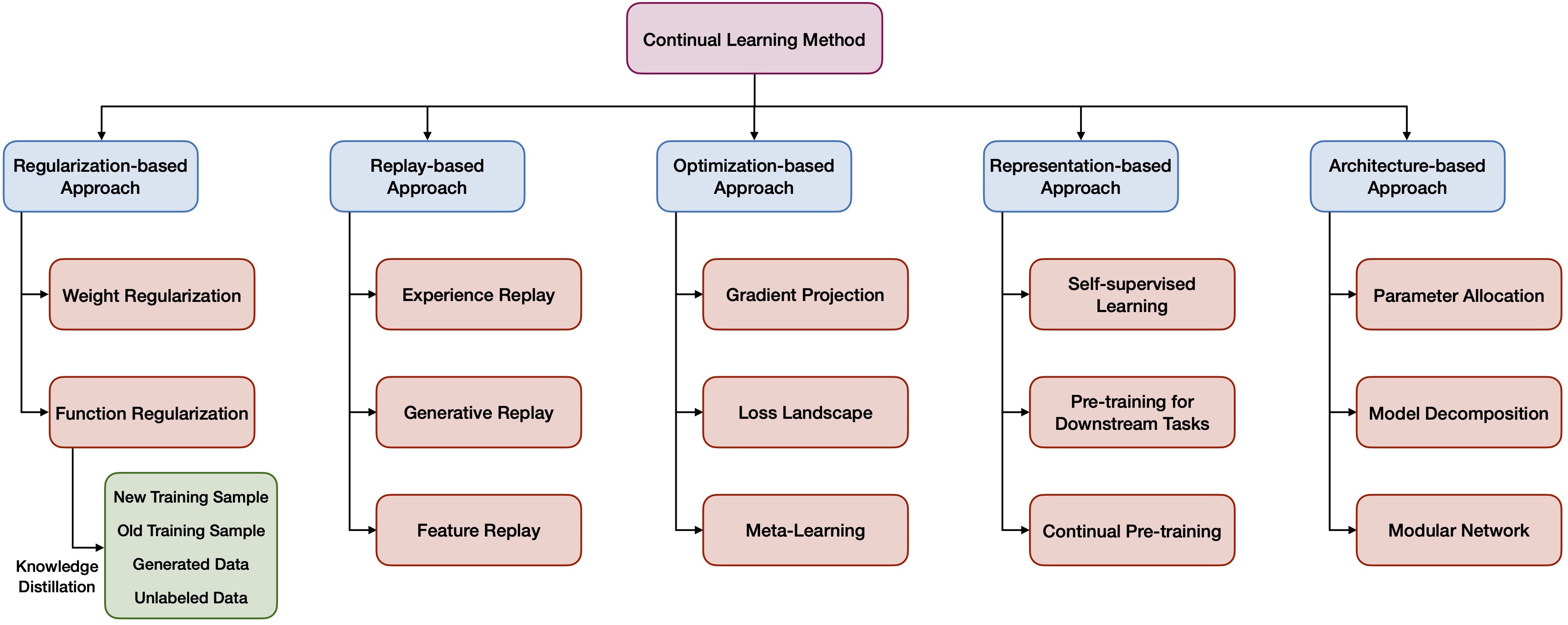} 
    \vspace{-.1cm}
	\caption{A state-of-the-art and elaborated taxonomy of representative continual learning methods. We have summarized five main categories (blue blocks), each of which is further divided into several sub-directions (red blocks).} %The solid black line demonstrates the hierarchical relationship of different (sub-)directions, while the dashed gray line denotes the connections between methods.
	\label{fig:Method}
    \vspace{-.2cm}
\end{figure*}

\subsection{Generalizability Analysis}
% \subsection{Solution Generalizability}
%\xx{Currently,} The theoretical analysis of continual learning is usually performed 
Current theoretical efforts for continual learning have primarily been performed on training sets of incremental tasks, assuming that their test sets follow similar distributions and the candidate solutions have similar generalizability. However, since the objective for learning multiple tasks is typically highly non-convex, there are many local optimal solutions that perform similarly on each training set but have significantly different generalizability on test sets~\cite{wang2022coscl,mirzadeh2020understanding}. 
For continual learning, a desirable solution requires not only \emph{intra-task generalizability} from training sets to test sets, but also \emph{inter-task generalizability} to accommodate to accommodate incremental changes of their distributions.
%For continual learning, a desirable solution needs not only to generalize from training sets to test sets, but also to accommodate distribution differences between tasks.
%For continual learning, the generalizability of a solution determines not only the generalization gap, i.e., the expected difference between training and testing errors, but also its compatibility with distribution differences between tasks. 

Here we provide a conceptual demonstration with a task-specific loss $\ell_t(\theta; \mathcal{D}_t)$ and its empirical optimal solution $\theta_t^{*} = {\arg \min}_{\theta} \ell_t(\theta; \mathcal{D}_t)$. When a task $i$ needs to accommodate another task $j$, the maximum sacrifice of its performance can be estimated by performing a second-order Taylor expansion of
$\ell_i(\theta; \mathcal{D}_i)$ around $\theta_i^*$:
\begin{equation}
\begin{split}
    \ell_i(\theta_j^*; \mathcal{D}_i) 
    &\approx \ell_i(\theta_i^*; \mathcal{D}_i) + (\theta_j^* - \theta_i^*)^\top \nabla_{\theta} \ell_i(\theta; \mathcal{D}_i) \big|_{\theta = \theta_i^*} \\
    & + \frac{1}{2}(\theta_j^* - \theta_i^*)^\top \nabla_{\theta}^2 \ell_i(\theta; \mathcal{D}_i) \big|_{\theta = \theta_i^*} (\theta_j^* - \theta_i^*) \\
    &\approx \ell_i(\theta_i^*; \mathcal{D}_i) + \frac{1}{2} \Delta \theta^{\top} \nabla_{\theta}^2 \ell_i(\theta; \mathcal{D}_i) \big|_{\theta = \theta_i^*} \Delta \theta,
\end{split}
\label{eqn_compatibility_error}
\end{equation}
where $\Delta \theta := \theta_j^* - \theta_i^*$ and $\nabla_{\theta} \ell_i(\theta; \mathcal{D}_i) \big|_{\theta = \theta_i^*} \approx \textbf{0}$. Then, the performance degradation of task $i$ is upper-bounded by
\begin{equation}
    \ell_i(\theta_j^*; \mathcal{D}_i) - \ell_i(\theta_i^*; \mathcal{D}_i) \leq \frac{1}{2} \lambda_i^{max} \| \Delta \theta \|^2,
\label{eqn_compatibility_bound}
\end{equation}
where $\lambda_i^{max}$ is the maximum eigenvalue of the Hessian matrix $\nabla_{\theta}^2 \ell_i(\theta; \mathcal{D}_i) \big|_{\theta = \theta_i^*}$. Note that the order of task $i$ and $j$ can be arbitrary, that is, Eq.~\eqref{eqn_compatibility_bound} demonstrates both forward and backward effects.
%the effects of both forward and backward adaptation. 
Therefore, the robustness of an empirical optimal solution $\theta_i^*$ to parameter changes is closely related to $\lambda_i^{max}$, which has been a common metric to describe the \textbf{flatness of loss landscape}~\cite{mirzadeh2020understanding,hochreiter1997flat,keskar2017large}. 

Intuitively, convergence to a local minima with a flatter loss landscape will be less sensitive to modest parameter changes and thus benefit both old and new tasks (see Fig.~\ref{fig:Theory}, c). To find such a \emph{flat minima}, 
there are three widely-used strategies in continual learning. 
% there are three main strategies that have been adopted in continual learning. 
The first is derived from its definition, i.e., the flatness metric. Specifically, the minimization of $\ell_t(\theta; \mathcal{D}_t)$ can be replaced by a robust task-specific loss $\ell_t^{b}(\theta; \mathcal{D}_t) := \max_{\| \delta \| \leq b} \ell_t(\theta+\delta; \mathcal{D}_t)$, thus the obtained solution guarantees low error not only at a specific point but also in its neighborhood with a ``radius'' of $b$. However, due to the great dimensionality of $\theta$, the calculation of $\ell_t^{b}(\theta; \mathcal{D}_t)$ cannot cover all possible $\delta$ but only a few directions~\cite{shi2021overcoming}, similar to the complexity issue of computing the Hessian matrix in Eq.~\eqref{eqn_compatibility_error}. %have to collect only a few directions
The alternatives include using an approximation of the Hessian~\cite{mirzadeh2020understanding,deng2021flattening} or calculating $\delta$ only along the trajectory of forward and backward parameter changes~\cite{huang2021altersgd,mirzadeh2020linear,cha2021swad,izmailov2018averaging,mehta2021empirical}. 
The second is to operate the loss landscape by constructing an ensemble model under the restriction of mode connectivity, i.e., integrating multiple minima in parameter or function space along the low-error path, since connecting them ensures flatness on that path~\cite{mirzadeh2020linear,wang2022coscl,izmailov2018averaging,garipov2018loss,wortsman2022model,cha2020cpr}. 
These two strategies are closely related to the \emph{optimization-based approach}.
The third comes down to obtaining well-distributed representations, which tend to be more robust to distribution differences in function space, such as by using pre-training~\cite{ramasesh2021effect,mehta2021empirical,hu2021well}, wider network architectures~\cite{ramasesh2020anatomy,mirzadeh2022wide,ramasesh2021effect} and self-supervised learning~\cite{hu2021well,pham2021dualnet,cha2021co2l,madaan2021representational}. 
Observing the substantial attention to large-scale pre-training and self-supervised learning, we group this direction into the \emph{representation-based approach} in Sec.~\ref{Sec.4.4_Representation}.

There are many other factors that are important for continual learning performance. As shown in Eq.~\eqref{eqn_compatibility_bound}, the upper bound of performance degradation also depends on the difference of the empirical optimal solution $\theta_t^{*} = {\arg \min}_{\theta} \ell_t(\theta; \mathcal{D}_t)$ for each task, i.e., the \textbf{discrepancy of task distribution} (see Fig.~\ref{fig:Theory}, d), which is further validated by a theoretical analysis of the forgetting-generalization trade-off~\cite{ramakrishnan2020relationship} and the PAC-Bayes bound of generalization errors~\cite{pentina2014pac,wang2022coscl}. Therefore, how to exploit task similarity is directly related to the performance of continual learning. 
The generalization error for each task can improve with synergistic tasks but deteriorate with competing tasks~\cite{ramesh2021model}, where learning all tasks equally in a shared solution tends to compromise each task in performance~\cite{ramesh2021model,wang2022coscl}. 

On the other hand, when model parameters are \emph{not} shared by all tasks (e.g., using a multi-head output layer), the impact of task similarity on continual learning will be complex. Some theoretical studies with the neural tangent kernel (NTK)~\cite{bennani2020generalisation,doan2021theoretical,karakida2022learning,lee2021continual} suggest that an increase in task similarity may lead to more forgetting. Since the output heads are independent for individual tasks, it becomes much more difficult to distinguish between two similar solutions~\cite{karakida2022learning,lee2021continual}.
Specifically, under the NTK regime from the $t$-th task up until the $k$-th task, the forgetting of old tasks is bounded by:
%\vspace{-0.4cm}
\begin{small}
\begin{equation}
\begin{split}
%&\left \| p(\mathcal{D}_t|\theta_k^{*})-p(\mathcal{D}_t|\theta_t^{*}) \right \|_F^2 \leq \sigma_{t,1}^2 \sum_{i=t+1}^{k} \left \| \Theta^{t\rightarrow i} \right \| _2^2 \left \| {\rm{RES}_i }\right \| _2^2, \nonumber,\\
&\left \| p(\mathcal{D}_k|\theta_k^{*})-p(\mathcal{D}_k|\theta_t^{*}) \right \|_F^2 \leq \\ %\nonumber
&\sigma_{t,\rm{|rep|+1}}^2 \sum_{i=t+1}^{k} \left \| \Theta^{t\rightarrow S(i,\rm{|rep|})} \right \| _2^2\left \| \Theta^{i\rightarrow S(i,\rm{|rep|})} \right \| _2^2 \left \| {\rm{RES}_i }\right \| _2^2. %\nonumber
\end{split}
\end{equation}
\end{small}
$\Theta^{t\rightarrow k}$ is a diagonal matrix where each diagonal element $\cos(\theta_{t,k})_r$ is the cosine of the $r$-th principal angle between the $t$-th and $k$-th tasks in the feature space. $\sigma_{t,\cdot}$ is the $\cdot$-th singular value of the $t$-th task. ${\rm{RES}}_i$ is the rotated residuals that remain to be learned, and $S(i,\cdot)$ represents the residuals subspace of order $\cdot$ until the $i$-th task. $\rm{|rep|}$ is the sample number of replay data. 
The complex impact of task similarity suggests the importance of model architectures for coordinating task-sharing and task-specific components.
%These results can inspire the design of model architecture for continual learning, with coordinating task-sharing and task-specific components.

%Many theoretical studies~\cite{bennani2020generalisation,doan2021theoretical,karakida2022learning,lee2021continual} have observed that for fully-connected networks with a multi-head output layer, an increase in task similarity may lead to more forgetting. Since the output heads are independent for individual tasks, it becomes much more difficult to distinguish between two similar solutions~\cite{karakida2022learning,lee2021continual}. The impact of task similarity can inspire the design of model architecture for continual learning, in terms of task-sharing and task-specific components.
%The above analysis could be generalized to the design of model architecture for continual learning, where the task-sharing and task-specific components have sophisticated interactions with task similarity.
%To facilitate knowledge transfer and prevent inter-task interference, a promising idea is to combine multiple continual learning models with a proper diversity of task expertise~\cite{ramesh2021model,wang2022coscl,doan2022efficient}, as an emerging direction of the \emph{architecture-based approach}.

Moreover, the complexity of finding a desirable solution for continual learning is determined to a large extent by the \textbf{structure of parameter space}. % corresponding to the model architecture. 
Learning all incremental tasks with a shared solution is equivalent to learning each new task in a constrained parameter space that prevents performance degradation of all old tasks. Such a classical continual learning problem has proven to be NP-hard in general~\cite{knoblauch2020optimal}, because the feasible parameter space tends to be narrow and irregular as more tasks are introduced, thus difficult to identify (see Fig.~\ref{fig:Theory}, e). This challenging issue can be mitigated by replaying representative old training samples~\cite{knoblauch2020optimal}, restricting the feasible parameter space to a hyperrectangle~\cite{wolczyk2022continual}, or alternating the model architecture of using a single parameter space (e.g., using multiple continual learning models)~\cite{ramesh2021model,wang2022coscl,doan2022efficient}. % corresponding to the model architecture.

To harmonize the important factors in continual learning, recent work presents a similar form of generalized bounds for learning and forgetting.
%recent work presents generalization bounds (especially learning and forgetting) in a similar form.
For example, with probability at least $1-\delta$ for any $\delta \in (0,1)$, an ideal continual learner~\cite{peng2023ideal} under the assumption that all tasks share a global minimizer with uniform convergence, i.e., $\lambda_i^{max} =\lambda$ for $\forall t=1,\cdots, k $ in Eq.~\eqref{eqn_compatibility_bound}, has the generalization bound
%Specifically, let $\delta \in (0,1)$ and $c_t^*= \ell_t(\theta_t^*;\mathcal{D}_t)$ be the minimum of the $t$-th task loss. Under the assumption that all tasks share a global minimizer with uniform convergence (i.e., $\lambda_i^{max} =\lambda$ for $\forall t=1,\cdots, k $), then with probability at least $1-\delta$, the ideal continual learner~\cite{peng2023ideal} has the generalization bound
\begin{equation}
\begin{split}
   c_t^* \leq  \mathbb{E}_{\mathcal{D}_t\sim\mathbb{D}_t} \ell_t(\theta;\mathcal{D}_t) \leq c_t^* + \zeta(N_t, \delta), \forall t=1,\cdots, k, %\nonumber
\label{eqn_icl_bound}
\end{split}
\end{equation}
where $c_t^*= \ell_t(\theta_t^*;\mathcal{D}_t)$ is the minimum loss of the $t$-th task, and $\theta$ is a global solution of the continually learned $1:k$ tasks by empirical risk minimization.
% , i.e., $(i.e., \ell_t(\theta; \mathcal{D}_t))$. $\zeta$ is the upper bound of convergence.
$\zeta=O(\frac{\lambda B\sqrt{|\theta|\log(N_t)\log(|\theta|k/\delta)}}{2\sqrt{N_t}} )$, and $\left \| \theta \right \| _2\leq B$.
Considering that the shared parameter space for many different tasks might be an empty set (see Fig.~\ref{fig:Theory}, e), i.e., $ \cup_{t=1}^{k} \theta_t=\emptyset$, the generalization bounds are further refined by assuming $K$ parameter spaces ($K\geq 1$ in general) to capture all tasks~\cite{wang2022coscl,wang2023incorporating}. For generalization errors of new and old tasks:

\vspace{-0.2cm}
\begin{footnotesize}
\begin{equation}
\begin{split}
    \mathbb{E}_{\mathcal{D}_t\sim\mathbb{D}_t} \ell_t(\theta;\mathcal{D}_t) \leq c_t^* + R(\sum_{i=1}^{t-1}\ell_i^b) +\sum_{i=1}^{t-1} \rm{Div}( \mathbb{D}_\textit{i}, \mathbb{D}_\textit{t}) +\zeta(\sum_{i=1}^{t-1} N_i, K/\delta), \\
    \sum_{i=1}^{t-1} \mathbb{E}_{\mathcal{D}_i\sim\mathbb{D}_i} \ell_i(\theta;\mathcal{D}_i)\leq \sum_{i=1}^{t-1} c_i^* + R(\ell_t^b) + \sum_{i=1}^{t-1} \rm{Div}( \mathbb{D}_\textit{t}, \mathbb{D}_\textit{i}) + \zeta( N_t, K/\delta), 
\end{split}
\end{equation}
\end{footnotesize}
where $R(\cdot)$ and $\rm{Div}$ are the functions of loss flatness and task discrepancy, respectively. The definitions of $\delta$, $\theta$ and $c_t^*$ are the same as Eq.~\eqref{eqn_icl_bound}.

These theoretical efforts suggest that, a desirable solution for continual learning should provide an appropriate stability-plasticity trade-off and an adequate intra/inter-task generalizability, motivating a variety of representative methods as detailed in the next section.

\section{Method}\label{Sec.4_Method}
In this section, we present an elaborated taxonomy of representative continual learning methods (see Fig.~\ref{fig:Method} and also Fig.~\ref{fig:CL_Framework}, c), analyzing extensively their main motivations, typical implementations and empirical properties.

%\hangx{we may keep the milestone papers while remove the others. Also, add some comments about the trends and limitations of current methods.}
%\hangx{when we read a survey, one may expect the fundamental venation and future opportunity.}
%\hangx{\textbf{it is imperative if we can identify some critical perspectives, and make a comprehensive comparison between different strategies  }}

\subsection{Regularization-Based Approach}\label{Sec.4.1_Regularization}

\begin{figure}[th]
	\centering
	\includegraphics[width=0.88\columnwidth]{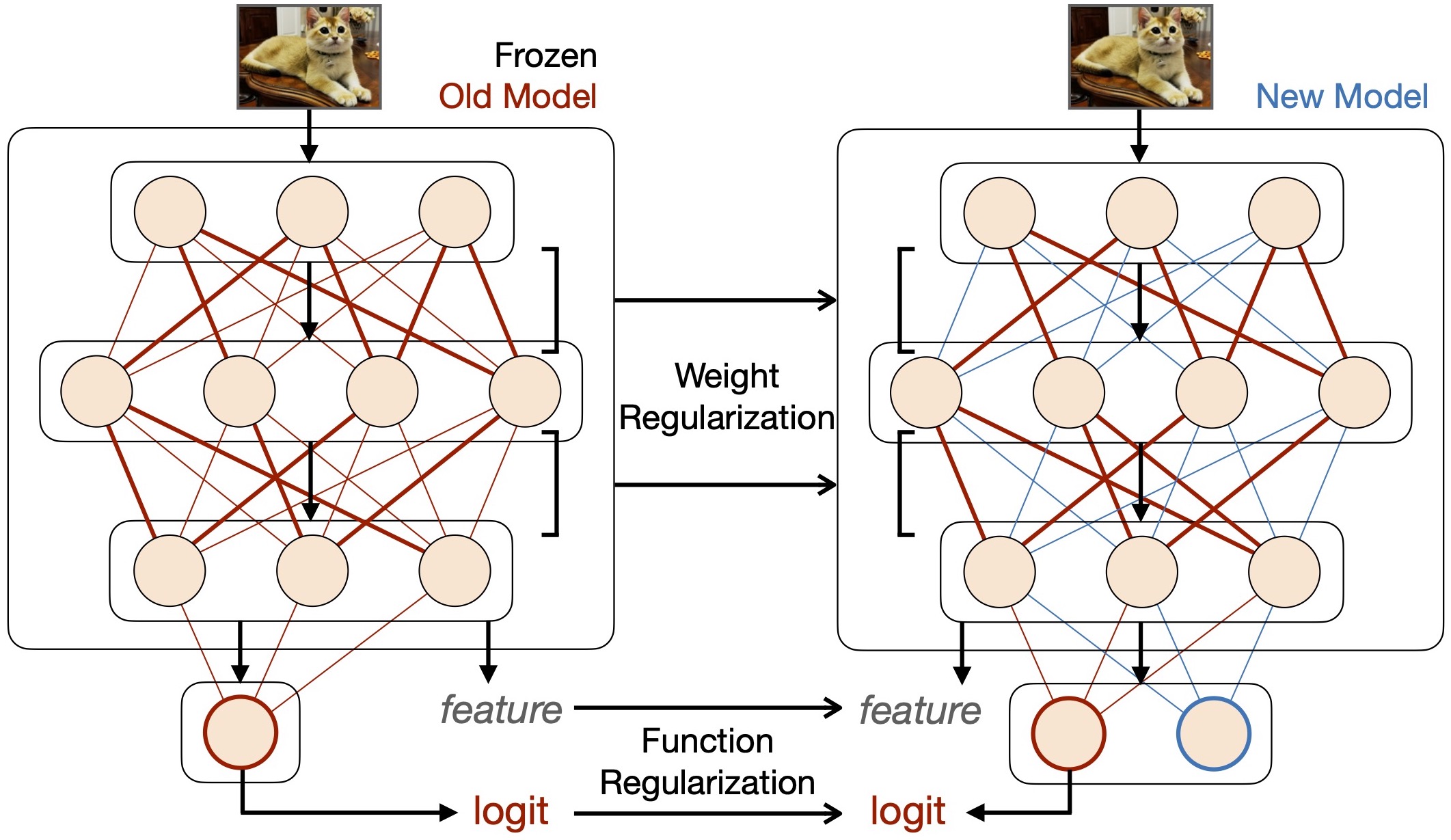} 
    \vspace{-.1cm}
	\caption{Regularization-based approach. This direction is characterized by adding explicit regularization terms to mimic the parameters (weight regularization) or behaviors (function regularization) of the old model.}
	\label{fig:Regularization}
    \vspace{-.1cm}
\end{figure}

This direction is characterized by adding explicit regularization terms to balance the old and new tasks, which usually requires storing a frozen copy of the old model for reference (see Fig.~\ref{fig:Regularization}). Depending on the target of regularization, such methods can be divided into two sub-directions.

The first is \textbf{weight regularization}, which selectively regularizes the variation of network parameters. A typical implementation is to add a quadratic penalty in loss function that penalizes the variation of each parameter depending on its contribution or ``importance'' to performing the old tasks, e.g., Eq.~\eqref{eqn_EWC}, in a form originally derived from online Laplace approximation of the posterior under the Bayesian framework. The importance can be calculated by the Fisher information matrix (FIM), such as EWC~\cite{kirkpatrick2017overcoming} and some more advanced versions~\cite{ritter2018online,schwarz2018progress}.
%with a diagonal approximation, i.e., EWC~\cite{kirkpatrick2017overcoming}, or a Kronecker-factored approximation~\cite{ritter2018online}. 
%Online EWC~\cite{schwarz2018progress} extended the original version of EWC by recursively updating the FIM without accessing the task label. 
Meanwhile, numerous efforts have been devoted to designing a better importance measurement. SI~\cite{zenke2017continual_si} online approximates the importance of each parameter by its contribution to the total loss variation and its update length over the entire training trajectory. MAS~\cite{aljundi2018memory_mas} accumulates the importance measurements based on the sensitivity of predictive results to parameter changes, which is both online and unsupervised. RWalk~\cite{chaudhry2018riemannian_rwalk} combines the regularization terms of SI~\cite{zenke2017continual_si} and EWC~\cite{kirkpatrick2017overcoming} to integrate their advantages. Interestingly, these importance measurements have been shown to be all tantamount to an approximation of the FIM~\cite{benzing2022unifying}, although stemming from different motivations. 

There are also several works refining the implementation of the quadratic penalty.
Since the diagonal approximation of the FIM in EWC might lose information about the old tasks, R-EWC~\cite{liu2018rotate} performs a factorized rotation of the parameter space to diagonalize the FIM. XK-FAC~\cite{lee2020continual} further considers the inter-example relations in approximating the FIM to better accommodate batch normalization. Observing the asymmetric effect of parameter changes on old tasks, ALASSO~\cite{park2019continual} designs an asymmetric quadratic penalty with one of its sides overestimated. 

Compared to learning each task within the constraints of the old model, which typically exacerbates the intransience, an \emph{expansion-renormalization} process of obtaining separately the new task solution and renormalizing it with the old model has been shown to provide a better stability-plasticity trade-off~\cite{lee2017overcoming_imm,lee2020residual,schwarz2018progress,wang2021afec,lin2022towards}. IMM~\cite{lee2017overcoming_imm} is an early attempt that incrementally matches the moment of the posterior distributions of old and new tasks, i.e., a weighted average of their solutions. % with transfer learning techniques such as weight transfer, L2-norm and dropout. 
ResCL~\cite{lee2020residual} extends this idea with a learnable combination coefficient.
P\&C~\cite{schwarz2018progress} learns each task individually with an additional network, and then distills it back to the old model with a generalized version of EWC. AFEC~\cite{wang2021afec} introduces a forgetting rate to eliminate the potential negative transfer from the original posterior $p(\theta | \mathcal{D}_{1:k-1})$ in Eq.~\eqref{eqn_cl_posterior}, and derives quadratic terms to penalize differences of the network parameters $\theta$ with both the old and new task solutions. To reliably average the old and new task solutions, a linear connector~\cite{lin2022towards} is constructed by constraining them on a linear low-error path.

Other forms of regularization that target the network itself also belong to this sub-direction. As discussed before, online VI of the posterior distribution can serve as an implicit regularization of parameter changes, such as VCL~\cite{nguyen2018variational,swaroop2019improving}, NVCL~\cite{tserannatural}, CLAW~\cite{adel2019continual}, GVCL~\cite{loo2020generalized}, KCL~\cite{derakhshani2021kernel} and VAR-GPs~\cite{kapoor2021variational}. Instead of consolidating parameters, NPC~\cite{paik2020overcoming} estimates the importance of each neuron and selectively reduces its learning rate. UCL~\cite{ahn2019uncertainty} and AGS-CL~\cite{jung2020continual} freeze
% \junz{freeze?} 
the parameters connecting the important neurons, equivalent to a hard version of weight regularization.

The second is \textbf{function regularization}, which targets the intermediate or final output of the prediction function.
%and is similar to the idea of knowledge distillation~\cite{gou2021knowledge}. 
This strategy typically employs the previously-learned model as the teacher and the currently-trained model as the student, while implementing knowledge distillation (KD)~\cite{gou2021knowledge} to mitigate catastrophic forgetting.
Ideally, the target of KD should be all old training samples, which are unavailable in continual learning. The alternatives can be new training samples~\cite{li2017learning,dhar2019learning,iscen2020memory,rannen2017encoder}, a small fraction of old training samples~\cite{rebuffi2017icarl,castro2018end,hou2019learning_lucir,douillard2020podnet}, external unlabeled data~\cite{lee2019overcoming}, generated data~\cite{wu2018memory_mrgan,zhai2019lifelong_gan}, etc., suffering from different degrees of distribution shift.

% (a modified version in terms of the output head)
As a pioneer work, LwF~\cite{li2017learning} and LwF.MC~\cite{rebuffi2017icarl} learn \emph{new training samples} while using their predictions from the output head of the old tasks to compute the distillation loss. %of their predictions from the old task output head. 
LwM~\cite{dhar2019learning} exploits the attention maps of new training samples for KD. %to transfer knowledge between old and new models.
EBLL~\cite{rannen2017encoder} learns task-specific autoencoders and prevents changes in feature reconstruction. % for new training samples.
GD~\cite{lee2019overcoming} further distills knowledge from massive \emph{unlabeled data} available in the wild. %, together with new training samples. %on the large stream of \emph{unlabeled data} 
When old training samples are faithfully recovered, the potential of function regularization can be largely released.
%The potential of function regularization is largely released when old training samples can be (partially) recovered. 
Thus, function regularization often collaborates with replaying a few \emph{old training samples}, such as iCaRL~\cite{rebuffi2017icarl}, EEIL~\cite{castro2018end}, LUCIR~\cite{hou2019learning_lucir}, PODNet~\cite{douillard2020podnet}, DER~\cite{buzzega2020dark}, etc., discussed latter in Sec.~\ref{Sec.4.2_Replay}. Besides, sequential Bayesian inference over function space can be seen as a form of function regularization, which generally requires storing some old training samples (called ``coreset'' in literature), such as FRCL~\cite{titsias2019functional}, FROMP~\cite{pan2020continual} and S-FSVI~\cite{rudner2022continual}. 
For conditional generation, the \emph{generated data} of previously-learned conditions and their output values are regularized to be consistent between the old and new models, such as MeRGANs~\cite{wu2018memory_mrgan}, DRI~\cite{wang2022continual_dri} and LifelongGAN~\cite{zhai2019lifelong_gan}.
%\hangx{comments on the limitations, and the pros. and cons. compared with alternative methods. }

%A significant limitation of regularization-based approaches is that the old model can only preserve the distribution learned from its objective, i.e., p(y|x) for discriminative model. A full distribution p(x,y) is generally required to perform class-incremental learning. (which can be achieved by replay). Hybrid regularization and replay can benefit each other: replay brings full distribution, (functional) regularization brings additional information of the old tasks. Weight or functional regularization can be flexibly combine with other methods and adapt to a broad range of scenarios.

\subsection{Replay-Based Approach}\label{Sec.4.2_Replay}
\begin{figure}[th]
	\centering
	\includegraphics[width=1\columnwidth]{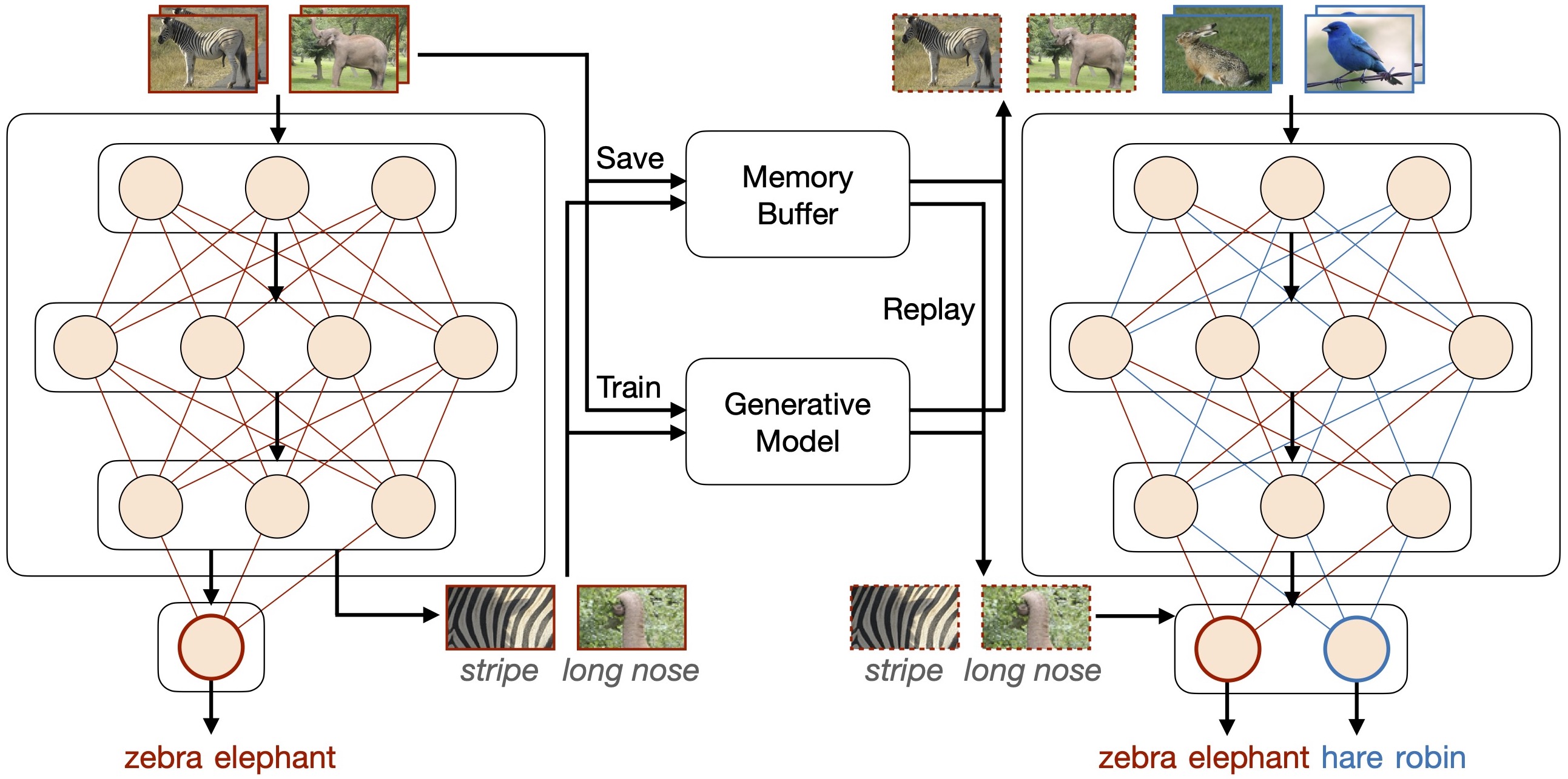} 
     \vspace{-.5cm}
	\caption{Replay-based approach. This direction is characterized by approximating and recovering the old data distributions. Typical sub-directions include experience replay, which saves a few old training samples in a memory buffer; generative replay, which trains a generative model to provide generated samples; and feature replay, which recovers the distribution of old features through saving prototypes, saving statistical information or training a generative model.}
	\label{fig:Replay}
     \vspace{-.45cm}
\end{figure}

We group the methods for approximating and recovering old data distributions into this category (see Fig.~\ref{fig:Replay}). Depending on the content of replay, these methods can be further divided into three sub-directions, each with its own targets and challenges.

The first is \textbf{experience replay}, which typically stores a few old training samples within a small memory buffer. Due to the extremely limited storage space, the key challenges consist of \emph{how to construct} and \emph{how to exploit} the memory buffer. As for construction, the preserved old training samples should be carefully selected, compressed, augmented, and updated, in order to recover adaptively the past information. Earlier work adopts fixed principles for \emph{sample selection}. For example, Reservoir Sampling~\cite{vitter1985random,chaudhry2019tiny,riemer2018learning} randomly preserves a fixed number of old training samples obtained from each training batch. Ring Buffer~\cite{lopez2017gradient_gem} further ensures an equal number of old training samples per class. %further ensured that an equal number of old training samples are randomly selected for each class.
Mean-of-Feature~\cite{rebuffi2017icarl} selects an equal number of old training samples that are closest to the feature mean of each class. There are many other fixed principles, such as k-means~\cite{chaudhry2019tiny}, plane distance~\cite{riemer2018learning} and entropy~\cite{riemer2018learning}, but all perform mediocrely~\cite{riemer2018learning,chaudhry2019tiny}. More advanced strategies are typically gradient-based or optimizable, by maximizing such as 
the sample diversity in terms of parameter gradients (GSS~\cite{aljundi2019gradient}), performance of corresponding tasks with cardinality constraints (CCBO~\cite{borsos2020coresets}), mini-batch gradient similarity and cross-batch gradient diversity (OCS~\cite{yoon2021online}), ability of optimizing latent decision boundaries (ASER~\cite{shim2021online}), diversity of robustness against perturbations (RM~\cite{bang2021rainbow}), similarity to the gradients of old training samples with respect to the current parameters (GCR~\cite{tiwari2022gcr}), etc. 

To improve \emph{storage efficiency}, AQM~\cite{caccia2020online} performs online continual compression based on a VQ-VAE framework~\cite{van2017neural} and saves compressed data for replay. MRDC~\cite{wang2021memory} formulates experience replay with data compression as determinantal point processes (DPPs)~\cite{kulesza2012determinantal}, and derives a computationally efficient way for online determination of an appropriate compression rate. RM~\cite{bang2021rainbow} adopts conventional and label mixing-based strategies of data augmentation to enhance the diversity of old training samples. RAR~\cite{kumariretrospective} synthesizes adversarial samples near the forgetting boundary and performs MixUp~\cite{zhang2018mixup} for data augmentation. 
The auxiliary information with low storage cost, such as class statistics (IL2M~\cite{belouadah2019il2m}, SNCL~\cite{gong2022continual}) and attention maps (RRR~\cite{ebrahimi2020remembering}, EPR~\cite{saha2020gradient}), can be further incorporated to maintain old knowledge. Besides, the old training samples could be continually modified to accommodate incremental changes, such as by making them more representative (Mnemonics~\cite{liu2020mnemonics}) or more challenging (GMED~\cite{jin2021gradient}) for separation.

%GSS~\cite{aljundi2019gradient} maximized sample diversity in terms of parameter gradients. OCS~\cite{yoon2021online} maximized mini-batch gradient similarity and cross-batch diversity. GCR~\cite{tiwari2022gcr} selected the samples that closely approximates the gradient of previous data with respect to current model parameters. ASER~\cite{shim2021online} scored the samples according to their ability to optimize latent decision boundaries.

As for exploitation, experience replay requires an adequate use of the memory buffer to recover the past information. There are many different implementations, closely related to other continual learning strategies.
%There are many different implementations, mainly through optimization, distillation, etc., closely connected to other representative continual learning approaches. 
First, the effect of old training samples in \emph{optimization} can be constrained to avoid catastrophic forgetting and facilitate knowledge transfer. For example, GEM~\cite{lopez2017gradient_gem} constructs individual constraints based on the old training samples for each task to ensure non-increase in their losses.
%constraints for the old training samples of each task to ensure their losses non-increase. 
A-GEM~\cite{chaudhry2018efficient_agem} replaces the individual constraints with a global loss of all tasks to improve training efficiency.
%further improves training efficiency by constraining the global loss of all tasks.
LOGD~\cite{tang2021layerwise} decomposes the gradient of each task into task-sharing and task-specific components to leverage inter-task information. To achieve a good trade-off in interference-transfer~\cite{riemer2018learning} (i.e., forgetting-generalization~\cite{ramakrishnan2020relationship}), MER~\cite{riemer2018learning} employs meta-learning for gradient alignment in experience replay. BCL~\cite{ramakrishnan2020relationship} explicitly pursues a saddle point of the cost of old and new training samples. MetaSP~\cite{sun2022exploring} leverages the Pareto optimum of example influence on stability-plasticity to control the model and storage updates.
To selectively utilize the memory buffer, MIR~\cite{aljundi2019mir} prioritizes the old training samples that subject to more forgetting, while HAL~\cite{chaudhry2021using} uses them as ``anchors'' and stabilizes their predictions. 

On the other hand, experience replay can be naturally combined with \emph{knowledge distillation} (KD), which additionally incorporates the past information from the old model. iCaRL~\cite{rebuffi2017icarl} and EEIL~\cite{castro2018end} are two early works that perform KD on both old and new training samples. Some subsequent improvements focus on different issues in experience replay. %D+R~\cite{hou2018lifelong} performs KD from an additional model dedicated to the current task to enhance learning plasticity. 
To mitigate data imbalance of the limited old training samples, LUCIR~\cite{hou2019learning_lucir} encourages similar feature orientation of the old and new models, while performing cosine normalization of the last layer and mining hard negatives of the current task. BiC~\cite{wu2019large_bic} and WA~\cite{zhao2020maintaining} attribute this issue to the bias of the last fully connected layer, and resolve it by either learning a bias correction layer with a balanced validation set~\cite{wu2019large_bic} or normalizing the output weights~\cite{zhao2020maintaining}. SS-IL~\cite{ahn2021ss} adopts separated softmax in the last layer and task-wise KD to mitigate the bias.
DRI~\cite{wang2022continual_dri} trains a generative model to supplement the old training samples with additional generated data.
To alleviate dramatic representation shifts, PODNet~\cite{douillard2020podnet} employs a spatial distillation loss to preserve representations throughout the model. Co2L~\cite{cha2021co2l} introduces a self-supervised distillation loss to obtain robust representations against catastrophic forgetting. % based on contrastive learning
GeoDL~\cite{simon2021learning} performs KD along a path that connects the low-dimensional projections of the old and new feature spaces for a smooth transition between them.
ELI~\cite{joseph2022energy} learns an energy manifold with the old and new models to realign the representation shifts for optimizing incremental tasks.
To adequately exploit the past information, AU~\cite{kurmi2021not} incorporates uncertainty and self-attention into the distillation loss, while CSC~\cite{ashok2022class} additionally leverages the structure of the feature space.
DDE~\cite{hu2021distilling} distills colliding effects from the features of the new training samples, which is causally equivalent to replaying more old training samples.
TAMiL~\cite{bhat2023task} adds task-specific attention in the feature space and performs consistency regularization to better preserve task-relevant information.
To further enhance learning plasticity, D+R~\cite{hou2018lifelong} performs KD from an additional model dedicated to each new task. 
FOSTER~\cite{wang2022foster} dynamically expands new modules to fit the residuals of the old model and then distills them into a single model.
Besides, weight regularization approaches can be combined with experience replay to achieve better performance and generality~\cite{chaudhry2018riemannian_rwalk,wang2021afec}.

%Besides, there are many other explorations. To selectively exploit the memory buffer, MIR~\cite{aljundi2019mir} prioritized the old training samples that tend to suffer more forgetting, while HAL~\cite{chaudhry2021using} used them as ``anchors'' and stabilized their predictions. Representative weight regularization approaches, such as EWC~\cite{kirkpatrick2017overcoming}, MAS~\cite{aljundi2018memory_mas}, SI~\cite{zenke2017continual_si}, RWalk~\cite{chaudhry2018riemannian_rwalk} and AFEC~\cite{wang2021afec}, can also be combined with experience replay to obtain better performance and generality. 

It is worth noting that the merits and potential limitations of experience replay remain largely open. In addition to the intuitive benefits of staying in the low-loss region of the old tasks~\cite{verwimp2021rehearsal}, theoretical analysis has demonstrated its contribution to resolving the NP-hard problem of optimal continual learning~\cite{knoblauch2020optimal}. However, it risks overfitting to only a few old training samples retained in the memory buffer, which potentially affects generalizability~\cite{verwimp2021rehearsal}. 
To alleviate overfitting, LiDER~\cite{bonicelli2022effectiveness} takes inspirations from adversarial robustness and enforces the Lipschitz continuity of the model to its inputs.
MOCA~\cite{yu2022continual} enlarges the variation of representations to prevent the old ones from shrinking in their space.
Interestingly, several simple baselines of experience replay can achieve considerable performance. DER/DER++~\cite{buzzega2020dark} and X-DER~\cite{boschini2022class} preserve old training samples together with their logits, and perform logit-matching throughout the optimization trajectory. GDumb~\cite{prabhu2020gdumb} greedily collects incoming training samples in a memory buffer and then uses them to train a model from scratch for testing. These efforts can serve as an evaluation criterion for more advanced strategies in this sub-direction.

The second is \textbf{generative replay} or pseudo-rehearsal, which usually requires training an additional generative model to replay generated data. This is closely related to continual learning of generative models themselves, as they also require incremental updates. 
%This sub-direction is closely related to continual learning of generative models themselves, as they also need to be updated sequentially. 
%, where representative continual learning approaches can be incorporated.
DGR~\cite{shin2017continual_dgr} provides an initial framework in which learning each generation task is accompanied with replaying generated data sampled from the old generative model, so as to inherit the previously-learned knowledge. MeRGAN~\cite{wu2018memory_mrgan} further employs replay alignment to enforce consistency of the generative data sampled with the same random noise between the old and new generative models, similar to the role of function regularization. %, which can be seen as a form of function regularization. 
Besides, other continual learning strategies can be incorporated into generative replay. 
Weight regularization~\cite{seff2017continual,nguyen2018variational,wang2021triple,wang2021ordisco} and experience replay~\cite{he2018exemplar,wang2021ordisco} have been shown to be effective in mitigating catastrophic forgetting of generative models. 
DGMa/DGMw~\cite{ostapenko2019learning_dgmw} adopts binary masks to allocate task-specific parameters for overcoming inter-task interference, and an extendable network to ensure scalability. 
If pre-training is available, it can provide a relatively stable and strong reference model for continual learning. For example, FearNet~\cite{kemker2018fearnet} and ILCAN~\cite{xiang2019incremental} additionally preserves statistical information of the old features acquired from a pre-trained feature extractor, while GAN-Memory~\cite{cong2020gan} continually adjusts a pre-trained generative model with task-specific parameters.
%trains additional task-specific parameters to adptively leverage a pre-trained generative model. 

The generative models for pseudo-rehearsal can be of various types, such as generative adversarial networks (GANs) and (variational) autoencoder (VAE).
%Generative replay can be based on various types of generative models, such as GANs and autoencoders. 
A majority of state-of-the-art approaches have focused on GANs to enjoy its advantages in fine-grained generation, but usually suffer from label inconsistency in continual learning~\cite{ostapenko2019learning_dgmw,ayub2020eec}. In contrast, autoencoder-based strategies, such as FearNet~\cite{kemker2018fearnet}, SRM~\cite{riemer2019scalable}, CLEER~\cite{rostami2019complementary}, EEC~\cite{ayub2020eec}, GMR~\cite{pfulb2021continual} and Flashcards~\cite{gopalakrishnan2022knowledge}, can explicitly control the labels of the generated data, albeit with relatively blurred granularity. L-VAEGAN~\cite{ye2020learning} instead employs a hybrid model for both high-quality generation and accurate inference. 
However, since continual learning of generative models is extremely difficult and requires significant resource overhead, generative replay is often limited to relatively simple datasets~\cite{van2020brain,wang2021triple}. An alternative is to convert the target of generative replay from data level to feature level, which can largely reduce the complexity of conditional generation and more adequately exploit semantic information. For example, GFR~\cite{liu2020generative} trains conditional GANs to replay generated features after the feature extractor. BI-R~\cite{van2020brain} incorporates context-modulated feedback connections in a standard VAE to replay internal representations.

%If pre-training is available, continual learning benefits from pre-training by obtaining relatively stable and separable representations. For example, 
%Storing extra old training samples such as ESGR~\cite{he2018exemplar}, while DRI used generative model for data augmentation and mitigate data imbalance.

In fact, maintaining feature-level rather than data-level distributions enjoys numerous benefits in terms of efficiency and privacy. We refer to this sub-direction as \textbf{feature replay}. However, a central challenge is the \emph{representation shift} caused by sequentially updating the feature extractor, which reflects the feature-level catastrophic forgetting. To address this issue, GFR~\cite{liu2020generative}, FA~\cite{iscen2020memory} and DSR~\cite{zhu2022self} perform feature distillation between the old and new models. IL2M~\cite{belouadah2019il2m} and SNCL~\cite{gong2022continual} recover statistics of feature representations (e.g., mean and covariance) on the basis of experience replay. 
RER~\cite{toldo2022bring} explicitly estimates the representation shift to update the preserved old features.
REMIND~\cite{hayes2020remind} and ACAE-REMIND~\cite{wang2021acae} instead fix the early layers of the feature extractor and reconstruct the intermediate representations to update the latter layers. 
FeTrIL~\cite{petit2022fetril} employs a fixed feature extractor learned from the initial task and replays generated features afterwards.

For continual learning from scratch, the required changes in representation are often dramatic, while stabilizing the feature extractor may interfere with accommodating new representations. 
In contrast, the use of strong pre-training can provide robust representations that are generalizable to downstream tasks and remain relatively stable in continual learning. 
An empirical study~\cite{ostapenko2022foundational} has systematically investigated feature replay for continual learning with large-scale pre-training. A more in-depth discussion of this topic is presented in Sec.~\ref{Sec.4.4_Representation}.
%Recent prompt-based approaches such as L2P~\cite{wang2022learning_l2p}, DualPrompt~\cite{wang2022dualprompt} and S-Prompts~\cite{wang2022sprompts} adopt a small set of learnable parameters (i.e., prompts) to instruct the representations of a pre-trained transformer and preserve such parameters in a memory space, which can be seen as an extension of this strategy.

\subsection{Optimization-Based Approach}\label{Sec.4.3_Optimization}
Continual learning can be achieved by not only adding additional terms to the loss function (e.g., regularization and replay), but also explicitly designing and manipulating the optimization programs.

%This direction is characterized by the design of optimization programs to prevent interference between old and new tasks. 
% the design of optimization algorithms
%an explicit manipulation of the optimization process to prevent interference between old and new tasks. 
A typical idea is to perform \textbf{gradient projection}. 
%The previously mentioned GEM~\cite{lopez2017gradient_gem}, A-GEM~\cite{chaudhry2018efficient_agem}, LOGD~\cite{tang2021layerwise} and MER~\cite{riemer2018learning} constrained parameter updates to align with the direction of experience replay, which can be seen as retaining the former gradient space through old training samples. 
Some replay-based approaches such as GEM~\cite{lopez2017gradient_gem}, A-GEM~\cite{chaudhry2018efficient_agem}, LOGD~\cite{tang2021layerwise} and MER~\cite{riemer2018learning} constrain parameter updates to align with the direction of experience replay, corresponding to preserving the previous input space and gradient space with some old training samples. 
%retained the former gradient space through old training samples and constrained parameter updates to align its direction.
In contrast to saving old training samples, OWM~\cite{zeng2019continual} and AOP~\cite{guo2022adaptive} modify parameter updates to the orthogonal direction of the previous input space. % by learning a projector matrix. 
OGD~\cite{farajtabar2020orthogonal} preserves the old gradient directions and rectifies the current gradient directions orthogonal to them. 
Orthog-Subspace~\cite{chaudhry2020continual} performs continual learning with orthogonal low-rank vector subspaces and keeps the gradients of different tasks orthogonal to each other.
GPM~\cite{saha2020gradient} maintains the gradient subspace important to the old tasks (i.e., the bases of core gradient space) for orthogonal projection in updating parameters. 
CGP~\cite{chen2022class} calculates such gradient subspace from individual classes to additionally mitigate inter-class interference.
FS-DGPM~\cite{deng2021flattening} dynamically releases unimportant bases of GPM~\cite{saha2020gradient} to improve learning plasticity and encourages the convergence to a flat loss landscape.
CUBER~\cite{lin2022beyond} selectively projects gradients to update the knowledge of old tasks that are positively related to the current task.
%employed different gradient projections to selectively update the knowledge of old tasks positively related to a new task.
TRGP~\cite{lin2021trgp} defines the ``trust region'' through the norm of gradient projection onto the subspace of previous inputs, so as to selectively reuse the frozen weights of old tasks.
%of old tasks and leveraged it to reuse the frozen weights of old tasks most relevant to a new task.
Adam-NSCL~\cite{wang2021training} instead projects candidate parameter updates into the current null space approximated by the uncentered feature covariance of the old tasks, while AdNS~\cite{kong2022balancing} considers the shared part of the previous and current null spaces.
NCL~\cite{kao2021natural} unifies Bayesian weight regularization and gradient projection, encouraging parameter updates in the null space of the old tasks while converging to a maximum of the Bayesian approximation posterior. 
Under the upper bound of the quadratic penalty in Bayesian weight regularization, RGO~\cite{liu2021continual} modifies gradient directions with a recursive optimization procedure to obtain the optimal solution.
Therefore, as regularization and replay are ultimately achieved by rectifying the current gradient directions, gradient projection corresponds to a similar modification but explicitly in parameter updates.

\begin{figure}[th]
	\centering
	\includegraphics[width=1\columnwidth]{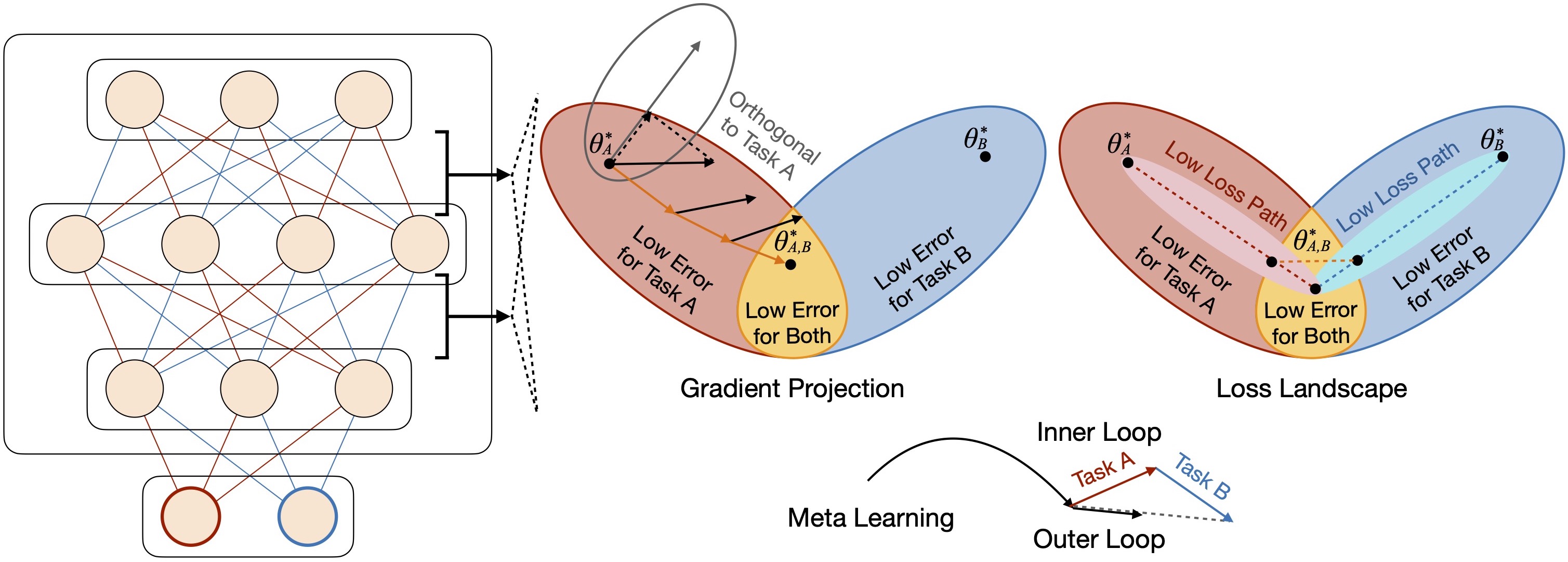} 
    \vspace{-.5cm}
	\caption{Optimization-based approach. This direction is characterized by explicit design and manipulation of the optimization programs, such as gradient projection with reference to the gradient space or input space of the old tasks (adapted from~\cite{farajtabar2020orthogonal}), meta-learning of sequentially arrived tasks within the inner loop, and exploitation of mode connectivity and flat minima in loss landscape (adapted from~\cite{mirzadeh2020linear,lin2022towards}). $\theta_A^*$, $\theta_B^*$ and $\theta_{A,B}^*$ are desirable solutions for task $A$, task $B$ and both of them, respectively. }
	\label{fig:Optimization}
    %\vspace{-.1cm}
\end{figure}

Another attractive idea is \textbf{meta-learning} or learning-to-learn for continual learning, which attempts to obtain a data-driven inductive bias for various scenarios, rather than designing it manually~\cite{hadsell2020embracing}.
OML~\cite{javed2019meta} provides a meta-training strategy that performs online updates on the sequentially arrived inputs and minimizes their interference, which can naturally obtain sparse representations suitable for continual learning.
ANML~\cite{beaulieu2020learning} extends this idea by meta-learning of a context-dependent gating function to selectively activate neurons with respect to incremental tasks.
AIM~\cite{lee2021few} learns a mixture of experts to make predictions with the representations of OML~\cite{javed2019meta} or ANML~\cite{beaulieu2020learning}, further sparsifying the representations at the level of architecture.
Meanwhile, meta-learning can work with experience replay to better utilize both the old and new training samples. For example, MER~\cite{riemer2018learning} aligns their gradient directions, while iTAML~\cite{rajasegaran2020itaml} applies a meta-updating rule to keep them in balance with each other. With the help of experience replay, La-MAML~\cite{gupta2020look} optimizes the OML~\cite{javed2019meta} objective in an online fashion with an adaptively modulated learning rate.
OSAKA~\cite{caccia2020osaka} proposes a hybrid objective of knowledge accumulation and fast adaptation, which can be resolved by obtaining a good initialization with meta-training and then incorporating knowledge of incremental tasks into the initialization.
Meta-learning can also be used to optimize specialized architectures. %task-specific and task-sharing weights.
MERLIN~\cite{kj2020meta} consolidates a meta-distribution of model parameters given the representations of each task, which allows to sample task-specific models and ensemble them for inference. Similarly, PR~\cite{henning2021posterior} adopts a Bayesian strategy to learn task-specific posteriors with a shared meta-model.
MARK~\cite{hurtado2021optimizing} maintains a set of shared weights that are incrementally updated with meta-learning and selectively masked to resolve specific tasks.
ARI~\cite{wang2022anti} combines adversarial attacks with experience replay to obtain task-specific models, which are then fused together through meta-training.

Besides, some other works refine the optimization process from a \textbf{loss landscape} perspective. %, such as flatness and mode connectivity. 
For example, rather than dedicating an algorithm, Stable-SGD~\cite{mirzadeh2020understanding} enables SGD to find a flat local minima by adapting the factors in training regime, such as dropout, learning rate decay and batch size.
%discovered that the factors in training regimes, such as dropout, learning rate decay and batch size, can have a great impact on continual learning performance via the flatness of the obtained local minima, and thus performing SGD with a proper combination of them can lead to significant improvements.
MC-SGD~\cite{mirzadeh2020linear} empirically demonstrates that the local minima obtained by multi-task learning (i.e., joint training of all incremental tasks) and continual learning can be connected by a linear path of low error, and applies experience replay to find a better solution along it.
Linear Connector~\cite{lin2022towards} adopts Adam-NSCL~\cite{wang2021training} and feature distillation to obtain respective solutions of the old and new tasks connected by a linear path of low error, followed by linear averaging. 
Further, un-/self-supervised learning (than traditional supervised learning)~\cite{gallardo2021self,hu2021well,madaan2021representational} and large-scale pre-training (than random initialization)~\cite{hu2021well,wu2022class,davari2022probing,ramasesh2021effect,mehta2021empirical} have been shown to suffer from less catastrophic forgetting.
Empirically, both can be attributed to obtaining a more robust (e.g., orthogonal, sparse and uniformly-scattered) representation~\cite{hu2021well,madaan2021representational,shi2022mimicking,ramasesh2021effect}, and converging to a wider loss basin~\cite{hu2021well,madaan2021representational,mehta2021empirical,neyshabur2020being,hao2019visualizing}, suggesting a potential link among the sensitivity of representations, parameters and task-specific errors. 
%suggesting a potential link among sensitivity to changes in representations, parameters and task-specific errors.
%a sufficiently flat loss landscape
Many efforts seek to leverage these advantages in continual learning, as we discuss next.

%self: more robust (e.g., uniformly-scattered and orthogonal) representations and flat minima~\cite{hu2021well,madaan2021representational}, 
%pre: wide minima~\cite{mehta2021empirical,hu2021well}, in the context of TL, pre-trained weights lead to a flat loss basin when fine-tuning on a single task~\cite{neyshabur2020being,hao2019visualizing}, uniformly scattered representations~\cite{shi2022mimicking}, more robust (e.g., uniformly-scattered and orthogonal) representations and flat minima~\cite{hu2021well}, 

\subsection{Representation-Based Approach}\label{Sec.4.4_Representation}
We group the approaches that create and exploit the strengths of representations for continual learning into this category.
%of creating and leveraging representational strengths for continual learning into this category. 
In addition to an earlier work that acquires sparse representations from meta-training~\cite{javed2019meta}, recent work has attempted to incorporate the advantages of self-supervised learning (SSL)~\cite{gallardo2021self,madaan2021representational,pham2021dualnet} and large-scale pre-training~\cite{mehta2021empirical,shi2022mimicking,wu2022class} to improve the representations in initialization and in continual learning.
Note that these two strategies are closely related, since the pre-training data is usually of a huge amount and without explicit labels, while the performance of SSL itself is mainly evaluated by fine-tuning on (a sequence of) downstream tasks. 
%Representative sub-directions are detailed below.
Below, we will discuss representative sub-directions.
%pre-training requires USL/SSL to process large amounts of data without explicit labels

The first is to implement \textbf{self-supervised learning} (basically with contrastive loss) for continual learning. %, which is generally based on the idea of contrastive learning. 
Observing that self-supervised representations are more robust to catastrophic forgetting, LUMP~\cite{madaan2021representational} acquires further improvements by interpolating between instances of the old and new tasks.
%For example, LUMP~\cite{madaan2021representational} observes the robustness of self-supervised representations to catastrophic forgetting and further improves it by interpolating between instances of the old and new tasks.
MinRed~\cite{purushwalkam2022challenges} further promotes the diversity of experience replay by de-correlating the stored old training samples.
%MinRed~\cite{purushwalkam2022challenges} identifies multiple additional challenges in this sub-direction and proposes to increase the diversity of experience replay by de-correlating the stored old training samples.
CaSSLe~\cite{fini2022self} converts the self-supervised loss to a distillation strategy by mapping the current state of a representation to its previous state. %, in order to improve the representations of SSL models in continual learning. 
Co2L~\cite{cha2021co2l} adopts a supervised contrastive loss to learn each task and a self-supervised loss to distill knowledge between the old and new models. 
DualNet~\cite{pham2021dualnet} trains a fast learner with supervised loss and a slow learner with self-supervised loss, with the latter helping the former to acquire generalizable representations.

The second is to use \textbf{pre-training for downstream continual learning}.
Several empirical studies suggest that downstream continual learning clearly benefits from the use of pre-training, which brings not only strong knowledge transfer but also robustness to catastrophic forgetting~\cite{gallardo2021self,ramasesh2021effect,mehta2021empirical,ostapenko2022foundational}. In particular, the benefits for downstream continual learning tend to be more apparent when pre-training with larger data size~\cite{ramasesh2021effect,ostapenko2022foundational}, larger model size~\cite{ramasesh2021effect} and contrastive loss~\cite{gallardo2021self,davari2022probing}.
However, a critical challenge is that the pre-trained knowledge needs to be adaptively leveraged for the current task while maintaining generalizability to future tasks. There are various strategies for this problem, depending on whether the pre-trained backbone is fixed or not.

\begin{figure}[th]
	\centering
    %\vspace{-.1cm}
	\includegraphics[width=0.92\columnwidth]{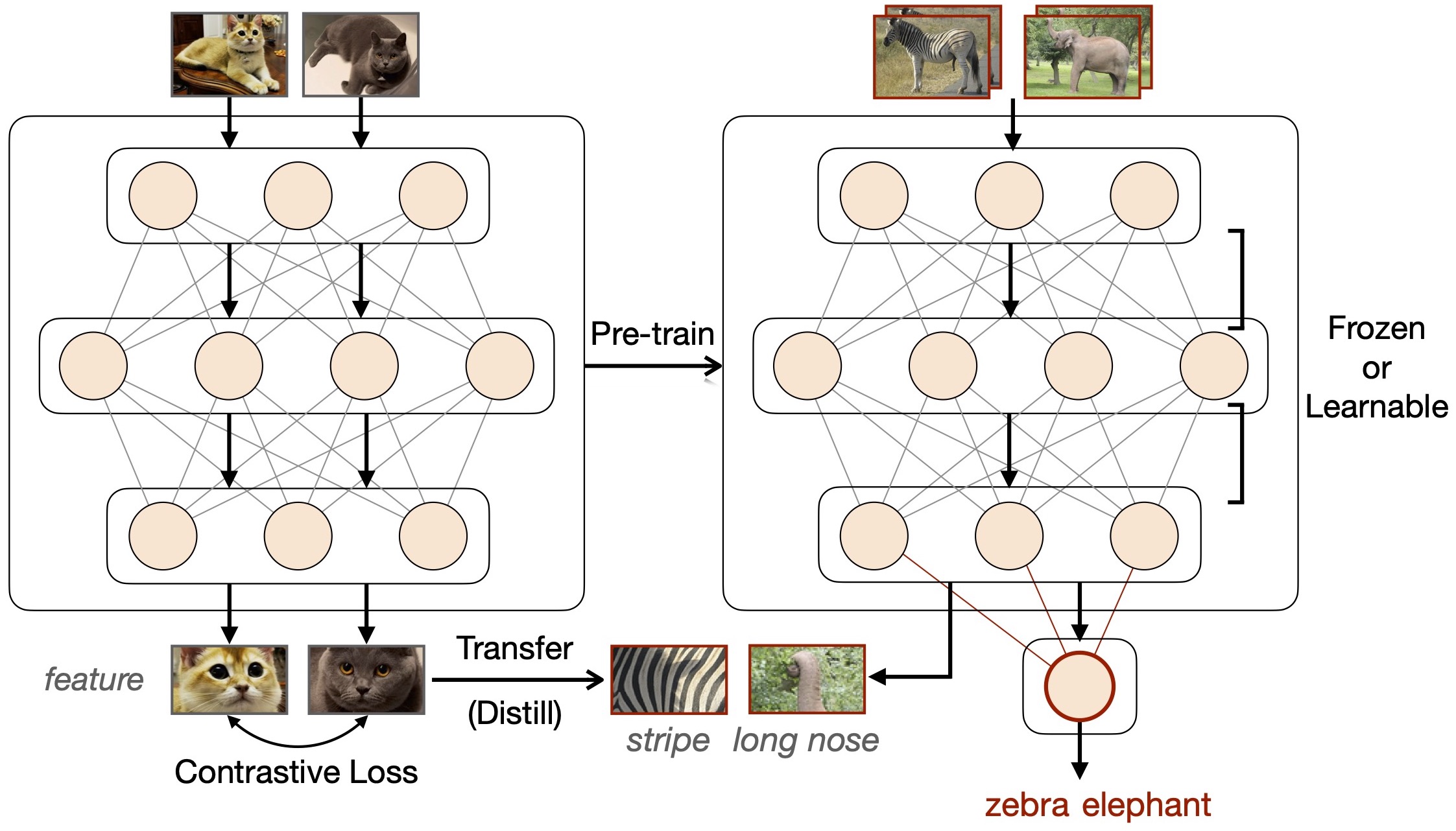}
    \vspace{-.2cm}
	\caption{Representation-based approach. This direction is characterized by creating and leveraging the strengths of representations for continual learning, such as by using self-supervised learning and pre-training. In particular, both upstream pre-training and downstream fine-tuning require continual learning, while the pre-trained representations are optionally fixed for learning specific downstream tasks.}
	\label{fig:Representation}
    \vspace{-.1cm}
\end{figure}

As for adapting a \emph{fixed} backbone, Side-Tuning~\cite{zhang2020side} and DLCFT~\cite{shon2022dlcft} train a lightweight network in parallel with the backbone and fuse their outputs linearly. 
TwF~\cite{boschini2022transfer} also trains a sibling network, but distills knowledge from the backbone in a layer-wise manner.
GAN-Memory~\cite{cong2020gan} takes advantage of FiLM~\cite{perez2018film} and AdaFM~\cite{zhao2020leveraging} to learn task-specific parameters for each layer of a pre-trained generative model, while ADA~\cite{ermis2022memory} employs Adapters~\cite{houlsby2019parameter} with knowledge distillation to adjust a pre-trained transformer.
Recent prompt-based approaches instruct the representations of a pre-trained transformer with a few prompt parameters. Such methods typically involve construction of task-adaptive prompts and inference of appropriate prompts for testing, by exploring prompt architectures to accommodate task-sharing and task-specific knowledge. Representative strategies include selecting the most relevant prompts from a prompt pool (L2P~\cite{wang2022learning_l2p}), performing a weighted summation of the prompt pool with attention factors (CODA-Prompt~\cite{smith2023coda}), using explicitly task-sharing and task-specific prompts (DualPrompt~\cite{wang2022dualprompt}) or only task-specific prompts (S-Prompts~\cite{wang2022sprompts}, HiDe-Prompt~\cite{wang2023hierarchical}), progressive expansion of task-specific prompts (Progressive Prompts~\cite{razdaibiedina2023progressive}), etc.
%For example, L2P~\cite{wang2022learning_l2p} constructs a prompt pool of fixed size and predicts the most relevant prompts with uninstructed representations. CODA-Prompt~\cite{smith2023coda} performs a weighted summation of the prompt pool with attention factors predicted from uninstructed representations
%Recent prompt-based approaches such as L2P~\cite{wang2022learning_l2p}, DualPrompt~\cite{wang2022dualprompt} and S-Prompts~\cite{wang2022sprompts} adopt a small set of learnable parameters (i.e., prompts) to instruct the representations of a pre-trained transformer
Besides, by saving prototypes, appending a nearest class mean (NCM) classifier to the backbone has proved to be a strong baseline~\cite{janson2022simple,pelosin2022simpler}, which can be further enhanced by transfer learning techniques such as the FiLM adapter~\cite{panos2023first}.
As for optimizing an \emph{updatable} backbone, F2M~\cite{shi2021overcoming} searches for flat local minima in the pre-training stage, and then learns incremental tasks within the flat region. 
CwD~\cite{shi2022mimicking} regularizes the initial-phase representations to be uniformly scattered, which can empirically mimic the representations of joint training. 
SAM~\cite{mehta2021empirical,foret2020sharpness} encourages finding a wide basin in downstream continual learning by optimizing the flatness metric.
SLCA~\cite{zhang2023slca} observes that slowly fine-tuning the backbone of a pre-trained transformer can achieve outstanding performance in continual learning, and further preserves prototype statistics to rectify the output layer.

The third is \textbf{continual pre-training} (CPT) or continual meta-training. 
As the huge amount of data required for pre-training is typically collected in an incremental manner, performing upstream continual learning to improve downstream performance is particularly important.
%A typical scenario is continual learning of vision-language pre-training models.
For example, an empirical study finds that self-supervised pre-training is more effective than supervised protocols for continual learning of vision-language models~\cite{cossu2022continual}, consistent with the results for only visual tasks~\cite{hu2021well}.
Since texts are generally more efficient than images, IncCLIP~\cite{yan2022generative} replays generated hard negative texts conditioned on images and performs multi-modal knowledge distillation for updating CLIP~\cite{radford2021learning}.
For CPT of language models, ECONET~\cite{han2021econet} designs a self-supervised framework with generative replay.
Meanwhile, continual meta-training needs to address a similar issue that the pre-trained knowledge of base classes is incrementally enriched and adapted.
IDA~\cite{liu2020incremental} imposes discriminants of the old and new models to be aligned relative to the old centers, and otherwise leaves the embedding free to accommodate new tasks.
ORDER~\cite{wang2022meta} employs unlabeled OOD data with experience replay and feature replay to cope with large inter-task differences.

\subsection{Architecture-Based Approach}\label{Sec.4.5_Architecture}
\begin{figure}[th]
	\centering
	\includegraphics[width=0.90\columnwidth]{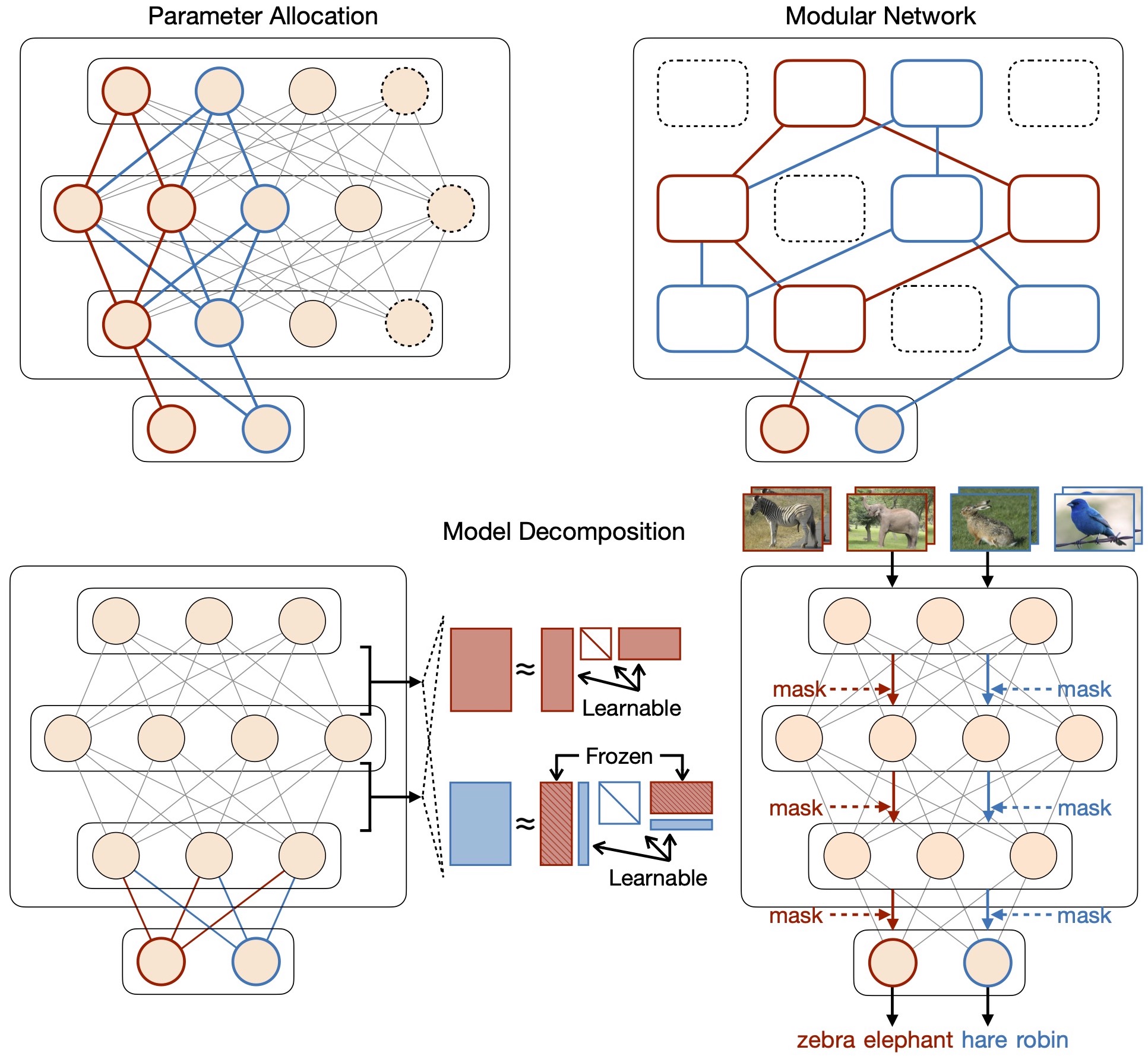} 
     \vspace{-.1cm}
	\caption{Architecture-based approach. This direction is characterized by constructing task-specific/adaptive parameters with a properly-designed architecture, such as assigning dedicated parameters to each task (parameter allocation), constructing task-adaptive sub-modules or sub-networks (modular network), and decomposing the model into task-sharing and task-specific components (model decomposition). Here we exhibit two types of model decomposition, corresponding to parameters (low-rank factorization, adapted from~\cite{hyder2022incremental}) and representations (masking of intermediate features).}
     \label{fig:Architecture}
    \vspace{-.1cm}
\end{figure}

The above strategies basically focus on learning all incremental tasks with a shared set of parameters (i.e., a single model as well as one parameter space), which is a major cause of the inter-task interference. In contrast, constructing task-specific parameters can explicitly resolve this problem.
%Since learning all incremental tasks with a shared set of parameters is a major cause of the interference between them, this direction tries to learn task-specific parameters explicitly to resolve this problem.
Previous work generally separates this category into \emph{parameter isolation} and \emph{dynamic architecture}, depending on whether the network architecture is fixed or not.
Here, we instead focus on the way of implementing task-specific parameters, extending the above concepts to parameter allocation, model decomposition and modular network (see Fig.~\ref{fig:Architecture}).

\textbf{Parameter allocation} features an isolated parameter subspace dedicated to each task throughout the network, where the architecture can be fixed or dynamic in size. 
Within a \emph{fixed} network architecture, Piggyback~\cite{mallya2018piggyback}, HAT~\cite{serra2018overcoming}, SupSup~\cite{wortsman2020supermasks}, MEAT~\cite{xue2022meta}, WSN~\cite{kang2022forget} and H$^2$~\cite{jin2022helpful} explicitly optimize a binary mask to select dedicated neurons or parameters for each task, with the masked regions of the old tasks (almost) frozen to prevent catastrophic forgetting. 
PackNet~\cite{mallya2018packnet}, UCL~\cite{ahn2019uncertainty}, CLNP~\cite{golkar2019continual}, AGS-CL~\cite{jung2020continual} and NISPA~\cite{gurbuz2022nispa} explicitly identify the important neurons or parameters for the current task and then release the unimportant parts to the following tasks, which can be achieved by iterative pruning~\cite{mallya2018packnet}, activation value~\cite{golkar2019continual,jung2020continual,gurbuz2022nispa}, uncertainty estimation~\cite{ahn2019uncertainty}, etc. 
%utilize the available network parameters to learn each task, 
%They need to identify and stabilize the important neurons or parameters for the current task, releasing the unimportant parts to the following tasks, which can be achieved by iterative pruning~\cite{mallya2018packnet}, activation level~\cite{golkar2019continual,jung2020continual,gurbuz2022nispa}, uncertainty estimation~\cite{ahn2019uncertainty}, etc. 
Since the network capacity is limited, ``free'' parameters tend to saturate as more incremental tasks are introduced. Therefore, these methods typically require sparsity constraints on parameter usage and selective reuse of the frozen old parameters, which might affect the learning of each task. 
%\cite{mallya2018packnet,mallya2018piggyback,serra2018overcoming,wortsman2020supermasks,kang2022forget,ahn2019uncertainty,golkar2019continual,jung2020continual,gurbuz2022nispa}
%\cite{serra2018overcoming,wortsman2020supermasks,jin2022helpful,golkar2019continual,gurbuz2022nispa}
To alleviate this dilemma, the network architecture can be \emph{dynamically expanded} if its capacity is not sufficient to learn a new task well, such as DEN~\cite{yoon2018lifelong}, CPG~\cite{hung2019compacting} and DGMa/DGMw~\cite{ostapenko2019learning_dgmw}.
The dynamic architecture can also be explicitly optimized to improve parameter efficiency and knowledge transfer, such as by reinforcement learning (RCL~\cite{xu2018reinforced}, BNS~\cite{qin2021bns}), architecture search (LtG~\cite{li2019learn}, BNS~\cite{qin2021bns}), variational Bayes (BSA~\cite{kumar2021bayesian}), etc. As the network expansion should be much slower than the task increase to ensure scalability, constraints on sparsity and reusability remain important.

\textbf{Model decomposition} separates a model explicitly into task-sharing and task-specific components, where the task-specific components are often expandable.
%where the latter is typically expandable.
%grows linearly with the number of incremental tasks. 
For a regular network, the task-specific components could be parallel branches (ACL~\cite{ebrahimi2020adversarial}, ReduNet~\cite{wu2021incremental}, EPIE-Net~\cite{du2022efficient}), adaptive layers (GVCL~\cite{loo2020generalized}, DyTox~\cite{douillard2022dytox}), masks or mask generators for intermediate features (CCLL~\cite{singh2020calibrating}, CCG~\cite{abati2020conditional}, MARK~\cite{hurtado2021optimizing}). 
%masks (CCLL~\cite{singh2020calibrating}) or mask generators (CCG~\cite{abati2020conditional}, MARK~\cite{hurtado2021optimizing}) for intermediate feature maps. 
Note that the feature masks for model decomposition do not operate in parameter space and are not binary for each task, thus fundamentally different from the binary masks for parameter allocation. 
Besides, the network parameters themselves can be decomposed into task-sharing and task-specific elements, such as by additive decomposition (APD~\cite{yoon2019scalable}), singular value decomposition (RCM~\cite{kanakis2020reparameterizing}), filter atom decomposition (FAS~\cite{miao2021continual}) and low-rank factorization (IBP-WF~\cite{mehta2021continual}, IRU~\cite{hyder2022incremental}).
As the number of task-specific components usually grows linearly with incremental tasks, their resource efficiency determines the scalability of this sub-direction.
%the scalability of this sub-direction depends on the resource efficiency of its form.
%their computational and storage efficiency determines the scalability.

%For example, the task-specific components could be dedicated modules of the network (ACL~\cite{ebrahimi2020adversarial}, ReduNet~\cite{wu2021incremental}, EPIE-Net~\cite{du2022efficient}), or layer-adaptive modifications (CCG~\cite{abati2020conditional}, CCLL~\cite{singh2020calibrating}, GVCL~\cite{loo2020generalized}, MARK~\cite{hurtado2021optimizing}, DyTox~\cite{douillard2022dytox})

\textbf{Modular network} leverages parallel sub-networks or sub-modules to learn incremental tasks in a differentiated manner, without pre-defined task-sharing or task-specific components. As an early work, Progressive Networks~\cite{rusu2016progressive} introduces an identical sub-network for each task and allows knowledge transfer from other sub-networks via adaptor connections. 
Expert Gate~\cite{aljundi2017expert} and a subsequent work~\cite{collier2020routing} employ a mixture of experts~\cite{jacobs1991adaptive} to learn incremental tasks, expanding one expert as each task is introduced. %, recognizing task relatedness with an undercomplete autoencoder~\cite{aljundi2017expert} or routing network~\cite{collier2020routing}. 
PathNet~\cite{fernando2017pathnet} and RPSNet~\cite{rajasegaran2019random} pre-allocate multiple parallel networks to construct a few candidate paths, i.e., layer-wise compositions of network modules, and select the best path for each task.
MNTDP~\cite{veniat2020efficient} and LMC~\cite{ostapenko2021continual} attempt to find the optimal layout from previous sub-modules and potentially new sub-modules.
Similar to parameter allocation, these efforts are intentional to construct task-specific models, while the combination of sub-networks or sub-modules allows explicit reuse of knowledge.
%reduce parameter usage and facilitate knowledge transfer.
In addition, the sub-networks can be encouraged to learn incremental tasks in parallel.
%each sub-network can be encouraged to learn all incremental tasks. 
Model Zoo~\cite{ramesh2021model} expands a sub-network to learn each new task with experience replay of the old tasks, and ensembles all sub-networks for prediction. 
CoSCL~\cite{wang2022coscl} and CAF~\cite{wang2023incorporating} ensembles multiple continual learning models and modulates the predictive similarity between them, proving to be effective in resolving the discrepancy of task distribution and improving the flatness of loss landscape.
%ensembles a fixed number of sub-networks to perform continual learning in parallel, and modulates the predictive similarity between sub-networks.

In a broader sense, sampling parameters from task-conditioned parameter distributions (MERLIN~\cite{kj2020meta}, PR~\cite{henning2021posterior}, PGMA~\cite{hu2019overcoming}, HNET~\cite{von2019continual}), as well as stabilizing important parameters with weight regularization, can be seen as a form of deriving task-specific/adaptive parameters.
In contrast to other directions, most architecture-based approaches amount to de-correlating incremental tasks in network parameters, which can almost avoid catastrophic forgetting but affect scalability and inter-task generalizability. In particular, task identities are often required to determine which set of parameters to use, thus greatly constraining their applications. 
To overcome this limitation, task identities can be inferred from the responses (e.g., predictive uncertainty) of each task-specific model~\cite{aljundi2017expert,henning2021posterior,kim2022theoretical}.
The function of task-identity prediction can also be learned explicitly from incremental tasks, using other continual learning strategies to mitigate catastrophic forgetting~\cite{collier2020routing,jin2022helpful,ebrahimi2020adversarial,abati2020conditional,mehta2021continual,henning2021posterior,kj2020meta}. 
%Also, the mapping function from representations to task labels can be explicitly learned from incremental tasks, using other continual learning strategies to mitigate catastrophic forgetting~\cite{collier2020routing,jin2022helpful,ebrahimi2020adversarial,abati2020conditional,mehta2021continual,henning2021posterior,kj2020meta}. 
%The function of task-label prediction (i.e., mapping from representations to task labels) can also be learned from incremental tasks, using other continual learning strategies to mitigate catastrophic forgetting~\cite{collier2020routing,jin2022helpful,ebrahimi2020adversarial,abati2020conditional,mehta2021continual,henning2021posterior,kj2020meta}.

Besides, the design and choice of a \textbf{basic architecture} can largely impact the continual learning performance. For example, wider networks tend to be more robust to catastrophic forgetting due to more orthogonality and sparsity in gradient directions, as well as a lazier training regime~\cite{mirzadeh2022wide,mirzadeh2022architecture}.
Batch Normalization (BN) layers~\cite{ioffe2015batch} tend to introduce biased moments towards the current task, resulting in sub-optimal performance of the previous tasks~\cite{pham2021continual,cha2022task,lyu2023overcoming}.
Dropout~\cite{srivastava2014dropout} in a stable network behaves like a gating mechanism to create task-specific pathways and thus can mitigate catastrophic forgetting~\cite{mirzadeh2020dropout}.

%\hangx{we may have a conclusive summary here to demonstrate what are the advantages and disadvantages for each strategy}

%\section{Scenario Challenge in Application: The Case of Visual Classification}
\section{Scenario Complexity in Application: The Case of Visual Classification}\label{Sec.5_Scenario}
%\hangx{this section could be more brief. Emphasize the challenges }

Due to the complexity of the real world, practical applications present a variety of specialized challenges. Here we categorize these challenges into \textbf{scenario complexity} and \textbf{task specificity}, with an extensive analysis of how continual learning methods are adapted to them.
In this section, we demonstrate the challenges of scenario complexity, using visual classification as an example.
%\subsection{Data/Domain/Task-Incremental Learning}
%Since task labels are provided in both training and testing, such settings generally do not have specialized requirements for continual learning methods. Empirical study can compare the efficacy of representative approaches.

%\subsection{Class-Incremental Learning}
%For continual learning, \emph{Task-Incremental Learning} (TIL) usually serves as a basic setting, i.e., task identities are provided in both training and testing. In contrast, \emph{Class-Incremental Learning} (CIL) is characterized by task-agnostic inference, i.e., testing without providing task identities, which is more natural but also more challenging for classification tasks. 
\subsection{Task-Agnostic Inference}%\label{Sec.5.1_CIL}
Continual learning usually has \emph{Task-Incremental Learning} (TIL) as a basic setup, i.e., task identities are provided in both training and testing. In contrast, task-agnostic inference that avoids the use of task identities for testing tends to be more natural but more challenging in practical applications, which is known as \emph{Class-Incremental Learning} (CIL) for classification tasks.
For example, let's consider two binary classification tasks: (1) ``zebra'' and ``elephant''; and (2) ``hare'' and ``robin''. After learning these two tasks, TIL needs to know which task it is and then classify the two classes accordingly, while CIL directly classifies the four classes at the same time. Therefore, CIL has received increasingly more attention and become almost the most representative scenario for continual learning.

\begin{figure}[ht]
    \vspace{-.1cm}
	\centering
	\includegraphics[width=0.98\columnwidth]{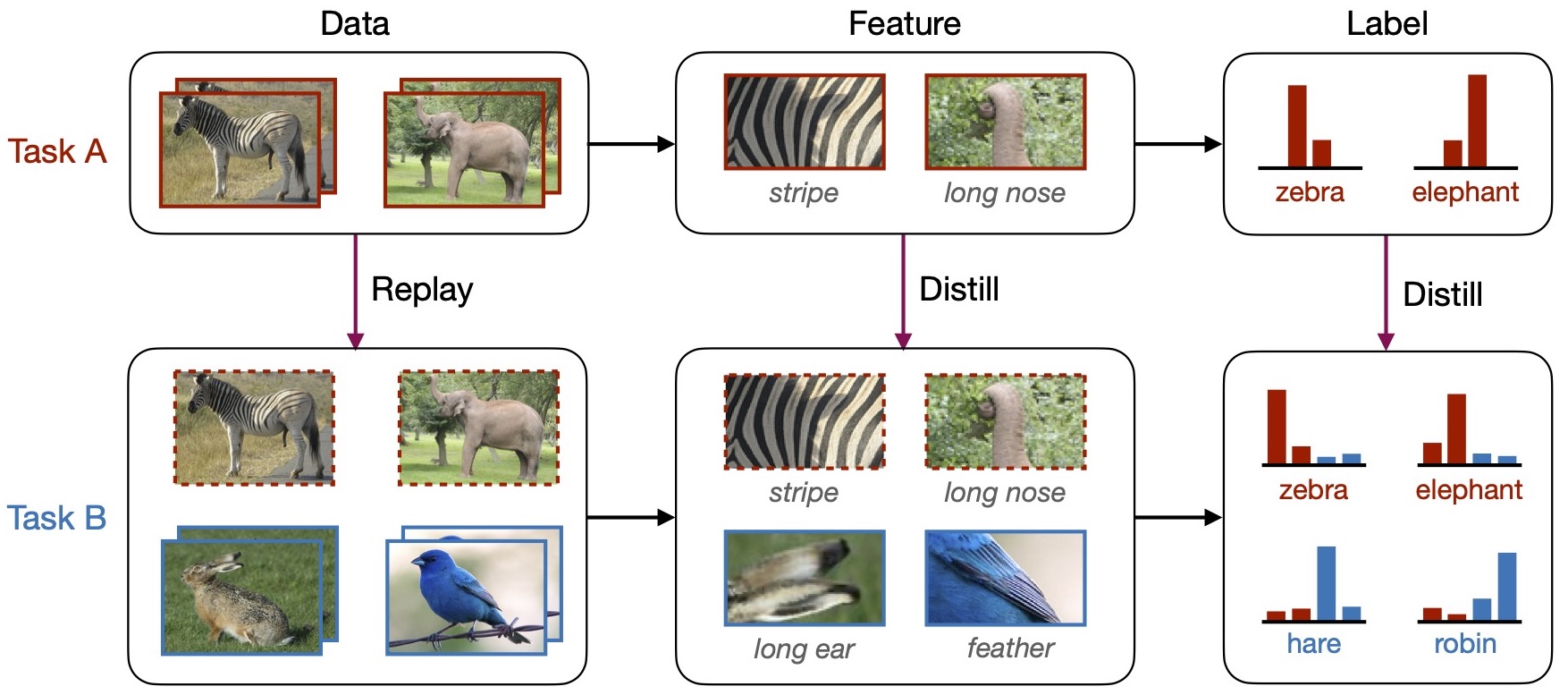} 
    \vspace{-.1cm}
	\caption{Representative strategies for class-incremental learning. Catastrophic forgetting can be mitigated with respect to data space (experience replay), feature space (knowledge distillation) and label space (knowledge distillation). This figure is adapted from~\cite{hu2021distilling}.}
	\label{fig:CIL}
    %\vspace{-.1cm}
\end{figure}

The CIL problem can be disentangled into \emph{within-task prediction} and \emph{task-identity prediction}, where the latter is a particular challenge posed by task-agnostic inference and has been shown to be closely related to the OOD detection~\cite{kim2022theoretical}. 
%To address these challenges
To overcome catastrophic forgetting in CIL, numerous efforts attempt to impose the behaviors of the previous model on the current model, in terms of data, feature, and label spaces (see Fig.~\ref{fig:CIL}).
Since the incremental classes are disjoint, the new training samples are usually OOD from the old ones. In this regard, replay of the old training samples is able to impose an end-to-end effect, while feature and label distillations of the new training samples are limited to the output side with a biased distribution~\cite{hu2021distilling}. 
Therefore, many state-of-the-art methods are built on the framework of \emph{experience replay} and then incorporate \emph{knowledge distillation}, as we have discussed in Sec.~\ref{Sec.4.2_Replay}. In Table~\ref{table:CIL}, we summarize these CIL methods based on their main focus. %main contributions to data, representation and prediction.

\begin{table}[th]
    \renewcommand\arraystretch{1.8}
    %\vspace{-0.1cm}
	\caption{Summary of representative CIL methods with the use of experience replay.
	These methods further improve the memory buffer, feature distillation or label distillation to achieve better performance, corresponding to data, feature and label spaces, respectively.} 
	%	We categorize them based on their main focus, including data space (modification of the memory buffer), feature space (distillation of feature representations) and label space (distillation of predicted logits).
     \vspace{-0.2cm}
       \centering
      \smallskip
      \renewcommand\arraystretch{1.8}
	%\resizebox{0.95 cm}{!}{ 
      \begin{tabular}{c|m{6.3cm}}
	 \hline
      Main Focus & Representative Method \\ %Main Contributions 
    \hline
     Data Space &iCaRL~\cite{rebuffi2017icarl}, GSS~\cite{aljundi2019gradient}, Mnemonics~\cite{liu2020mnemonics}, TPCIL~\cite{tao2020topology}, GDumb~\cite{prabhu2020gdumb}, DER++~\cite{buzzega2020dark}, RMM~\cite{liu2021rmm}, HAL~\cite{chaudhry2021using}, MRDC~\cite{wang2021memory}, CSI~\cite{kim2022theoretical}, X-DER~\cite{boschini2022class}  \\
      \hline
     Feature Space&LUCIR~\cite{hou2019learning_lucir}, PODNet~\cite{douillard2020podnet}, TPCIL~\cite{tao2020topology}, PCL~\cite{hu2021continual}, AANets~\cite{liu2021adaptive}, DER~\cite{yan2021dynamically}, DDE~\cite{hu2021distilling}, GeoDL~\cite{simon2021learning}, PASS~\cite{zhu2021prototype}, Co2L~\cite{cha2021co2l}, AFC~\cite{kang2022class}, SP-CIL~\cite{wu2022class}, ELI~\cite{joseph2022energy}, CwD~\cite{shi2022mimicking}, CSCCT~\cite{ashok2022class}, FOSTER~\cite{wang2022foster}, FASP~\cite{miao2021continual}, CLS-ER~\cite{arani2021learning}  \\
      \hline
     Label Space&LwF~\cite{li2017learning}, iCaRL~\cite{rebuffi2017icarl}, GEM~\cite{lopez2017gradient_gem}, A-GEM~\cite{chaudhry2018efficient_agem}, EEIL~\cite{castro2018end}, BiC~\cite{wu2019large_bic}, WA~\cite{zhao2020maintaining}, DER++~\cite{buzzega2020dark}, ScaIL~\cite{belouadah2020scail}, S\&B~\cite{kim2021split}, SS-IL~\cite{ahn2021ss}, Coil~\cite{zhou2021co} \\
     % ER-ACE~\cite{caccia2021new}
     %Catastrophic Forgetting&LwF~\cite{li2017learning}, iCaRL~\cite{rebuffi2017icarl}, EEIL~\cite{castro2018end}, TPCIL~\cite{tao2020topology}, DER~\cite{buzzega2020dark}   \\
     %Stability-Plasticity Trade-off & Mnemonics~\cite{liu2020mnemonics}  \\
     %Representation Shift & PODNet~\cite{douillard2020podnet}  \\
     %Semantic Shift \& Imbalance & BiC~\cite{wu2019large_bic}, LUCIR~\cite{hou2019learning_lucir}, WA~\cite{zhao2020maintaining}   \\
    \hline
	\end{tabular}
	%}
	\label{table:CIL}
    \vspace{-.2cm}
\end{table}

To avoid the additional resource overhead and potential privacy issues of retaining old training samples, many efforts attempt to perform CIL without experience replay, i.e., \emph{Data-Free} CIL. An intriguing idea is to replay synthetic data produced by inverting a frozen copy of the old classification model, such as DeepInversion~\cite{yin2020dreaming}, ABD~\cite{smith2021always}, RRL~\cite{gao2022r} and CF-IL~\cite{pourkeshavarzi2021looking}, which usually further incorporate knowledge distillation to compensate the lost information in model inversion.
Other methods exploit the class-wise statistics of feature representations to obtain a balanced classifier, such as by imposing the representations to be transferable and invariant (SPB~\cite{wu2021striking}, IL2A~\cite{zhu2021class}, SSRE~\cite{zhu2022self}, FeTrIL~\cite{petit2022fetril}), or compensating explicitly the representation shifts (SDC~\cite{yu2020semantic}, RER~\cite{toldo2022bring}).
Besides, the use of adequate pre-training provides well-distributed representations, which allows strong CIL performance without the need of experience replay. Representative strategies include prompting~\cite{wang2022learning_l2p,smith2023coda,wang2022dualprompt,wang2023hierarchical} or carefully fine-tuning~\cite{zhang2023slca,shi2021overcoming,mehta2021empirical} the backbone, saving prototypes to rectify the predictions~\cite{janson2022simple,pelosin2022simpler,zhang2023slca}, incorporating transfer learning techniques~\cite{panos2023first,ermis2022memory,boschini2022transfer}, etc., as discussed in Sec.~\ref{Sec.4.4_Representation}.

For continual learning (especially CIL) of visual classification tasks, the current state-of-the-art methods mainly focus on image classification with attention to relatively complex and large-scale datasets, such as ILSVRC2012~\cite{krizhevsky2012imagenet} and its derivatives. 
There are also many benchmarks for video classification~\cite{lomonaco2017core50,roady2020stream,park2021class,villa2022vclimb}, varying in size and purpose.
%differing in their size and purpose.
%such as CORe50~\cite{lomonaco2017core50}, Stream-51~\cite{roady2020stream}~\cite{park2021class}, vCLIMB~\cite{villa2022vclimb} 

%are no longer constrained to simple datasets such as MNIST, SVHN, CIFAR-10 and even CIFAR-100, but focus on larger-scale datasets such as ImageNet 

% \subsection{Limited Labeling}
\subsection{Scarcity of Labeled Data}
Most of the current continual learning settings assume that incremental tasks have sufficiently large amounts of labeled data, which is often expensive and difficult to obtain in practical applications. For this reason, there is a growing body of work focusing on the scarcity of labeled data in continual learning. 

A representative scenario is \emph{Few-Shot Continual Learning} (FSCL) or further specified as \emph{Few-Shot} CIL (FSCIL)~\cite{tao2020few}, where the model first learns some base classes for initialization with a large number of training samples, and then learns a sequence of novel classes with only a small number of training samples. 
The extremely limited training samples exacerbate the overfitting of previously-learned representations to subsequent tasks, which can be alleviated by recent work such as preserving the topology of representations (TOPIC~\cite{tao2020few}), constructing an exemplar relation graph for knowledge distillation (ERL~\cite{dong2021few}), selectively updating only unimportant parameters (FSLL~\cite{mazumder2021few}) or stabilizing important parameters (LCwoF~\cite{kukleva2021generalized}), cooperating fast-slow weights (MgSvF~\cite{zhao2021mgsvf}), updating parameters within the flat region of loss landscape (F2M~\cite{shi2021overcoming}), meta-learning of a good initialization (MetaFSCIL~\cite{chi2022metafscil}), and generative replay of old data distributions (ERDFR~\cite{liu2022few}).

There are many other efforts keeping the backbone fixed in subsequent continual learning, so as to decouple the learning of \emph{representation} and \emph{classifier}. Under this framework, representative strategies can be conceptually separated into two aspects. The first is to obtain compatible and extensible representations from massive base classes, such as by enforcing the representations compatible with simulated incremental tasks (SPPR~\cite{zhu2021self}, LIMIT~\cite{zhou2022few}), reserving the feature space with virtual prototypes for future classes (Fact~\cite{zhou2022forward}), using angular penalty loss with data augmentation (ALICE~\cite{peng2022few}), using self-supervised learning (S3C~\cite{kalla2022s3c}), providing extra constraints from margin-based representations (CLOM~\cite{zou2022margin}), etc. The second is to obtain an adaptive classifier from a sequence of novel classes, such as by evolving the classifier weights with a graph attention network (CEC~\cite{zhang2021few}), developing a tree-based hierarchical classifier with Gaussian processes (GP-Tree~\cite{achituve2021gp}), performing hyperdimensional computing (C-FSCIL~\cite{hersche2022constrained}), and sampling stochastic classifiers from a weight distribution (S3C~\cite{kalla2022s3c}). 
Besides, auxiliary information such as semantic word vectors (SKD~\cite{cheraghian2021semantic}, Mixture of Subspaces~\cite{cheraghian2021synthesized}, Subspace Reg~\cite{akyurek2021subspace}, FSIL-GAN~\cite{agarwal2022semantics}) and sketch exemplars (DIY~\cite{bhunia2022doodle}) can be incorporated to enrich the limited training samples.

In addition to a few labeled data, there is usually a large amount of unlabeled data available and collected over time. The first practical setting is called \emph{Semi-Supervised Continual Learning} (SSCL)~\cite{wang2021ordisco}, which considers incremental data as partially labeled. As an initial attempt, ORDisCo~\cite{wang2021ordisco} learns a semi-supervised classification model together with a conditional GANs for generative replay, and regularizes discriminator consistency to mitigate catastrophic forgetting. Subsequent work includes training an adversarial autoencoder to reconstruct images (AAE~\cite{lechat2021semi}), imposing predictive consistency among augmented and interpolated data (CCIC~\cite{boschini2022continual}), leveraging the nearest-neighbor classifier to distill class-instance relationships (NNCSL~\cite{kang2022soft}), etc.
The second scenario assumes that there is an external unlabeled dataset to facilitate supervised continual learning, e.g., by knowledge distillation (GD~\cite{lee2019overcoming}) and data augmentation (L2I~\cite{tang2022learning}).
The third scenario is to learn representations from incremental unlabeled data~\cite{cossu2022continual,hu2021well}, i.e., \emph{Unsupervised Continual Learning} (UCL), which becomes an increasingly important topic for updating pre-trained knowledge in foundation models.

\subsection{Generic Learning Paradigm}%\label{Sec.5.3_GCL}
Potential challenges of the learning paradigm can be summarized in a broad concept called \emph{General Continual Learning} (GCL)~\cite{aljundi2019task,de2021continual_review,buzzega2020dark}, where the model observes incremental data in an online fashion without explicit task boundaries. Correspondingly, GCL consists of two interconnected settings: \emph{Task-Free Continual Learning} (TFCL)~\cite{aljundi2019task}, where the task identities are not accessible in either training or testing; and \emph{Online Continual Learning} (OCL)~\cite{aljundi2019gradient}, where the training samples are observed in an one-pass data stream. 
Since TFCL usually accesses only a small batch of training samples at a time for gradual changes in task distributions, while OCL usually requires only the data label rather than the task identity for each training sample, many representative methods for TFCL and OCL are indeed compatible (see Table~\ref{table:gcl} for their target scenarios). 
%Here we summarize these methods with their target scenarios in Table~\ref{table:gcl}. 
\begin{table}[th]
    \renewcommand\arraystretch{1.8}
     \centering
    \vspace{-0.1cm}
	\caption{Summary of TFCL, OCL and GCL (both TFCL and OCL) methods. Of note, some methods are not designed for specific scenarios, but are used as strong baselines in subsequent work.} 
     \vspace{-.2cm}
      \smallskip
      \renewcommand\arraystretch{1.8}
	%\resizebox{0.95 cm}{!}{ 
      \begin{tabular}{c|m{6.7cm}}
	 \hline
     Scenario & Representative Method \\ %Main Contributions 
    \hline
     TFCL & CN-DPM~\cite{lee2019neural_cndpm},~\cite{jerfel2019reconciling}, VariGrow~\cite{ardywibowo2022varigrow}, ODDL~\cite{ye2022task} \\
      \hline
     OCL & InstAParam~\cite{chen2020mitigating}, CBRS~\cite{chrysakis2020online}, ASER~\cite{shim2021online}, InfoRS~\cite{sun2021information}, DVC~\cite{gu2022not}, CTN~\cite{pham2020contextual}, NCCL~\cite{yin2021mitigating}, ILOS~\cite{he2020incremental}, SCR~\cite{mai2021supervised}, CVT~\cite{wang2022online}, OCM~\cite{guo2022online}, ER-ACE~\cite{caccia2021new}, RAR~\cite{zhang2022simple}, PoLRS~\cite{cai2021online}, AOP~\cite{guo2022adaptive}, CV~\cite{he2022online} \\
      \hline
     GCL & GSS~\cite{aljundi2019gradient}, MIR~\cite{aljundi2019mir}, CLIB~\cite{koh2021online}, GMED~\cite{jin2021gradient}, DRO~\cite{wang2022improving}, GDumb~\cite{prabhu2020gdumb}, DER~\cite{buzzega2020dark}, GEM~\cite{lopez2017gradient_gem}, A-GEM~\cite{chaudhry2018efficient_agem}, DSDM~\cite{pourcel2022online}, CoPE~\cite{de2021continual}, BLD~\cite{fini2020online} \\
    \hline
	\end{tabular}
	%}
	\label{table:gcl}
    %\vspace{-.1cm}
\end{table}

Specifically, some of them attempt to learn specialized parameters in a growing architecture. For TFCL, CN-DPM~\cite{lee2019neural_cndpm} adopts Dirichlet process mixture models to construct a growing number of neural experts, while a concurrent work~\cite{jerfel2019reconciling} derives such mixture models from a probabilistic meta-learner. VariGrow~\cite{ardywibowo2022varigrow} employs an energy-based novelty score to decide whether to extend a new expert or update an old one. ODDL~\cite{ye2022task} estimates the discrepancy between the current memory buffer and the previously-learned knowledge as an expansion signal. For OCL, InstAParam~\cite{chen2020mitigating} selects and consolidates appropriate network paths for individual training samples. 
%DSDM provides an associative content-addressable memory model and dynamically expands address nodes as a function of the input.

%\emph{Task-Free Continual Learning} (TFCL) and \emph{Online Continual Learning} (OCL) are two realistic and interconnected settings. They both belong to the broader concept of General Continual Learning (GCL), where the model observes incremental data in an online fashion without explicit task boundaries. 

On the other hand, many TSCL methods and most OCL methods are built on experience replay, focusing on construction, management and exploitation of a memory buffer. As the training samples arrive in small batches, the information of task boundaries is less effective, and reservoir sampling usually serves as an effective baseline strategy for sample selection. More advanced strategies prioritize the replay of those training samples that are informative (InfoRS~\cite{sun2021information}), diversified in parameter gradients (GSS~\cite{aljundi2019gradient}) or previously-learned knowledge (ODDL~\cite{ye2022task}), balanced in class labels (CBRS~\cite{chrysakis2020online}, GDumb~\cite{prabhu2020gdumb}, CoPE~\cite{de2021continual}), and beneficial to latent decision boundaries (ASER~\cite{shim2021online}). Meanwhile, the memory buffer can be dynamically managed, such as by removing less important training samples (CLIB~\cite{koh2021online}), editing the old training samples (GMED~\cite{jin2021gradient}) or their distributions (DRO~\cite{wang2022improving}) to be more likely forgotten, and retrieving the old training samples that are susceptible to interference (MIR~\cite{aljundi2019mir}, DVC~\cite{gu2022not}).
To adaptively exploit the memory buffer, representative strategies include using additional parameters to construct task-specific features (CTN~\cite{pham2020contextual}) or calibrate the network (NCCL~\cite{yin2021mitigating}), performing knowledge distillation (ILOS~\cite{he2020incremental}, DER~\cite{buzzega2020dark}), evolving prototypes in feature space (CoPE~\cite{de2021continual}),
improving representations with contrastive learning (SCR~\cite{mai2021supervised}, DVC~\cite{gu2022not}, CVT~\cite{wang2022online}), maximizing mutual information between input data and their label predictions (OCM~\cite{guo2022online}), using asymmetric cross-entropy (ER-ACE~\cite{caccia2021new}) or constrained gradient directions (GEM~\cite{lopez2017gradient_gem}, A-GEM~\cite{chaudhry2018efficient_agem}) of the old and new training samples, repeated rehearsal with data augmentation (RAR~\cite{zhang2022simple}), properly adjusting the learning rate (PoLRS~\cite{cai2021online}, CLIB~\cite{koh2021online}),
training from scratch with the memory buffer (GDumb~\cite{prabhu2020gdumb}), etc. 

In addition, there are some OCL methods without using experience replay, such as by distilling knowledge from the current batch of training samples (BLD~\cite{fini2020online}), and using a pre-trained backbone for orthogonal gradient projection (AOP~\cite{guo2022adaptive}) or feature replay (CV~\cite{he2022online}).
%\vspace{-.3cm}

%\section{Task-Specific Challenge in Application}
\section{Task Specificity in Application}\label{Sec.6_Task}
In this section, we describe the challenges of task specificity in various realistic applications of continual learning.

\subsection{Object Detection}
\emph{Incremental Object Detection} (IOD) is a typical extension of continual learning for object detection, where the training samples annotated with different classes are introduced in sequence, and the model needs to correctly locate and identify the object instances belonging to the previously-learned classes. 
%A typical extension of continual learning for object detection is called \emph{Incremental Object Detection} (IOD), where the training samples annotated with different classes are introduced in sequence, and the model needs to correctly locate and identify the objects belonging to the previously-learned classes. 
Unlike visual classification with only one object instance appearing in each training sample, object detection usually has multiple object instances belonging to the old and new classes appearing together.
Such \emph{co-occurrence} poses an additional challenge for IOD, where the object instances of old classes are annotated as the ``background'' when learning new classes, thus exacerbating catastrophic forgetting.
On the other hand, this makes knowledge distillation a naturally powerful strategy for IOD, since the object instances of old classes can be obtained from new training samples to constrain the differences in responses between the old and new models. %LwF-like~\cite{li2017learning} 

As an early work, ILOD~\cite{shmelkov2017incremental_ilod} distills the responses for old classes to prevent catastrophic forgetting on Fast R-CNN~\cite{girshick2015fast}, followed by RKT~\cite{ramakrishnan2020relationship} that further distills the relation of co-occurrence in selected proposals.
The idea of knowledge distillation is then introduced to other object detectors, such as SID~\cite{peng2021sid} on CenterNet~\cite{zhou2019objects}, RILOD~\cite{li2019rilod} on RetinaNet~\cite{lin2017focal}, ERD~\cite{feng2022overcoming} on GFLV1~\cite{li2020generalized}, CIFRCN~\cite{hao2019end}, Faster ILOD~\cite{peng2020faster}, DMC~\cite{zhang2020class}, BNC~\cite{dong2021bridging} and IOD-ML~\cite{kj2021incremental} on Faster R-CNN~\cite{ren2015faster}, etc.
%In particular, unlabeled in-the-wild data can be exploited to distill the old and new models into a shared model
Some methods exploit the unlabeled in-the-wild data to distill the old and new models into a shared model, in order to bridge potential non co-occurrence (BNC~\cite{dong2021bridging}) and to achieve a better stability-plasticity trade-off (DMC~\cite{zhang2020class}). 
To further mitigate the negative effects of knowledge distillation on learning plasticity, IOD-ML~\cite{kj2021incremental} adopts meta-learning to reshape parameter gradients into a balanced direction between the old and new classes.

IOD is not only applicable for 2D images, but also for 3D images~\cite{zhao2022static} and videos~\cite{wang2021wanderlust}. Besides, there are many other related settings, such as incremental few-shot detection~\cite{perez2020incremental}, where a pre-trained object detector registers new classes with only a few annotated data; and open world object detection~\cite{joseph2021towards}, where the object detector needs to identify potential object instances of unknown classes and register them after receiving corresponding annotations.
%ONCE~\cite{perez2020incremental} focuses on Incremental Few-Shot Detection, where a pre-trained detection model registers new classes with only a few annotated training samples, through meta-learning a class-specific code generator for flexible adaptation on a class-generic feature extractor. ORE~\cite{joseph2021towards} considers Open World Object Detection, where the model identifies the unknown classes and then registers these classes after receiving the corresponding labels, and address theses challenges by energy-based unknown identifications and experience replay. 

\subsection{Semantic Segmentation}
\emph{Continual Semantic Segmentation} (CSS) aims at pixel-wise prediction of classes and learning new classes in addition to the old ones.
Similar to IOD, the object instances of old and new classes can appear together. 
Some early efforts utilize full annotations of both the old and new classes in continual learning~\cite{michieli2019incremental,michieli2021knowledge}. 
However, due to the significant expense and time cost of re-annotating the old classes, more attention has been focused on using annotations of only the new classes, which leads to the old classes being treated as the background (known as the \emph{background shift}) and thus exacerbates catastrophic forgetting.

A common strategy is to distill knowledge adaptively from the old model, which can faithfully distinguish unannotated old classes from the background.
For example, MiB~\cite{cermelli2020modeling} calibrates regular cross-entropy (CE) and knowledge distillation (KD) losses of the background pixels with predictions from the old model. 
ALIFE~\cite{oh2022alife} further improves the calibrated CE and KD with logit regularization, and fine-tunes the classifier with feature replay.
RCIL~\cite{zhang2022representation} reparameterizes the network into two parallel branches, where the old branch is frozen for KD between intermediate layers. 
SDR~\cite{michieli2021continual} and UCD~\cite{yang2022uncertainty} introduce contrastive learning into distillation of latent representations, where pixels of the same class are clustered and pixels of different classes are separated.
PLOP~\cite{douillard2021plop}, RECALL~\cite{maracani2021recall}, SSUL~\cite{cha2021ssul}, EM~\cite{yan2021framework}, Self-Training~\cite{yu2022self}, UCD~\cite{yang2022uncertainty} and WILSON~\cite{cermelli2022incremental} explicitly use the old model to generate pseudo-labels of the old classes. 
Auxiliary data resources such as a web crawler (RECALL-Web~\cite{maracani2021recall}), a pre-trained generative model (RECALL-GAN~\cite{maracani2021recall}), large amounts of unlabeled data (Self-Training~\cite{yu2022self}), and a few old training samples (EM~\cite{yan2021framework}, SSUL~\cite{cha2021ssul}, ILLR~\cite{fortincontinual}) have been exploited to facilitate KD and prevent catastrophic forgetting. 
Besides, saliency maps are commonly used to locate unannotated objects in CSS, in response to weak supervision of only image-level annotations (WILSON~\cite{cermelli2022incremental}, ILLR~\cite{fortincontinual}), as well as defining unknown classes within the background to benefit learning plasticity (SSUL~\cite{cha2021ssul}).

In addition to the regular CSS, other relevant settings include unsupervised domain adaptation for CSS~\cite{stan2021unsupervised}, where a pre-trained backbone adapts to the unannotated data of the target domain without using the annotated data of the source domain; incremental few-shot semantic segmentation~\cite{shi2022incremental}, which performs CSS with only a few annotated data; incremental instance segmentation~\cite{gu2021class}, which incrementally learns new classes but requires individual segmentation for each instance; as well as incremental few-shot instance segmentation~\cite{ganea2021incremental}.

%\vspace{-.1cm}
\subsection{Conditional Generation}
\emph{Continual Learning for Conditional Generation} (CLCG) is closely related to generative replay, which can mitigate catastrophic forgetting of the currently-trained generative model and/or discriminative model through recovering the previously-learned data distributions.
This usually requires saving a frozen copy of the old generative model for conditional generation during a new training process.
Since the sampling distributions are often provided in both training and testing, CLCG is similar to the setting of TIL, without the challenge of task-identity prediction.

As a generic framework, DGR~\cite{shin2017continual_dgr} performs generative replay for continual learning of both discriminative and generative models.
Many efforts focus on CLCG with GANs, such as by weight regularization (ORDisCo~\cite{wang2021ordisco},~\cite{seff2017continual}), generative replay (MeRGANs~\cite{wu2018memory_mrgan}, L-VAEGAN~\cite{ye2020learning}), knowledge distillation (LifelongGAN~\cite{zhai2019lifelong_gan}, Hyper-LifelongGAN~\cite{zhai2021hyper}), parameter allocation (PiggybackGAN~\cite{zhai2020piggyback}, DGMa/DGMw~\cite{ostapenko2019learning_dgmw}, TMNs~\cite{wang2021triple}), model decomposition (Hyper-LifelongGAN~\cite{zhai2021hyper}, FILIT~\cite{chen2022few}, GAN-Memory~\cite{cong2020gan}), etc.
% adding task-specific parameters on a pre-trained backbone (GAN-Memory~\cite{cong2020gan})
Other work performs CLCG with VAE, which often relies on a multi-head architecture with encoder expansion, as well as weight regularization (VCL~\cite{nguyen2018variational}) or generative replay (CURL~\cite{rao2019continual}, VASE~\cite{achille2018life}) to overcome catastrophic forgetting.
BooVAE~\cite{egorov2021boovae} instead adopts a static architecture through incorporating each task into an additive aggregated posterior and using it as the prior for the next task.

%In general, continual learning of generative models can help discriminative models (e.g., classification models) to overcome catastrophic forgetting, although high-quality generation is not necessary. On the other hand, there are some works focusing on the performance of conditional generation itself, while the prediction of discriminative models can serve as an evaluation metric.

\vspace{-.3cm}
\subsection{Reinforcement Learning}
\emph{Continual Reinforcement Learning} (CRL) is required to address dynamic data distributions between and within tasks (corresponding to the settings of TIL and TFCL, respectively), because of the over-time interactions of the states, actions and environments.
%since the states, actions, and environments interact and change over time. 
Various types of continual learning strategies have been shown to work not only for visual classification but also for reinforcement learning (RL), such as EWC~\cite{kirkpatrick2017overcoming}, MAS~\cite{aljundi2018memory_mas}, VCL~\cite{nguyen2018variational}, P\&C~\cite{schwarz2018progress}, AFEC~\cite{wang2021afec}, UCL~\cite{ahn2019uncertainty}, AGS-CL~\cite{jung2020continual}, CPR~\cite{cha2020cpr}, Progressive Networks~\cite{rusu2016progressive}, PackNet~\cite{mallya2018packnet}, ER, A-GEM~\cite{chaudhry2018efficient_agem}, etc. 

With respect to specialized strategies, bio-inspired weight regularization (Benna-Fusi~\cite{kaplanis2018continual}) and function regularization (PC~\cite{kaplanis2019policy}) are proposed to mitigate within-task catastrophic forgetting. 
To cope with multiple tasks, OWL~\cite{kessler2022same} implements EWC~\cite{kirkpatrick2017overcoming} in a shared feature extractor of separate output heads, and adopts multi-armed bandit for task-identity prediction. 
An empirical study~\cite{isele2018selective} evaluates different principles of selecting old experiences for replay, with matching training distributions usually performing best.
%where matching training distributions usually performs the best.
%, where matching training distributions generally performs the best. 
CLEAR~\cite{rolnick2019experience} mixes off-policy learning from old experiences and on-policy learning from novel experiences, with behavioral cloning to further promote stability. 
MTR~\cite{kaplanis2020continual} employs a cascade of interacting sub-buffers to accumulate experiences at different timescales.
LPG-FTW~\cite{mendez2020lifelong} instead factorizes a policy gradient model into task-sharing and task-specific parameters.
ClonEx-SAC~\cite{wolczyk2022disentangling} investigates the effects of actor, critic, exploration and experience replay on knowledge transfer and provides a set of general recommendations.  
COMPS~\cite{berseth2021comps} considers continual meta-training of RL, using behavioral cloning to quickly learn new tasks from the knowledge of previous incremental tasks.

A distinctive feature of CRL is its high diversity of applications and benchmarks, including continuous control~\cite{kaplanis2019policy,lu2019adaptive,wolczyk2021continual,mendez2020lifelong}, maze navigation~\cite{rusu2016progressive, lomonaco2020continual}, video games (Atari~\cite{kirkpatrick2017overcoming,rusu2016progressive,wang2021afec,ahn2019uncertainty,jung2020continual,cha2020cpr}, StarCraft~\cite{schwarz2018towards}, Minecraft~\cite{tessler2017deep}, Hanabi~\cite{nekoei2021continuous}), etc., varying widely in task type and computation overhead~\cite{wolczyk2021continual}. A systematic comparison of CRL methods in various contexts is a promising future work.

%Jelly Bean World~\cite{platanios2019jelly}, robotics~\cite{wolczyk2021continual,mendez2020lifelong}, etc. 

\subsection{Natural Language Processing}
Continual learning in \emph{Natural Language Processing} (NLP) has received increasing attentions in recent years, which shares many typical settings and representative methods proposed in visual domains.
Specifically, a range of continual learning scenarios have been considered for NLP, including DIL, TIL, CIL, OCL and CPT \cite{ke2022continual_review}. Accordingly, many representative methods are adapted to these scenarios and proven to be effective, such as weight regularization (RMR-DSE\cite{li2022overcoming}, SRC~\cite{liu2019continual}), knowledge distillation (ExtendNER\cite{monaikul2021continual}, CFID~\cite{li2022continual}, CID~\cite{liu2021lifelong_cid}, PAGeR~\cite{varshney2022prompt}, LFPT5~\cite{qin2021lfpt5}, DnR~\cite{sun2020distill}, CL-NMT~\cite{cao2021continual}, COKD~\cite{shao2022overcoming}), experience replay (CFID~\cite{li2022continual}, CID~\cite{liu2021lifelong_cid}, ELLE~\cite{qin2022elle}, IDBR~\cite{huang2021continual}, MBPA++~\cite{de2019episodic}, Meta-MBPA++~\cite{wang2020efficient}, EMAR~\cite{xu2020continual}, DnR~\cite{sun2020distill}, ARPER~\cite{mi2020continual}, Total Recall~\cite{li2021total}), generative replay (PAGeR~\cite{varshney2022prompt}, LAMOL~\cite{sun2019lamol}, ACM~\cite{zhang2022continual}, NER~\cite{wang2022few}), parameter allocation (TPEM~\cite{geng2021continual}), modular network (ProgModel~\cite{shen2019progressive}), meta-learning (Meta-MBPA++~\cite{wang2020efficient}, MeLL~\cite{wang2021mell}, CML~\cite{wu2021curriculum}), etc.

On the other hand, continual learning in NLP is characterized by the extensive use of pre-training in transformer architectures, which motivates the implementations of parameter-efficient fine-tuning techniques. The pre-trained transformer can effectively accommodate successive arrivals by learning only a few task-specific parameters at a time, including \emph{adaptor-tuning} \cite{houlsby2019parameter} with inserted fully-connected layers (CPT~\cite{ke2022continual}, CLIF~\cite{jin2021learn}, AdapterCL~\cite{madotto2021continual}, ACM~\cite{zhang2022continual}, ADA~\cite{ermis2022memory}) and \emph{prompt-tuning} \cite{lester2021power} with trainable prompt tokens (C-PT~\cite{zhu2022continual}, LFPT5~\cite{qin2021lfpt5}, EMP~\cite{liu2022incremental}). 
Task specificity can also be acquired by \emph{instruction} \cite{efrat2020turking} in continual learning, which adds a short text to describe the core concept of each task (PAGeR~\cite{varshney2022prompt}, ConTinTin~\cite{yin2022contintin}, ENTAILMENT~\cite{xia2021incremental}).
Due to the great success of pre-trained foundation models, these techniques are increasingly being used and further facilitate continual learning in visual domains \cite{wang2022learning_l2p,wang2022dualprompt,wang2022continual_lifelongvit,wang2022sprompts,ermis2022memory,wang2022online,wang2023hierarchical}.

%On the other hand, the specialized architectures of (pre-trained) transformer has derived multiple NLP-adaptive techniques. For example, a pre-trained transformer can effectively accommodate downstream tasks by learning only a few task-specific parameters, including \emph{adaptor-tuning}~\cite{houlsby2019parameter} with inserted fully-connected layers (CTR~\cite{ke2021achieving}, B-CL~\cite{ke2021adapting}, CLASSIC~\cite{ke2020continual}, CPT~\cite{ke2022continual}, CLIF~\cite{jin2021learn}, AdapterCL~\cite{madotto2021continual}, ACM~\cite{zhang2022continual}, ADA~\cite{ermis2022memory}), and \emph{prompt-tuning}~\cite{lester2021power} with trainable prompt tokens (C-PT~\cite{zhu2022continual}, LFPT5~\cite{qin2021lfpt5}, EMP~\cite{liu2022incremental}). Task specificity can also be provided by an extra short text that describes the core concept of each task, i.e., \emph{instruction}~\cite{efrat2020turking}, such as PAGeR~\cite{varshney2022prompt}, ConTinTin~\cite{yin2022contintin} and ENTAILMENT~\cite{xia2021incremental}. In particular, these techniques are increasingly being used in vision domains due to the growing interest in pre-training and vision transformer~\cite{wang2022learning_l2p,wang2022dualprompt,wang2022continual_lifelongvit,wang2022sprompts,cong2020gan,ermis2022memory,wang2022online}. 

Besides, the applications of NLP in conjunction with continual learning are highly diversified, providing unique opportunities for subsequent research. Representative tasks that have been explored include dialogue system \cite{wang2019incremental,madotto2021continual,geishauser2022dynamic,geng2021continual,mi2020continual,liu2021lifelong}, text classification \cite{pasunuru2021continual,huang2021continual,xia2021incremental,de2019episodic}, sentence generation \cite{zhang2022continual,li2022overcoming,mi2020continual}, relation learning \cite{xu2020continual,qin2022continual,wu2021curriculum,ren2020two}, neural machine translation \cite{gu2020investigating,cao2021continual,garcia2021towards,shao2022overcoming}, named entity recognition \cite{wang2022few,monaikul2021continual,qin2021lfpt5}, etc.
Some other work considers multi-modality of vision and language for continual pre-training \cite{cossu2022continual,yan2022generative} or downstream tasks \cite{greco2019psycholinguistics,del2020ratt,srinivasan2022climb}.
We refer readers to their original papers for technical details.

%The applications of continual learning in NLP are also highly diverse, such as dialogue system~\cite{wang2019incremental,madotto2021continual,geishauser2022dynamic,geng2021continual,mi2020continual,liu2021lifelong,liu2021lifelong}, text classification~\cite{pasunuru2021continual,huang2021continual,xia2021incremental,de2019episodic}, sentence generation~\cite{zhang2022continual,li2022overcoming,mi2020continual}, relation learning~\cite{xu2020continual,qin2022continual,wu2021curriculum,ren2020two}, neural machine translation~\cite{gu2020investigating,cao2021continual,garcia2021towards,shao2022overcoming}, named entity recognition~\cite{wang2022few,monaikul2021continual,qin2021lfpt5}, etc. Some other work considers multimodality of vision and language for continual pre-training~\cite{cossu2022continual,yan2022generative} or downstream tasks~\cite{greco2019psycholinguistics,del2020ratt,srinivasan2022climb}. We refer readers to their original papers for technical details.

%(e.g., VQA~\cite{greco2019psycholinguistics}, image captioning~\cite{del2020ratt}, CLiMB~\cite{srinivasan2022climb})
%where adaptor-tuning~\cite{houlsby2019parameter} (inserted fully-connected layers) and prompt-tuning~\cite{lester2021power} (trainable prompt tokens) have been widely used to construct task-specific parameters in continual learning, such as CTR~\cite{ke2021achieving}, B-CL~\cite{ke2021adapting}, CLASSIC~\cite{ke2020continual}

%\subsection{Cross-Directional Inspiration}
\subsection{Beyond Task Performance}
%\hangx{Important, and could be moved to section 7 .}
Continual learning can benefit many considerations beyond task performance, such as efficiency, privacy and robustness.
A major purpose of continual learning is to avoid retraining all old training samples and thus improve \textbf{resource efficiency} of model updates, which is not only applicable to learning multiple incremental tasks, but also important for regular single-task training. 
Due to the nature of gradient-based optimization, a network tends to ``forget'' the observed training samples and thus requires repetitive training to capture a distribution, especially for some hard examples~\cite{hadsell2020embracing,arpit2017closer,chang2017active}.
Recent work has shown that the one-pass performance of visual classification can be largely improved by experience replay of hard examples~\cite{hu2021one} or orthogonal gradient projection~\cite{min2022one}.
Similarly, resolving within-task catastrophic forgetting can facilitate reinforcement learning~\cite{kaplanis2018continual,kaplanis2019policy} and stabilize the training of GANs~\cite{thanh2020catastrophic,lee2021infomax}.

Meanwhile, continual learning is relevant to two important directions of \textbf{privacy protection}. The first is \emph{Federated Learning}~\cite{mcmahan2017communication,li2020federated}, where the server and clients are not allowed to communicate with data. A typical scenario is that the server aggregates the locally trained parameters from multiple clients into a single model and then sends it back. As the incremental data collected by clients is dynamic and variable, federated learning needs to overcome catastrophic forgetting and facilitate knowledge transfer across clients, i.e., federated continual learning. 
To achieve these aims, FedWeIT~\cite{yoon2021federated} decomposes network parameters into global federated parameters and sparse task-specific parameters, with the latter selectively aggregated. 
FedSpeech~\cite{jiang2021fedspeech} adopts a set of gradual pruning masks for parameter allocation and a set of selective masks to reuse old knowledge. 
GLFC~\cite{dong2022federated} overcomes catastrophic forgetting through knowledge distillation of a few old training samples, and alleviates class imbalance through gradient compensation.
%uses knowledge distillation and experience replay to overcome catastrophic forgetting, alleviating class imbalance through gradient compensation.
CFeD~\cite{macontinual} instead performs knowledge distillation with auxillary unlabeled data.

The second is \emph{Machine Unlearning}, which aims to eliminate the influence of specific training samples when their access is lost while without affecting other knowledge. Representative methods in this direction are closely related to continual learning, such as learning separate models with subsets of training samples~\cite{bourtoule2021machine}, utilizing historical parameters and gradients~\cite{wu2020deltagrad}, removing old knowledge from parameters with Fisher information matrix~\cite{golatkar2020eternal}, adding adaptive parameters to a pre-trained backbone~\cite{golatkar2021mixed}, etc.
In the context of their intersection, retaining all old knowledge for continual learning may suffer from data leakage and privacy invasion, especially regarding data-level replay. 
Mnemonic Code~\cite{shibata2021learning} embeds a class-specific code when learning each class, enabling to selectively forget them through discarding the corresponding codes.
LIRF~\cite{ye2022learning} designs a distillation framework to remove specific old knowledge and store it in a pruned lightweight network for selective recovery.
%To selectively forget old classes in CIL, Mnemonic Code~\cite{shibata2021learning} embeds a class-specific code when learning each class, and selectively forget them through discarding the corresponding codes. 

%As different aspects of accommodating real-world dynamics, continual learning can also benefit model robustness by eliminating or resisting disturbance.
As a strategy for adapting to variable inputs, continual learning can assist a robust model to eliminate or resist external disturbances.
For example, PIGWM~\cite{zhou2021image} learns adaptive mapping functions for image de-raining and overcomes catastrophic forgetting by weight regularization. 
NACL~\cite{rostami2021detection} enables face antispoofing systems to detect potential novel attacks and remember the old ones with experience replay.
To cope with noisy labels of incremental data, SPR~\cite{kim2021continual} obtains self-supervised representations to purify a memory buffer and then performs supervised fine-tuning.
In fact, \textbf{robustness} and continual learning are intrinsically linked, as they correspond to generalizability in the spatial and temporal dimensions, respectively.
Many ideas for improving robustness to adversarial examples have been used to improve continual learning, such as flat minima~\cite{mirzadeh2020understanding,buzzega2020dark}, model ensemble~\cite{wang2022coscl}, Lipschitz continuity~\cite{bonicelli2022effectiveness} and adversarial training~\cite{yu2022continual}.
Subsequent work could further interconnect excellent ideas from both fields, e.g., designing particular algorithms and training paradigms to actively ``forget''~\cite{wang2021afec} external disturbances.

%SPR~\cite{kim2021continual} copes with noisy labeled incremental data by obtaining self-supervised representations to purify the memory buffer for experience replay and supervised fine-tuning.
%NACL~\cite{rostami2021detection} enables face antispoofing systems to detect potential novel attacks and remember the old ones through experience replay.

%with experience replay to remember the old ones.
%Continual learning can also benefit model robustness to better accommodate real-world dynamics. %, where forgetting is not always catastrophic but graceful.
%In particular, robustness theory and continual learning

\section{Discussion}\label{Sec.7_Discussion}
In this section, we present an in-depth discussion of related topics in continual learning, including our observation of current trends, cross-directional prospects and interdisciplinary connections with neuroscience.
%our observations of current trends in continual learning, and then discuss prospects for incorporating continual learning with other directions. 

\subsection{Observation of Current Trend}
As continual learning is directly affected by catastrophic forgetting, previous efforts seek to address this problem by promoting memory stability of the old knowledge. 
% , which historically provides the initial motivation~\cite{mccloskey1989catastrophic,mcclelland1995there}
% and adaptations to various applications
However, recent work has increasingly focused on facilitating learning plasticity and inter-task generalizability. % of the obtained solution. 
This essentially advances the understanding of continual learning:
a desirable solution requires a proper balance between the old and new tasks, with adequate generalizability to accommodate their distribution differences.
%in order to capture the distributions of both the old and new tasks, a desirable solution requires a proper balance between them and an adequate generalizability of their distribution differences.
%a desirable solution should be sufficiently compatible with both old and new tasks, which requires a proper balance between these two aspects and an adequate generalization of their distribution differences.

To promote learning plasticity on the basis of memory stability, emergent strategies include renormalization of the old and new task solutions~\cite{wang2021afec,schwarz2018progress,lin2022towards,hou2018lifelong}, balanced exploitation of the old and new training samples~\cite{hou2019learning_lucir,wu2019large_bic,zhao2020maintaining,ahn2021ss,wang2022continual_dri,hu2021distilling}, reserving space for subsequent tasks~\cite{ahn2019uncertainty,jung2020continual,zhou2022forward}, etc. %balanced incorporation
On the other hand, solution generalizability can be explicitly improved by converging to a flat loss landscape, where representative strategies include optimizing the flatness metric~\cite{shi2021overcoming,mehta2021empirical,mirzadeh2020understanding,deng2021flattening,mirzadeh2020linear}, constructing an ensemble model at either spatial scale~\cite{cha2020cpr,wang2022coscl,wang2023incorporating} or temporal scale~\cite{buzzega2020dark,boschini2022class}, and obtaining well-distributed representations~\cite{ramasesh2021effect,mehta2021empirical,hu2021well,pham2021dualnet,cha2021co2l,madaan2021representational}.
In particular, since self-supervised and pre-trained representations are naturally more robust to catastrophic forgetting~\cite{hu2021well,ramasesh2021effect,madaan2021representational,mehta2021empirical}, creating, improving and exploiting such representational advantages has become a promising direction.
%have been shown to be naturally robust to catastrophic forgetting~\cite{hu2021well,ramasesh2021effect,madaan2021representational,mehta2021empirical}, such representation-based approaches receive increasing attentions.

We also observe that the applications of continual learning are becoming more diverse and widespread.
In addition to various scenarios of visual classification, current extensions of continual learning have covered many other vision domains, as well as other areas such as RL, NLP and ethic considerations.
We present only some representative applications in the above two sections, with other more specialized and cross-cutting scenarios to be explored, such as robotics~\cite{lesort2020continual,churamani2020continual,ayub2022few}, graph learning~\cite{wang2022lifelong,kimdygrain,zhang2022cglb}, bioimaging~\cite{zhang2022epicker}, etc.
Notably, existing work on applications has focused on providing basic benchmarks and baseline methods. 
Future work could develop more specialized methods to obtain stronger performance, or evaluate the generality of current methods in different applications.

\subsection{Cross-Directional Prospect}
Continual learning demonstrates vigorous vitality, as most of the state-of-the-art AI models require flexible and efficient updates, and their advances have contributed to the development of continual learning. %exuberance and sustainability
Here, we discuss some attractive intersections of continual learning with other topics of the broad AI community: %other AI domains:

\emph{Diffusion Model}~\cite{song2020score,dhariwal2021diffusion,bao2021analytic,lu2022dpm-solver} is a rising state-of-the-art generative model, which constructs a Markov chain of discrete steps to progressively add random noise for the input and learns to gradually remove the noise to restore the original data distribution. This provides a new target for continual learning of generative models, and its outstanding performance in conditional generation can further facilitate the efficacy of generative replay.

%\emph{Neural Compression}~\cite{mentzer2020high,agustsson2017soft,wang2022compress} instead exploits generative models to optimize the storage overhead of input data without much distortion, which can benefit both efficacy and efficiency of experience replay~\cite{wang2021memory}. A promising direction is to learn the mapping function of data compression adaptively from the distribution of incremental data, in order to obtain a better compression rate.
%In contrast, the efficacy and storage efficiency of experience replay can benefit from advanced \emph{Neural Compression}~\cite{wang2021memory}, which also relies on generative models, but optimizes the storage overhead of training samples without much distortion~\cite{mentzer2020high,agustsson2017soft}. 

\emph{Foundation Model}~\cite{devlin2018bert,brown2020language,radford2021learning} acquires impressive results in a variety of downstream tasks due to the effective use of large-scale pre-training.
%relies on large-scale pre-training to acquire impressive results in downstream tasks~\cite{bommasani2021opportunities,han2021pre}. 
%, such as BERT~\cite{devlin2018bert}, GPT~\cite{brown2020language}, CLIP~\cite{radford2021learning}, etc.,
The pre-training data is usually huge in volume and collected incrementally, creating urgent demands for efficient updates. Meanwhile, increasing the scale of pre-training would facilitate knowledge transfer and mitigate catastrophic forgetting between downstream tasks~\cite{ramasesh2021effect,mehta2021empirical}. However, fine-tuning the foundation model tends to forget pre-trained knowledge \cite{lin2023speciality}, which has become a central challenge for its application and requires specialized continual learning strategies. %to fit downstream tasks 

\emph{Transformer-Based Architecture}~\cite{liu2021swin,vaswani2017attention} has proven effective for both language and vision domains, and become the dominant choice for state-of-the-art foundation models. This requires specialized designs to overcome catastrophic forgetting~\cite{wang2022continual_lifelongvit,wang2022online} while providing new insights for maintaining task specificity in continual learning~\cite{wang2022dualprompt,wang2022sprompts}. The parameter-efficient transfer learning techniques developed in NLP can serve as a good reference and are being widely adapted to continual learning.

%\emph{Vision Transformer}~\cite{liu2021swin,khan2022transformers} extends the success of transformer-based architectures in NLP to numerous vision domains. This requires specialized designs to overcome catastrophic forgetting~\cite{wang2022continual_lifelongvit,wang2022online} while providing new insights for maintaining task specificity in continual learning~\cite{wang2022learning_l2p,wang2022dualprompt,wang2022sprompts}. Similarly, specialized techniques for continual learning in NLP can serve as a good reference.

\emph{Multi-Modality}~\cite{radford2021learning,openai2023gpt4,kirillov2023segment}, especially for contrastive learning of vision-language pairs and language-grounded applications in vision domains, has largely advanced many directions in machine learning. As an additional source of supervision, stabilization of multi-modal information can potentially mitigate catastrophic forgetting. The use of (multi-modal) large-language models (LLMs), in particular, provides strong ability of predicting task identity in both training and testing.

\emph{Embodied AI}~\cite{franklin1997autonomous, chrisley2003embodied, duan2022survey}, an emerging paradigm shift from the era of ``internet AI'', aims to enable AI algorithms and agents to learn through interactions with their environments rather than datasets of images, videos or texts collected primarily from the Internet. 
Of important, the study of general continual learning helps the embodied agents to learn from an egocentric perception similar to humans, and provides a unique opportunity for researchers to pursue the essence of lifelong learning by observing the same person in a long time span~\cite{wang2021wanderlust,gadre2022continuous}.

%The transformer-based architecture initially succeeds in NLP and is recently extend to vision domains, i.e., Vision Transformer.

%Here, we discuss some attractive cross-directional opportunities.
%Previous efforts on continual learning of generative models and generative replay were built on VAE or GANs. In contrast, \emph{Diffusion Model} is a rising state-of-the-art generative model that constructs a Markov chain of discrete steps to progressively add random noise for the input and learns to gradually remove the noise to restore the original data distribution.

%ai field  (diffusion for generative replay, neural compression for experience replay, clip / gpt for representation, transformer architecture for CV / NLP, adversarial robustness, embodied ai)

\subsection{Connection with Neuroscience}
Inspirations from neuroscience have played an important role in the development of continual learning. As biological learning is naturally on a continual basis~\cite{kudithipudi2022biological,hadsell2020embracing,zhao2023genetic}, its underlying mechanisms provide an excellent reference for AI models. 
Here, we briefly overview the neurological basis of continual learning strategies, from the levels of synaptic plasticity to regional collaboration. %synaptic plasticity to coordinating brain regions.

Biological neural networks are capable of flexibly modulating \textbf{synaptic plasticity} in response to dynamic inputs, including
% the following aspects:
(1) stabilization of previously-learned synaptic changes to overcome subsequent interference~\cite{zhang2018active,yang2009stably,hayashi2015labelling}, which motivates the strategies of weight regularization to selectively penalize parameter changes~\cite{kirkpatrick2017overcoming,aljundi2018memory_mas,zenke2017continual_si}; (2) number expansion and pruning of functional connections to provide flexibility for new memory formation~\cite{richards2017persistence,dong2016inability,bailey2015structural}, which derives the expansion-renormalization paradigm of creating extra space for learning the current task and renormalizing it with the previous one~\cite{wang2021afec,schwarz2018progress,wang2023incorporating}; (3) activity-dependent and persistent regulation of synaptic plasticity, i.e., meta-plasticity or ``plasticity of synaptic plasticity''~\cite{abraham1996metaplasticity,abraham2008metaplasticity,clem2008ongoing}, corresponding to the use of meta-learning~\cite{finn2017model,javed2019meta,riemer2018learning}; and
(4) inhibitory synapses for excited neurons to reduce the activity of other neurons~\cite{cayco2019re,arevian2008activity,inada2017origins,espinoza2018parvalbumin}, which acts similarly to the binary mask for parameter allocation~\cite{serra2018overcoming,wang2021triple}.
%meta-plasticity of updating a pair of fast and slow weights over time, consistent with the use of meta-learning to explicitly balance specificity and generalization;

As for \textbf{regional collaboration}, the \emph{complementary learning system} (CLS) theory~\cite{mcclelland1995there,kumaran2016learning} has been widely used to inspire continual learning, which attributes the advantages of biological learning and memory to the complementary functions of hippocampus and neocortex. The hippocampus is responsible for rapid acquisition of separated representations of specific experiences, while the neocortex enables progressive acquisition of structured knowledge for generalization. In particular, the hippocampus can replay neural representations of the previous experiences, so its function is often represented by a memory buffer 
\cite{rostami2019complementary,rolnick2019experience,wang2021memory,wang2022continual_dri,boschini2022class} or a generative model~\cite{wang2021triple,van2020brain,shin2017continual_dgr,wu2018memory_mrgan,rostami2020generative} to recover the old data distributions. 
Instead of using the original data format, however, the hippocampal replay is in a temporally compressed form (e.g., an experience of around $6$ seconds can be compressed into $0.2$ seconds)~\cite{kudithipudi2022biological,davidson2009hippocampal,carr2011hippocampal}, which potentially improves resource efficiency. Such biological advantage is further incorporated by replaying compressed data~\cite{wang2021memory} or feature representations~\cite{ayub2020eec,hayes2020remind}.
In contrast, the neocortex-dependent memories are typically long-term and generalizable, where the representations can be progressively acquired in an unsupervised fashion~\cite{doya1999computations,doya2000complementary} and exhibit orthogonality~\cite{flesch2022orthogonal,xie2022geometry}, consistent with the advantages of self-supervised and pre-trained representations in continual learning~\cite{pham2021dualnet,hu2021well,ramasesh2021effect,madaan2021representational,mehta2021empirical}. 
% topology preservation
Besides, some regions of the biological brain have a modular architecture similar to the mixture of experts~\cite{jacobs1991adaptive}, such as the prefrontal cortex (i.e., a neocortex region) regulating sensory information from various cortical modules~\cite{tsuda2020modeling,frankland2005organization,ott2019dopamine} and the mushroom body (i.e., a memory center in fruit fly) coordinating multiple parallel compartments to process the sequentially-arrived experiences~\cite{cohn2015coordinated,waddell2016neural,modi2020drosophila,aso2016dopaminergic,aso2014mushroom}, corresponding to the use of modular networks in continual learning~\cite{wang2022coscl,aljundi2017expert,ramesh2021model}.

%Encouragingly, there is also a growing number of neuroscience research inspired by continual learning~\cite{hadsell2020embracing,kudithipudi2022biological}, such as the function of sleep in overcoming catastrophic forgetting. We expect to see the continual learning as a bridge between AI and BI. 

\section{Conclusion}\label{Sec.8_Conclusion}
In this work, we present an up-to-date and comprehensive survey of continual learning, bridging the latest advances in theory, method and application. We summarize both general objectives and particular challenges in this field, with an extensive analysis of how representative methods address them.
Encouragingly, we observe a growing and widespread interest in continual learning from the broad AI community, bringing novel understandings, diversified applications and cross-directional opportunities. 
%With exploring and exploiting the neurological underpinnings, 
Based on such a holistic perspective, we expect the development of continual learning to eventually empower AI systems with human-like adaptability, responding flexibly to real-world dynamics and evolving themselves in a lifelong manner.
\ifCLASSOPTIONcompsoc
  \section*{Acknowledgments}
\else
  \section*{Acknowledgment}
\fi
This work was supported by the National Key Research and Development Program of China (2020AAA0106000, 2020AAA0104304, 2020AAA0106302, 2021YFB2701000), NSFC Projects (62061136001, 62106123, 62076147, U19B2034, U1811461, U19A2081, 61972224), BNRist (BNR2022RC01006), the Tsinghua Institute for Guo Qiang, Tsinghua-OPPO Joint Research Center for Future Terminal Technology, and the High Performance Computing Center, Tsinghua University. L.W. was also supported by Shuimu Tsinghua Scholar.

\ifCLASSOPTIONcaptionsoff
  \newpage
\fi

% the document is modified latter


%\bibliographystyle{IEEEtran}

%\clearpage
\bibliographystyle{ieee_fullname}
\bibliography{egbib}

%\begin{thebibliography}{1}
%H.~Kopka and P.~W. Daly, \emph{A Guide to \LaTeX}, 3rd~ed.\hskip 1em plus 0.5em minus 0.4em\relax Harlow, England: Addison-Wesley, 1999.
%\end{thebibliography}


%\begin{IEEEbiography}{Liyuan Wang}
%\end{IEEEbiography}

\clearpage
\vspace{+0.2cm}

\begin{IEEEbiography}
[{\includegraphics[width=1in,height=1.25in,clip,keepaspectratio]{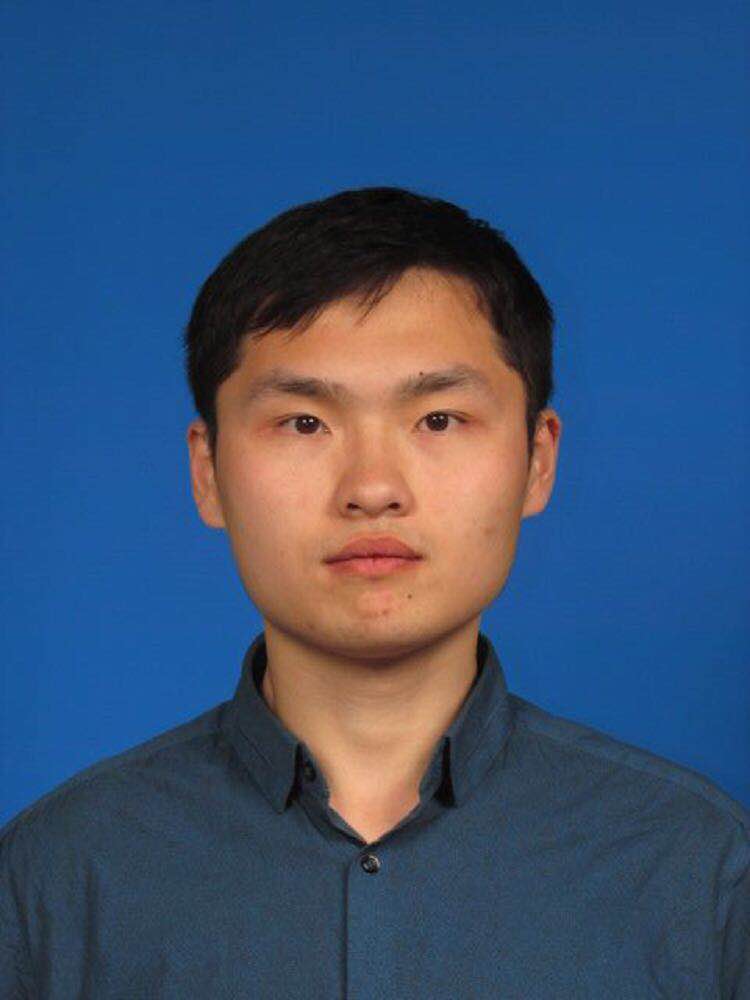}}]{Liyuan Wang}
is currently a postdoc in Tsinghua University, working with Prof.~Jun Zhu at the Department of Computer Science and Technology. Before that, he received the BS and PhD degrees from Tsinghua University. His research interests include continual learning, incremental learning, lifelong learning and brain-inspired AI. His work in continual learning has been published in major conferences and journals in related fields, such as Nature Machine Intelligence, NeurIPS, ICLR, CVPR, ICCV, ECCV, etc.
%is currently a PhD student at Tsinghua University, co-advised by Prof. Yi Zhong at the school of life sciences and Prof. Jun Zhu at the department of computer science and technology. Before that, he received the BS degree from Tsinghua University. His research interests include continual learning, incremental learning, lifelong learning and brain-inspired AI. His work in continual learning has been published in major conferences and journals in related fields, such as NeurIPS, ICLR, CVPR, ECCV, TNNLS, etc.
\end{IEEEbiography}

\begin{IEEEbiography}
[{\includegraphics[width=1in,height=1.25in,clip,keepaspectratio]{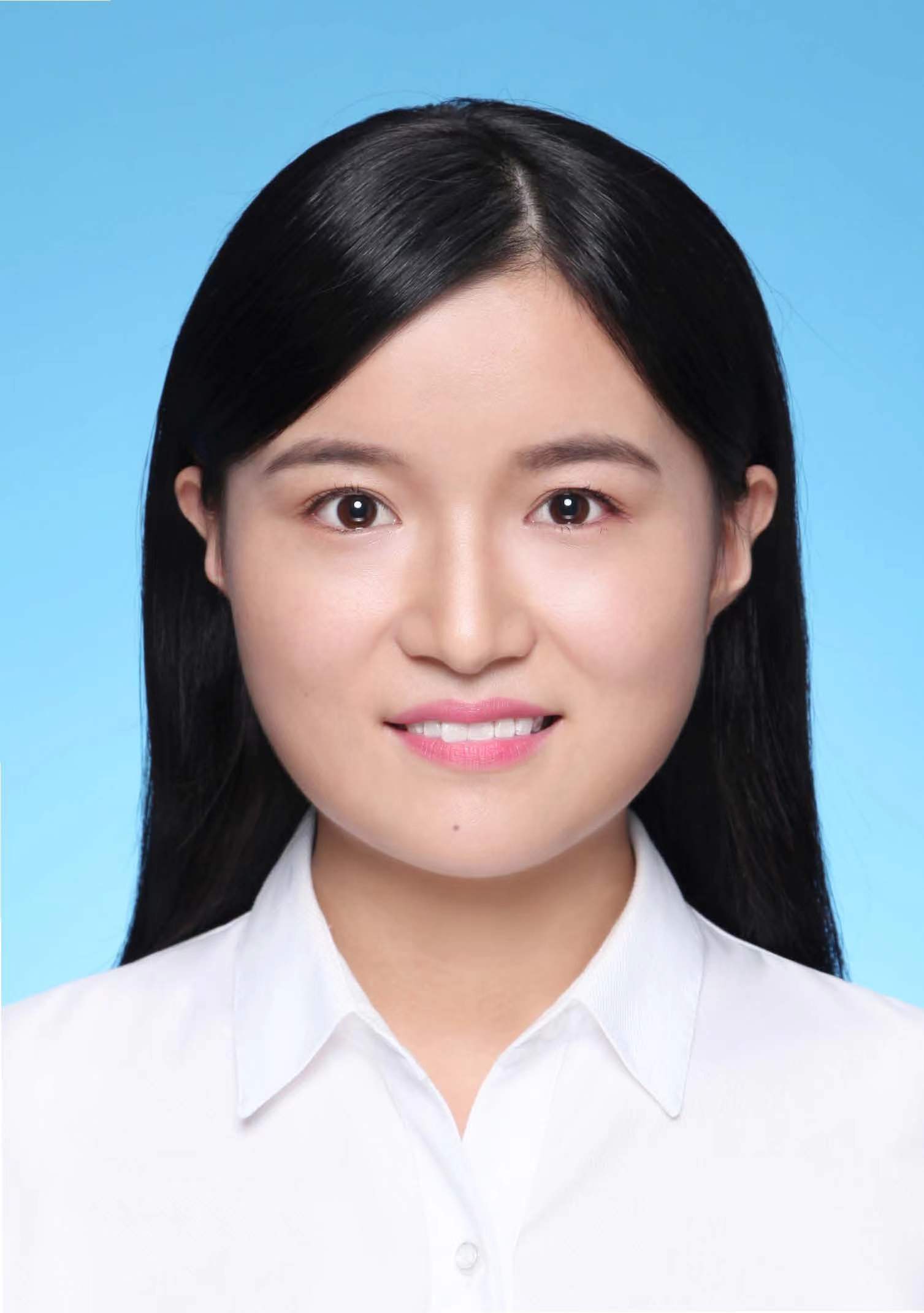}}]{Xingxing Zhang}
received the Ph.D. degree in signal and information processing from the Institute of Information Science, Beijing Jiaotong University (BJTU), Beijing, China, in 2020, and B.E. degree in 2015. She was also a Visiting Student with the Department of Computer Science, University of Rochester, USA, from 2018 to 2019. She was a Postdoc in the department of computer science, Tsinghua University, Beijing, China, from 2020 to 2022. Her research interests include continual learning, zero/few-shot learning, and data selection. She has received the excellent Ph.D. thesis award from the Chinese Institute of Electronics (CIE) in 2020.
\end{IEEEbiography}

\begin{IEEEbiography}
[{\includegraphics[width=1in,height=1.25in,clip,keepaspectratio]{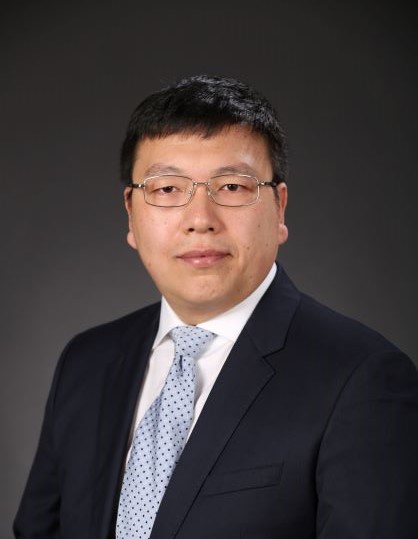}}]{Hang Su},
IEEE member, is an associated professor in the department of computer science and technology at Tsinghua University. His research interests lie in the adversarial machine learning and robust computer vision, based on which he has published more than 50 papers including CVPR, ECCV, TMI, etc. He has served as area chair in NeurIPS and the workshop co-chair in AAAI22. he received ``Young Investigator Award” from MICCAI2012, the ``Best Paper Award'' in AVSS2012, and ``Platinum Best Paper Award'' in ICME2018.
\end{IEEEbiography}

\begin{IEEEbiography}
[{\includegraphics[width=1in,height=1.25in,clip,keepaspectratio]{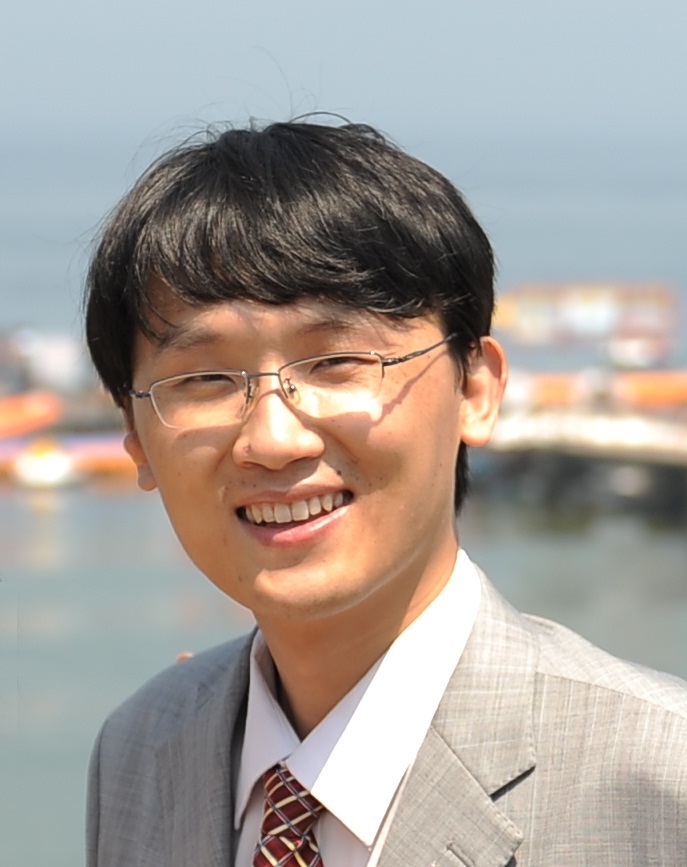}}]{Jun Zhu}  
received his BS and PhD degrees from the Department of Computer Science and Technology in Tsinghua University, where he is currently a Bosch AI professor. He was an adjunct faculty and postdoctoral fellow in the Machine Learning Department, Carnegie Mellon University. His research interest is primarily on developing machine learning methods to understand scientific and engineering data arising from various fields. He regularly serves as senior Area Chairs and Area Chairs at prestigious conferences, including ICML, NeurIPS, ICLR, IJCAI and AAAI. He was selected as ``AI's 10 to Watch'' by IEEE Intelligent Systems. He is a Fellow of the IEEE and an associate editor-in-chief of IEEE TPAMI. 
\end{IEEEbiography}



%\vfill


\end{document}